\documentclass[onefignum,onetabnum]{siamonline190516}

\usepackage{amsfonts}
\usepackage{graphicx}
\usepackage{epstopdf}

\ifpdf
  \DeclareGraphicsExtensions{.eps,.pdf,.png,.jpg}
\else
  \DeclareGraphicsExtensions{.eps}
\fi

\usepackage[inline]{enumitem}
\setlist[enumerate]{leftmargin=.5in}
\setlist[itemize]{leftmargin=.5in}

\newsiamremark{remark}{Remark}
\newsiamremark{hypothesis}{Hypothesis}
\newsiamremark{example}{Example}
\crefname{hypothesis}{Hypothesis}{Hypotheses}
\newsiamthm{claim}{Claim}

\headers{UQ for Large Linear Inverse Problems}{C. Travelletti, D. Ginsbourger, and N. Linde}

\title{Uncertainty Quantification and Experimental Design for Large-Scale Linear Inverse Problems under Gaussian Process Priors
\thanks{\funding{This work was funded by the Swiss National Science Fundation (SNF) through project no.~178858.}}}

\author{C\'{e}dric Travelletti\thanks{Institute of Mathematical Statistics and
    Actuarial Science, University of
    Bern, Bern, Switzerland 
  (\email{cedric.travelletti@stat.unibe.ch}).}
\and David Ginsbourger\thanks{Institute of Mathematical Statistics and
    Actuarial Science, University of
    Bern, Bern, Switzerland 
  (\email{david.ginsbourger@stat.unibe.ch})}
\and Niklas Linde\thanks{Institute of Earth Sciences, University of
    Lausanne, Lausanne, Switzerland 
  (\email{niklas.linde@unil.ch})}}

\usepackage{amsopn}

\usepackage{bm}
\usepackage{amssymb}

\usepackage{color}

\usepackage{array}
\usepackage{booktabs}

\usepackage{natbib}

\usepackage{tensor}

\usepackage{amsmath,mathtools}

\usepackage[noend]{algpseudocode}

\usepackage{subcaption}

\usepackage{xifthen}

\newcommand{\domain}{D}

\newcommand{\density}{\rho}

\newcommand{\unknownfunc}{\rho}

\newcommand{\noise}{\bm{\epsilon}}

\newcommand{\site}{s}

\newcommand{\candidateSites}{\bm{S}_c}

\newcommand{\nextSite}{\site_{\stage+1}}

\newcommand{\fwd}[1][]{
    \ifthenelse{\isempty{#1}}
    {G}
    {G_{#1}}
}

\newcommand{\predpts}{\bm{X}}
\newcommand{\predptsprime}{\bm{X}'}

\newcommand{\ptsfwd}{\bm{W}}

\newcommand{\datvals}{\bm{y}}

\newcommand{\randdatvals}{\bm{Y}}

\newcommand{\invop}{R}
\newcommand{\pushfwd}{\bar{K}}

\newcommand{\dimpred}{m}

\newcommand{\dimgrid}{m}

\newcommand{\datdim}{p}
\newcommand{\datdimTot}{P}

\newcommand{\dimfwd}{r}

\newcommand{\dimA}{q}

\newcommand{\somestage}{i}

\newcommand{\stage}{n}

\newcommand{\sumIndexI}{i}
\newcommand{\sumIndexJ}{j}
\newcommand{\sumIndexK}{k}

\newcommand{\matrixFwdN}{\underbar{G}_{\stage}}

\newcommand{\ptsfwdN}{\ptsfwd_{\stage}}

\newcommand{\pushfwdN}{\pushfwd_{\stage}}

\newcommand{\datvalsN}{\datvals_{\stage}}
\newcommand{\invopN}{\invop_{\stage}}
\newcommand{\invopUpdt}{S_{\stage}}
\newcommand{\covN}{K^{(\stage)}}
\newcommand{\muN}{m^{(\stage)}}

\newcommand{\meanPrior}{m^{(0)}}
\newcommand{\covPrior}{K^{(0)}}

\newcommand{\datdimN}{\datdim_{\somestage}}

\newcommand{\covNprev}{K^{(\stage-1)}}
\newcommand{\muNprev}{m^{(\stage-1)}}

\newcommand{\matrixFwdI}{\underbar{G}_{\somestage}}

\newcommand{\ptsfwdI}{\ptsfwd_{\somestage}}
\newcommand{\datvalsI}{\datvals_{\somestage}}
\newcommand{\randdatvalsI}{\randdatvals_{\somestage}}
\newcommand{\invopI}{\invop_{\somestage}}
\newcommand{\pushfwdI}{\pushfwd_{\somestage}}

\newcommand{\covI}{K^{(\somestage)}}
\newcommand{\muI}{m^{(\somestage)}}

\newcommand{\covIprev}{K^{(\somestage-1)}}

\newcommand{\datdimI}{\datdim_{\somestage}}

\newcommand{\currmean}[1][]{
   \ifthenelse{\isempty{#1}}
   {\mu_{\predpts}^{(\stage)}}
   {\mu_{#1}^{(\stage)}}
}
\newcommand{\prevmean}[1][]{
   \ifthenelse{\isempty{#1}}
   {\mu_{\predpts}^{(\stage-1)}}
   {\mu_{#1}^{(\stage-1)}}
}

\newcommand{\currcov}[1][]{
   \ifthenelse{\isempty{#1}}
   {K_{\predpts\predptsprime}^{(\stage)}}
   {K_{#1}^{(\stage)}}
}
\newcommand{\prevcov}[1][]{
   \ifthenelse{\isempty{#1}}
   {K_{\predpts\predptsprime}^{(\stage-1)}}
   {K_{#1}^{(\stage-1)}}
}

\newcommand{\currweights}[1]{
    \lambda_{\stage}\left(#1\right)
}

\newcommand{\currweightsI}[1]{
    \lambda_{\somestage}\left(#1\right)
}

\newcommand{\invopNexpr}{
    \matrixFwdN \covNprev_{\ptsfwdN \ptsfwdN}
        \matrixFwdN^T + \datvar\id
}
\newcommand{\id}{\bm{I}}

\newcommand{\datvar}{\tau^2}

\newcommand{\trueExcuSet}{\Gamma^{*}}
\newcommand{\excuSet}{\Gamma}
\newcommand{\thresh}{T}

\newcommand{\vol}{V}

\newcommand{\dimChunk}{\datdim_c}

\newcommand{\nchunks}{n_c}

\newcommand{\gp}[1][]{
    \ifthenelse{\isempty{#1}}
    {Z}
    {Z_{#1}}
}

\newcommand{\expec}[1]{\mathbb{E}
\left[#1\right]}

\newcommand{\covFunN}[1][]{
    \ifthenelse{\isempty{#1}}
    {p_{\Gamma}^{(\somestage)}}
    {p_{\Gamma}^{(\somestage)}({#1})}
}

\DeclareMathOperator*{\argmin}{arg\,min}

\ifpdf
\hypersetup{
  pdftitle={UQ for Large Linear Inverse Problems},
  pdfauthor={C. Travelletti, D. Ginsbourger, and N. Linde}
}
\fi

\begin{document}

\maketitle

\begin{abstract}
We consider the use of Gaussian process (GP) priors for solving inverse problems in a Bayesian framework. As is well known, the computational complexity of GPs scales cubically in the number of datapoints. We here show that in the context of inverse problems involving integral operators, one faces additional difficulties that hinder inversion on large grids. Furthermore, in that context, covariance matrices can become too large to be stored. 
By leveraging recent results about sequential disintegrations of Gaussian measures, we are able 
to introduce an implicit representation of posterior covariance matrices that reduces the memory footprint by only storing low rank intermediate matrices, while allowing individual elements to be accessed on-the-fly without needing to build full posterior covariance matrices. Moreover, it allows for fast sequential inclusion of new observations. These features are crucial when considering sequential experimental design tasks. We demonstrate our approach by computing sequential data collection plans for excursion set recovery for a gravimetric inverse problem, where the goal is to provide fine resolution estimates of high density regions inside the Stromboli volcano, Italy. Sequential data collection plans are computed by extending the weighted integrated variance reduction (wIVR) criterion to inverse problems. Our results show that this criterion is able to significantly reduce the uncertainty on the excursion volume, reaching close to minimal levels of residual uncertainty.
Overall, our techniques allow the advantages of probabilistic models to be brought to bear on large-scale inverse problems arising in the natural sciences. Particularly, applying the latest developments in Bayesian sequential experimental design on realistic large-scale problems opens new venues of research at a crossroads between mathematical modelling of natural phenomena, statistical data science and active learning.
\end{abstract}

\section{Introduction}\label{sec:introduction}
Gaussian processes (GP) provide a powerful Bayesian approach to regression
\citep{rasmussen_williams}. While traditional regression considers pointwise evaluations
of an unknown function, GPs can also include data in the
form of linear operators \citep{solak2003derivative,sarkka2011linear,jidling2017linearly,mandelbaum,TARIELADZE2007851,hairer_stuart,owhadi_scovel,klebanov_sullivan}. This allows GPs to provide a Bayesian framework to address inverse problems
\citep{tarantola,stuart_2010,stuart_dashti}. Even though GP priors have been shown to perform well on smaller-scale inverse problems, difficulties arise when one tries to apply them to large inverse problems (be it high dimensional or high resolution) and these get worse when one considers sequential data assimilation settings such as in \citet{chevalier_2014_fast}. The main goal of this work is to overcome these difficulties and to provide solutions for scaling GP priors to large-scale Bayesian inverse problems.\\

Specifically, we focus on a triple intersection of 
i) linear operator data, 
ii) large 
number of prediction points and iii) sequential data acquisition.
There exists related works that focus on some of these items individually. For example, methods for extending Gaussian processes to large datasets
\citep{hensman_big,exact_million} or to a large number of prediction points
\citep{wilson_large_sampling} gained a lot of attention over the last years. Also, much work has been devoted to extending GP regression to 
include linear constraints \citep{jidling2017linearly} or integral 
observations \citep{hendriks2018evaluating,jidling2019deep}. On the sequential side, methods have been developed relying on infinite-dimensional state-space representations 
\citep{sarkka_2013} and have also been extended to variational GPs \citep{hamelijnck2021spatiotemporal}. 
There are also works that focus on the three aspects at the same time \citep{solin_2015}. We note that all these approaches rely on approximations, while our method does not.

The topic of large-scale sequential assimilation of linear
operator data has also been of central interest in the Kalman filter community.
To the best of our knowledge, techniques employed in this framework usually 
rely on a low rank representation of the
covariance matrix, obtained either via factorization
\citep{kitanidis_large_kalman} or from an ensemble estimate \citep{mandel_efficient_ensemble_kalman}. 
Our goal in this work is to elaborate similar methods for Gaussian processes
without relying on a particular factorization of the covariance matrix. As stated above, we
focus on the case where: 
i) the number of prediction points is large, 
ii) the data has to be assimilated sequentially, and 
iii) it comes in the form of integral operators observations.
Integral operators are harder to handle than pointwise observations since, when discretized on a grid (which is the usual inversion approach), they turn into a matrix with entries that are not predominantly null, in our
case non-zero for most grid points. For the
rest of this work, we will only consider settings that enjoy these three properties.
This situation is typical 
of Bayesian large-scale inverse problems because those are often solved on a discrete
grid, forcing one to consider a large number of prediction points when inverting
at high resolution; besides, the linear operators found in
inverse problems are often of integral form (e.g. gravity,
magnetics).
~\\
\begin{example*}
    For the rest of this work, we will use as red thread a real-world inverse problem that enjoys the above properties. This problem is that of reconstructing the underground mass density inside the Stromboli volcano, Italy, from observations of the vertical component of the gravity field at different locations of the surface of the volcano (with respect to a reference station), see \cref{fig:problem_overview_intro}.
\begin{figure}[tbh!p]
\centering
\subfloat[]{\includegraphics[scale=0.09]{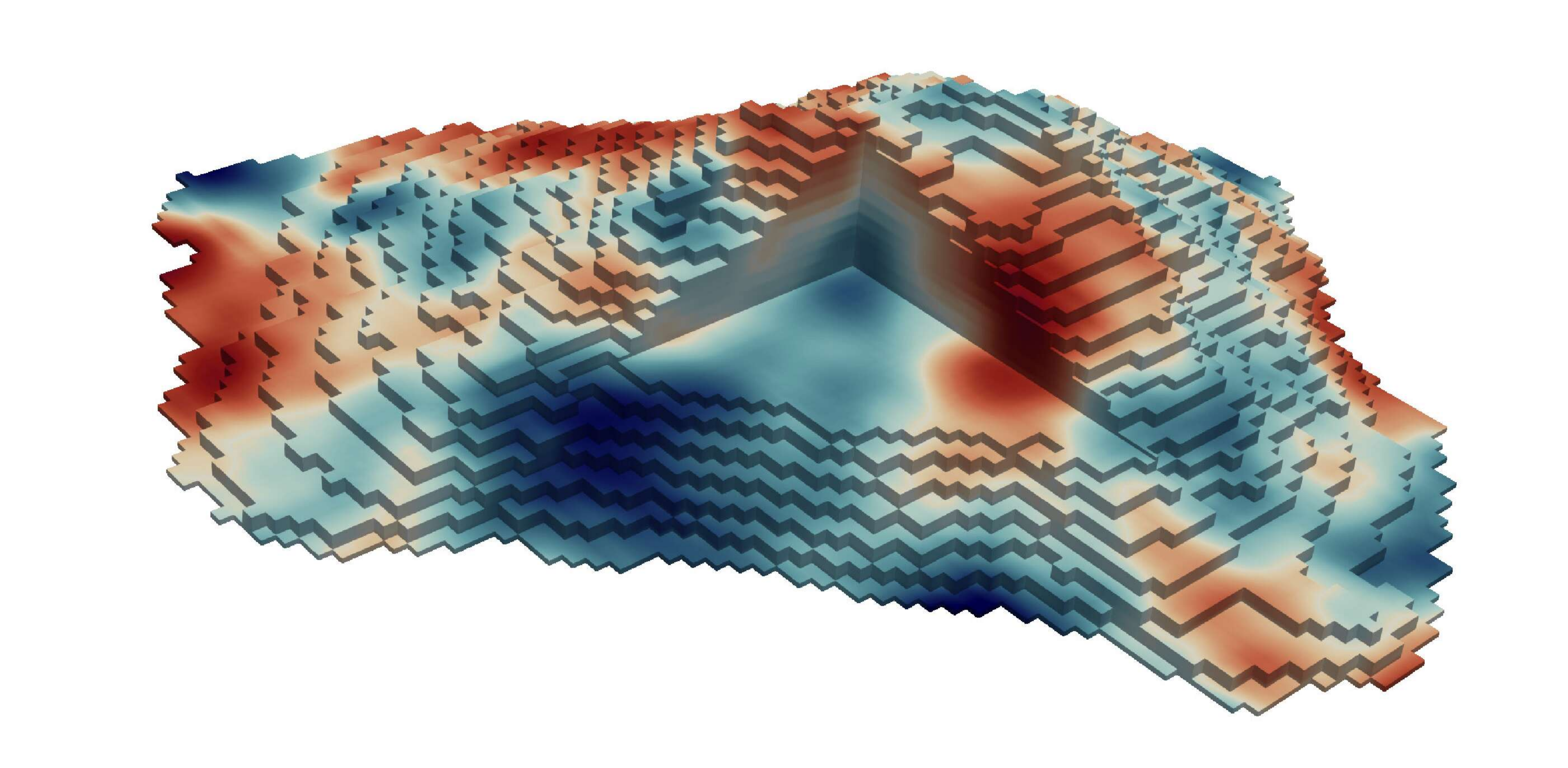}}
\subfloat[]{\includegraphics[scale=0.09]{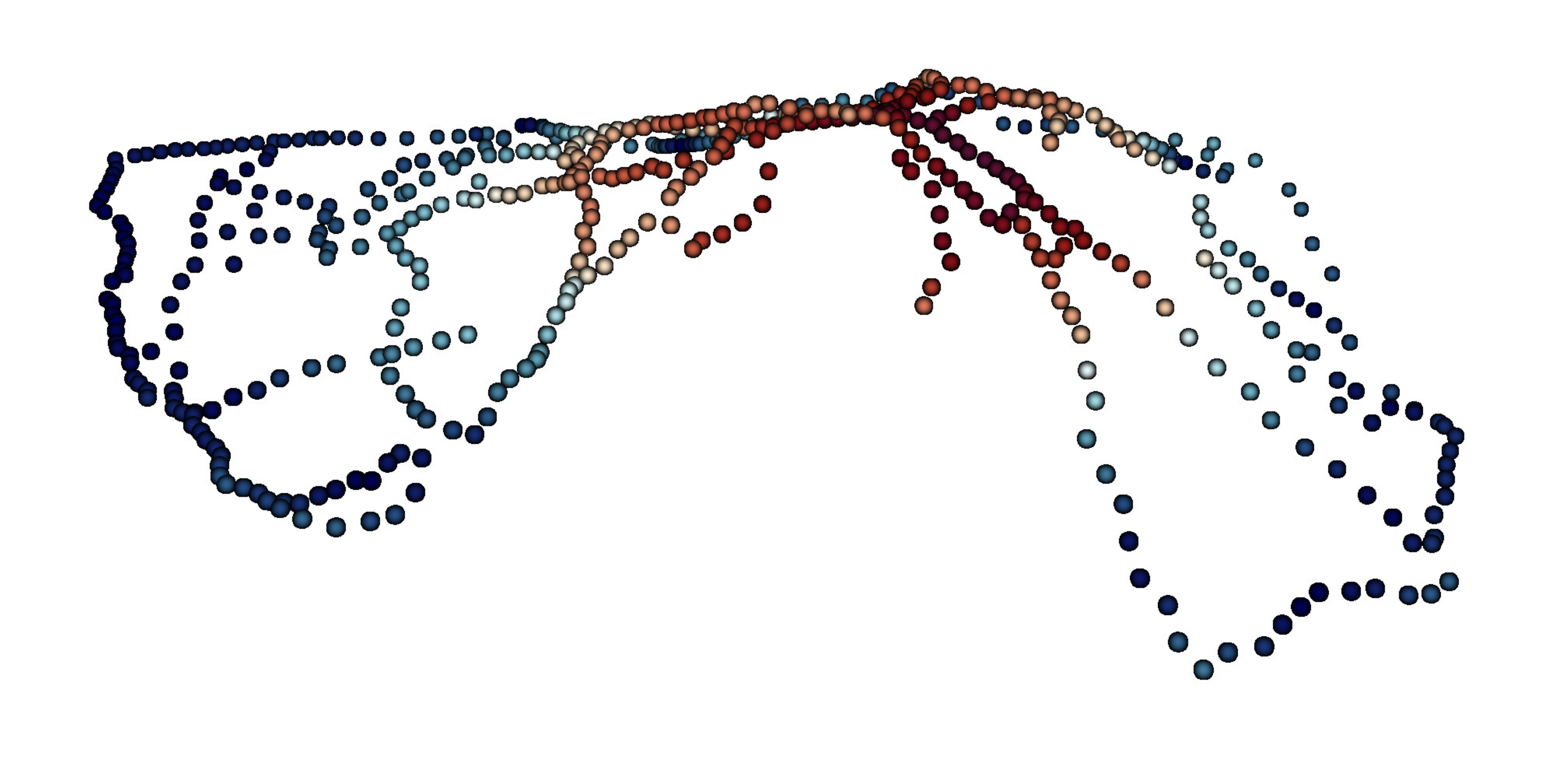}}\\
\caption{Example inverse problem: (a) Simulated underground mass density inside the Stromboli volcano (realisation from GP prior). (b) vertical intensity of the generated gravity field at locations where data has been gathered \citep{linde}. Colorscales were chosen arbitrarily.}
\label{fig:problem_overview_intro}
\end{figure}
We will use a real dataset of gravimetric observations that was collected during a field campaign in 2012 (courtesy of \citet{linde}). In \citet{linde} the inversion domain is discretized at $50~[m]$ resolution, which, as we explain in \cref{sec:applications} results in larger-than-memory covariance matrices, calling for the methods developed in this work.
\end{example*}

Our main contribution to overcome the above difficulties is the introduction of 
an implicit representation of the posterior covariance matrix that only
requires storage of low rank intermediate matrices and allows 
individual elements to be accessed on-the-fly, without ever storing the full
matrix. Our method relies on an extension of the \textit{kriging update formulae} 
\citep{update_chevalier,update_emery,update_gao,update_barnes} 
to linear operator observations. 
As a minor contribution, we also provide a technique for computing posterior means on fine 
discretizations using a chunking technique and explain
how to perform posterior simulations in the considered setting. The developed implicit representation allows for fast updates of posterior covariances under linear
operator observations on very large grids. This is particularly useful when
computing sequential data acquisition plans for inverse problems, which
we demonstrate by computing sequential experimental designs for excursion set learning in
gravimetric inversion. We find that our method provides significant
computational time savings over brute-force conditioning and scales to problem
sizes that are too large to handle using state-of-the-art techniques.\\

Although the contribution of this work is mostly methodological, the techniques developed here 
can be formulated in a rigorous abstract framework by using the language of Gaussian 
measures an disintegrations. Though less well-known than GPs, 
Gaussian measures provide an equivalent approach \citep{rajput_gp_vs_measures} 
that offers several advantages. First, it is more natural for discretization-independent formulations; second, it provides a clear description (in terms of dual spaces) of the observation operators for which a posterior can be defined and third in integrates well with excursion set estimation. 
Readers interessted in the technical details are refered to the results in \citep{travelletti_technical} upon which we build our implicit representation framework.\\

To demonstrate our method, we apply it to the Stromboli inverse problem described above. In this context, we show how it allows computing the posterior at high resolution, how hyperparameters of the prior can be trained and how we can sample from the posterior on a large grid. Finally, we illustrate how our method may be applied to a state-of-the-art sequential experimental design task for excursion set recovery. Sequential design criterion for excursion sets have gained a lot of attention recently \citep{azzimonti_adaptive}, but, to the best of our knowledge, this is the first time that sequential experimental design for set estimation is considered in the setting of Bayesian inverse problems.
\newpage

\section{Computing the Posterior under Operator Observations: Bottlenecks and Solutions}\label{sec:compute_posterior}
In this section, we focus on the challenges that arise when using Gaussian process priors to 
solve Bayesian inverse problems with linear operators observations and propose solutions 
to overcome these. Here an inverse problem is the task of recovering some unknown function $\unknownfunc \in C(D)$ from observation of a bounded linear operator $G:C(D)\rightarrow \mathbb{R}^{\datdim}$ applied to $\unknownfunc$. To solve the problem within a Bayesian framework, one puts a Gaussian process prior on $D$ and uses the conditional law of the process, conditional on the data, to approximate the unknown $\unknownfunc$.\\

Even if one can formulate the problem in an infinite dimensional setting \citep{travelletti_technical} and discretization should take place as late as possible \citep{stuart_2010}, when solving inverse problems in practice there is always some form of discretization involved, be it through quadrature methods \citep{hansen_discrete} or through basis expansion \citep{WAGNER2021110141}. It turns out that, regardless of the type of discretization used, one quickly encounters computational bottlenecks arising from memory limitations when trying to scale inversion to real-world problems. 
We here focus on inverse problems discretized on a grid, but stress that the computational difficulties described next also plague spectral approaches.\\

Let $\bm{W}=\left(w_1, ..., w_r\right)\in D^r$ be a given set of discretization points, we consider observation operators that (after discretization) may be written as linear combinations of Dirac delta functionals
\begin{equation}\label{eq:discretized_fwd}
    G:C(D)\rightarrow \mathbb{R}^{\datdim},~ G=\left(\sum_{j=1}^r g_{ij}\delta_{w_j}\right)_i,~i=1,..., \datdim,
\end{equation}
with arbitrary coefficients $g_{ij}\in\mathbb{R}$. Using \cite[Corollary 5]{travelletti_technical} we can compute the conditional law of a GP $\gp\sim Gp(m, k)$ on $D$ conditionally on the data 
$\bm{Y} = \underbar{G} Z_{\bm{W}} + \bm{\epsilon}$, where $\bm{\epsilon}\sim\mathcal{N}\left(\bm{0}, \tau^2\bm{I}_{\datdim}\right)$ is some observational noise. The conditional mean and covariance are then given by:
\begin{align}
    \tilde{m}_{\bm{X}} &=
    m_{\bm{X}}
    + K_{\bm{X} \bm{W}} \underbar{G}^T
    \left(\underbar{G} K_{\bm{W} \bm{W}} \underbar{G}^T + \tau^2 \bm{I}_{\datdim}\right)^{-1}
    \left(\bm{y} - \underbar{G} m_{\bm{W}}\right)\label{eq:cond_mean_matrix},\\
    \tilde{K}_{\predpts\predptsprime} &= K_{\predpts\predptsprime} 
    - K_{\predpts\bm{W}}\underbar{G}^T 
     \left(\underbar{G} K_{\bm{W} \bm{W}} \underbar{G}^T + \tau^2 \bm{I}_{\datdim}\right)^{-1} 
     K_{\bm{W}\predptsprime}.\label{eq:cond_cov_matrix}
\end{align}

\textbf{Notation:} When working with discrete operators as in \cref{eq:discretized_fwd} it is more convenient to use matrices. Hence we will use $\underbar{G}$ to denote the $\datdim \times r$ matrix with elements $g_{ij}$. Relation \cref{eq:discretized_fwd} will then be written compactly as $\rho\mapsto \underbar{G}\rho_{\bm{W}}$. Similarly, given a Gaussian process $Z\sim GP(m, k)$ on $D\subset \mathbb{R}^d$ and another set of points $\bm{X}=\left(x_1, ...,
x_m\right)^T\in D^m$, we will
use $K_{\bm{X}\bm{W}}$ to denote the $m\times r$ matrix obtained by evaluating 
the covariance function at all couples of points  $K_{ij}=k(x_i, w_j)$.
In a similar fashion, let $Z_{\bm{X}}\in\mathbb{R}^m$ denote the vector obtained by concatenating the values of the field at the different points.
From now on, boldface letters will be used to denote concatenated
quantities (usually datasets). The identity matrix will be denoted by $\id$, the dimension
being infered from the context.\\

Even if \cref{eq:cond_mean_matrix,eq:cond_cov_matrix} only involve basic matrix operations, their computational cost depends heavily on the number of prediction points $\bm{X}=\left(x_1, ..., x_m\right)$ and on the number of discretization points $\bm{W}=\left(w_1, ..., w_r\right)$, making their application to real-world inverse problems a non-trivial task. Indeed, when both $m$ and $r$ are big, there are two main difficulties that hamper the computation of the conditional distribution:

\begin{itemize}
        \item the $\dimfwd \times \dimfwd$ matrix $K_{\ptsfwd\ptsfwd}$ may never be
        built in memory due to its size, and
    \item the $\dimpred\times\dimpred$ posterior covariance $K_{\bm{X}\bm{X}}$ may be too large to
        store.
\end{itemize}

The first of these difficulties can be solved by performing the product
$K_{\ptsfwd\ptsfwd}\fwd^T$ in chunks, as described in \Cref{sec:chunking}. 
The second one only becomes of particular interest in sequential data
assimilation settings as considered in \Cref{sec:seq_assimilation}. In
\Cref{sec:implicit} we will introduce an (almost) matrix-free implicit representation of the
posterior covariance matrix enabling us to overcome both these memory
bottlenecks.

\subsection{Sequential Data Assimilation in Bayesian Inverse Problems and Update Formulae}\label{sec:seq_assimilation}
We now consider a sequential data assimilation problem where data is made available in
discrete stages.
At each stage, a set of
observations described by a (discretized) operator $\matrixFwdI$ is made and one observes a 
realization $\datvalsI$ of 
\begin{equation}\label{eq:seq_data_model}
    \randdatvalsI = \matrixFwdI \gp[\ptsfwdI] + \noise_{\somestage},
\end{equation}
where $\ptsfwdI$ is some set of points in $D$. 
Then, the posterior mean and covariance after each stage of observation may be
obtained by performing a low rank update of their counterparts at the previous
stage. Indeed, \cite[Corollary 6]{travelletti_technical} provides an extension of \citet{update_chevalier,update_emery,update_gao,update_barnes} to linear
operator observations, and gives:

\begin{theorem}\label{th:seq_update}
    Let $\gp\sim Gp(m, k)$ and 
    let $\muN$ and $\covN$ denote the conditional mean and covariance
    function conditional on the data $\lbrace \randdatvalsI = \datvalsI:i=1,
    ...,\stage\rbrace$ with $\randdatvalsI$ defined as in
    \cref{eq:seq_data_model}, where $\stage\geq 1$ and 
    $\meanPrior$ and $\covPrior$ are used to denote the prior mean and
    covariance.
    Then:
    \small
    \begin{align*}
        \muN_{\predpts} &= \muNprev_{\predpts} +
        \currweights{\predpts}^T\left(\datvalsN - \matrixFwdN~ \muNprev_{\ptsfwdN}\right),\\
        \covN_{\predpts\predptsprime} &= \covNprev_{\predpts\predptsprime} -
        \currweights{\predpts}^T
        \invopUpdt \currweights{\predptsprime},
    \end{align*}
with $\currweights{\predpts}$, $\invopUpdt$ defined as:
    \begin{align*}
        \currweights{\predpts} &=
        \invopUpdt^{-1} \matrixFwdN \covNprev_{\ptsfwdN \predpts},\\
        \invopUpdt &= \invopNexpr.
    \end{align*}
    \normalsize
\end{theorem}

At each stage $\somestage$, these formulae require computation of the $\datdimN \times
\dimpred$ matrix $\currweightsI{\predpts}$, which involves a
$\datdimN \times \datdimN$ matrix inversion, where
$\datdimN$ is the dimension of the operator $\matrixFwdN$ describing the
current dataset to be included. This allows computational savings by reusing already computed quantities, avoiding inverting the full dataset at each stage, which would require a 
$\datdim_{tot}^2$ matrix inversion, where $\datdim_{tot}=\sum_{\somestage=1}^{\stage} \datdimI$.\\

In order for these update equations to bring computational savings, one has to be able to store
the past covariances $\prevcov[\ptsfwdN\ptsfwdN]$ \citep{foxy}. This makes their application to large-scale sequential Bayesian inverse problems difficult, since the covariance matrix on the
full discretization may become too large for storage above a certain number of discretization points. 
The next section presents our main contributions to overcome this limitation.
They rely on an implicit representation of the posterior covariance that allows 
the computational savings offered by the kriging update formulae to be brought 
to bear on large scale inverse problems.

\subsection{Implicit Representation of Posterior
Covariances for Sequential Data Assimilation in Large-Scale Bayesian Inverse Problems}\label{sec:implicit}
We consider the same sequential data assimilation setup as in the previous section, 
and for the sake of simplicity we assume that $\bm{W}_1,..., \bm{W}_{\stage}=\bm{X}$
and use the lighter notation $\muI:=\muI_{\predpts}$ and
$\covI:=\covI_{\predpts\predpts}$. The setting we are interested in here is the one where $\predpts$ is so large that 
the covariance matrix gets bigger than the
available computer memory.\\

Our key insight is that instead of building the
full posterior covariance $\covN$ at each stage $\stage$, one can 
just maintain a routine that computes the product of the current posterior
covariance with any other low rank matrix.
More precisely, at each stage $\stage$, we provide a routine
$\textrm{CovMul}_{\stage}$ (\cref{alg:mult}), that allows to compute the product of the
current covariance  matrix with any \textit{thin} matrix $A \in\mathbb{R}^{\dimpred
\times \dimA}, ~\dimA \ll \dimpred$:
\[
    \textit{CovMul}_{\stage}: A \mapsto K^{(\somestage)}A,
\]
where \textit{thin} is to be understood as small enough so that the result of
the multiplication can fit in memory.\\

This representation of the posterior covariance was inspired by the covariance operator of Gaussian measures. Indeed, if we denote by $C_{\mu^{(\stage)}}$ the  covariance operator of the Gaussian measure associated to the posterior distribution of the GP at stage $\stage$, then 
\[
    \left(K^{(\stage)}A\right)_{\sumIndexI \sumIndexJ}=\sum_{\sumIndexK=1}^{\dimpred} \left\langle C_{\mu^{(\stage)}}\delta_{x_{\sumIndexI}}, \delta_{x_{\sumIndexK}} \right\rangle A_{\sumIndexK \sumIndexJ}.
\]
Hence, the procedure $\textrm{CovMul}_{\stage}$ may be thought of as computing the action of the covariance operator of the Gaussian measure associated to the posterior on the Dirac delta functionals at the discretization points.

This motivates us to think in terms of an \textit{updatable covariance} object, 
where the inclusion of new observations (the updating) amounts to redefining 
a right-multiplication routine.
It turns out that by grouping terms appropriately in \cref{th:seq_update} such a routine may be defined by only storing low rank
matrices at each data acquisition stage.
~\\
\begin{lemma}\label{th:sequential}
For any $\stage \in \mathbb{N}$ and any $\dimpred \times \dimA$ matrix $A$:
\[
    \covN A = K^{(0)} A - \sum_{\somestage=1}^{\stage}
    \pushfwdI \invopI^{-1} \pushfwdI^T A,
\]
with intermediate matrices $\pushfwdI$ and 
$\invopI^{-1}$ defined as:
\begin{align*}
    \pushfwdI :&= \covIprev \matrixFwdI^T,\\
    \invopI^{-1} :&= \left( \matrixFwdI \covIprev \matrixFwdI^T + \datvar \id \right)^{-1}.
\end{align*}
\end{lemma}
Hence, in order to compute products with the posterior covariance at
stage $\stage$, one only has to store $\stage$ matrices $\pushfwdI$, each
of size $\dimpred\times \datdimI$ and $\stage$ matrices
$\invopI^{-1}$ of
size $\datdimI \times \datdimI$, where $\datdimI$ is the number of observations
made at stage $\somestage$ (i.e. the number of lines in $\matrixFwdI$).
In turn, each of these objects is defined by multiplications with the
covariance matrix at previous stages, so that one may recursively update the
multiplication procedure $\textrm{CovMul}_{\stage}$. 
\Cref{alg:mult,,alg:update_interm,alg:seq_cond_mean} may be used for 
multiplication with the current covariance matrix, update of the representation
and update of the posterior mean.

\begin{algorithm}[h]
    \caption{\textit{Covariance Right Multiplication Procedure $\textrm{CovMul}_{\stage}$}}\label{alg:mult}
    \begin{algorithmic}
\Require
\State Precomputed matrices $\pushfwdI,~\invopI^{-1}$, $\somestage=1,\dotsc,\stage$.
\State Prior multiplication routine $\textrm{CovMul}_0$.
\State Input matrix $A$.
\Ensure $\covN A$.
\Procedure{$\textrm{CovMul}_n$}{A}
\State Compute $\covPrior A=\textrm{CovMul}_0(A)$.
\State \textbf{Return} $\covPrior A - \sum_{i=1}^n \pushfwdI \invopI^{-1} \pushfwdI^T A$.
\EndProcedure
    \end{algorithmic}
\end{algorithm}

\begin{algorithm}[h]
    \caption{\textit{Updating intermediate quantities at conditioning stage $\stage$}}
    \label{alg:update_interm}
    \begin{algorithmic}
\Require
\State Last multiplication routine $\textrm{CovMul}_{\stage-1}$.
\State Measurement matrix $\matrixFwdN$, noise variance $\datvar$.
\Ensure Step $\stage$ intermediate matrices $\pushfwdN$ and $\invopN$
\Procedure{$\textrm{Update}_{\stage}$}{}
\State Compute $\pushfwdN = \textrm{CovMul}_{\stage-1} \matrixFwdN^T$.
\State Compute $\invopN^{-1}=\left( \matrixFwdN \pushfwdN +
\datvar \id\right)^{-1}$.
\EndProcedure
    \end{algorithmic}
\end{algorithm}

\begin{algorithm}[h!]
    \caption{\textit{Computation of conditional mean at step $n$}}
    \label{alg:seq_cond_mean}
    \begin{algorithmic}
\Require
\State Previous conditional mean $\muNprev$.
\State Current data $\datvalsN$ and forward $\matrixFwdN$.
\State Intermediate matrices $\pushfwdN$ and $\invopN^{-1}$.
\Ensure Step $n$ conditional mean $\muN$.
\Procedure{$\textrm{MeanUpdate}_{\stage}$}{}
\State \textbf{Return} $\muNprev + \pushfwdN \invopN^{-1} \left(\datvalsN -
\matrixFwdN \muNprev \right)$.
\EndProcedure
    \end{algorithmic}
\end{algorithm}

\subsection{Prior Covariance Multiplication Routine and
Chunking}\label{sec:chunking}
To use \Cref{alg:mult}, one should be able to compute
products with the prior covariance matrix $\covPrior$. This may be achieved by
chunking the set of grid points into $\nchunks$ subsets
$\predpts=\left(\predpts_1, ..., \predpts_{\nchunks}\right)$, where each
$\predpts_i$ contains a subset of the points. Without loss of generality, we
assume all subsets to have the same size $\dimpred_c$. We may then write the product as
\begin{equation*}
    \covPrior_{\predpts\predpts}A = \left(\covPrior_{\predpts_1\predpts}A,
    \dotsc, \covPrior_{\predpts_{\nchunks}\predpts}A\right)^T.\\
\end{equation*}
Each of the subproducts may then be performed separately and the results
gathered together at the end. The individual products then involve matrices of
size $\dimpred_c\times \dimpred$ and $\dimpred\times\dimA$. One can then choose
the number of chunks so that these matrices can fit in memory. 
Each block $\covPrior_{\predpts_i\predpts}$ may be built on-demand provided 
$\covPrior_{\predpts\predpts}$ is defined through a given covariance function.\\
~\\
This ability of the prior covariance to be built quickly on-demand 
is key to our
method. The fact that the prior covariance matrix does not need to be stored 
allows us to handle larger-than-memory posterior covariances by
expressing products with it as a multiplication with the prior and a sum of
multiplications with lower rank matrices.\\
~\\
\begin{remark}[Choice of Chunk Size]
Thanks to chunking, the product may be computed in
parallel, allowing for significant performance improvements 
in the presence of multiple computing devices (CPUs, GPUs,
...). In that case, the chunk size should be chosen as large as possible to 
limit data transfers, but small enough so that the subproducts may fit on the
devices.
\end{remark}

\subsection{Computational Cost and Comparison to Non-Sequential Inversion}\label{sec:cost_analysis}
For the sake of comparison, assume that all \textit{n} datasets have the same size
$\dimChunk$ and let $\datdim = \stage \dimChunk$ denote the total data size. The cost of 
computing products with the current posterior
covariance matrix at some intermediate stage is given by:

\begin{lemma}[Multiplication Cost]\label{lemma:mult}
Let $A$ be an $\dimpred\times\dimA$ matrix. Then, the cost of computing 
$K_{\stage}A$ at some stage $\stage$ using \cref{alg:mult,alg:update_interm} is 
$\mathcal{O}\left(\dimpred^2\dimA + \stage(\dimpred \dimChunk
\dimA + \dimChunk^2\dimA)\right)$.
\end{lemma}
Using this recursively, we can then compute the cost of creating the implicit representation
of the posterior covariance matrix at stage $\stage$:
\begin{lemma}[Implicit Representation Cost]\label{lemma:repr_cost}
	To leading order in $\dimpred$ and
    $\datdim$, the cost of defining $\textrm{CovMul}_{\stage}$ is 
$\mathcal{O}\left(\dimpred^2\datdim + \dimpred \datdim^2 +
    \datdim^2\dimChunk\right)$. This is also the cost of computing $\muN$.
\end{lemma}

\begin{figure}[ht]
\centering
  \centering
  \includegraphics[width=0.85\linewidth,height=0.6\linewidth]{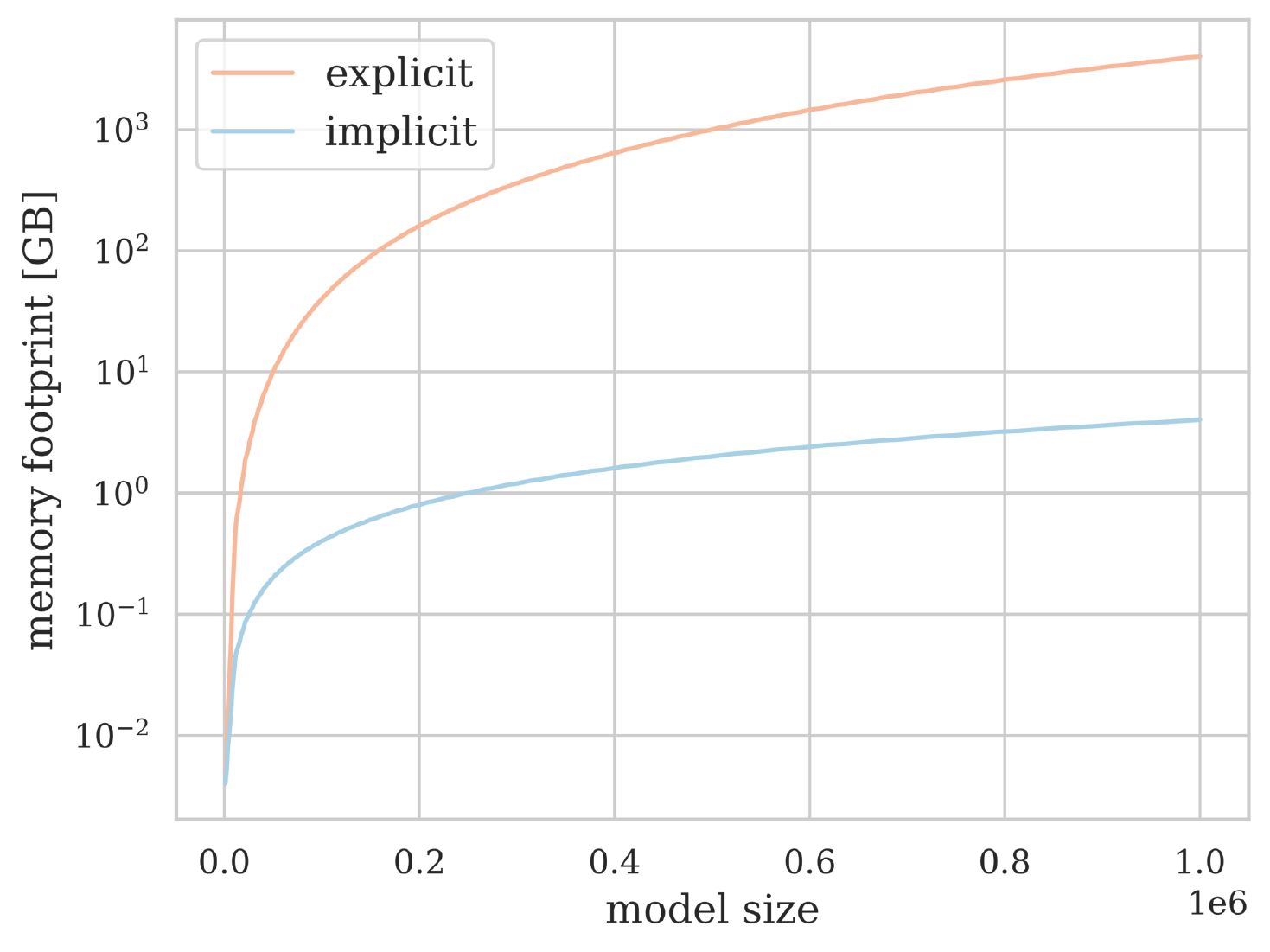}
  \caption{Memory footprint of the posterior covariance matrix as a function of
  discretization size for explicit and implicit representation.}\label{fig:memory}
\end{figure}

This can then be compared with a non-sequential approach where all datasets
would be concatenated into a single dataset of dimension $\datdim$. More
precisely, define the $\datdim\times \dimgrid$ matrix $\underbar{\fwd}$ 
and the $\datdim$-dimensional vector $\bm{y}$ as the concatenations 
of all the measurements and data vectors into a single operator, respectively
vector. Then computing the posterior mean using \cref{eq:cond_mean_matrix} with
those new observation operators and data vector the cost is, 
to leading order in $\datdim$ and $\dimpred$:
\[
    \mathcal{O}\left(\dimpred^2 \datdim + \dimpred \datdim^2 +
        \datdim^3
    \right).
\]
In this light, we can now sum up the two main advantages of the proposed sequential
approach:
\begin{itemize}
    \item the cubic cost
    $\mathcal{O}\left(\datdim^3\right)$ arising from the
    inversion of the data covariances is decreased to
    $\mathcal{O}\left(\datdim^2\dimChunk\right)$ in the sequential approach
    \item if a new set of observations has to be included, then the direct approach
        will require the $\mathcal{O}\left(\dimpred^2 \datdim\right)$
        computation of the
product $K\underbar{\fwd}^T$, which can become prohibitively expensive when the number of
prediction points is large, whereas the sequential approach will only require a
marginal computation of $\mathcal{O}\left(\dimpred^2 \dimChunk\right)$.
\end{itemize}
Aside from the computational cost, our implicit representation also provides
significant memory savings compared to an \textit{explicit} approach where
the full posterior covariance matrix would be stored. The storage requirement
for the implicit-representation as a function of the number of discretization
points $\dimpred$ is shown in \Cref{fig:memory}.

\subsection{Toy Example: 2D Fourier Transform}\label{sec:Fourier2D}
To illustrate the various methods presented in \cref{sec:implicit}, we here apply them to a toy two-dimensional example: 
we consider a GP discretized on a large square grid $\lbrace 0, ...,M\rbrace \times \lbrace 0, ..., M\rbrace$ and try to learn it through various types of linear operator data.\\
~\\
More precisely, we will here allow two types of observations: pointwise field values and Fourier coefficient data. 
Indeed, the field values at the nodes $Z_{kl},~k,l\in\lbrace 1, ..., M\rbrace$ are entirely determined by its discrete Fourier transform (DFT):
\begin{align}
    F_{uv} = \sum_{k=1}^M\sum_{l=1}^M Z_{kl} e^{-2\pi i \left(\frac{uk}{M} + \frac{vl}{M}\right)},~ u,v\in\lbrace 1, ...,M\rbrace.
\end{align}
Note that the above may be viewed as the discretized version of some operator and thus fits our framework for sequential conditioning. One may then answer questions such as \textit{how much uncertainty is  left after the first 10 Fourier coefficient $F_{kl},~k,l=1,...,10$ have been observed?} Or \textit{which Fourier coefficient provide the most information about the GP?}
\begin{figure}[h!]
    \centering
    \subfloat[ground truth]{\includegraphics[scale=0.3]{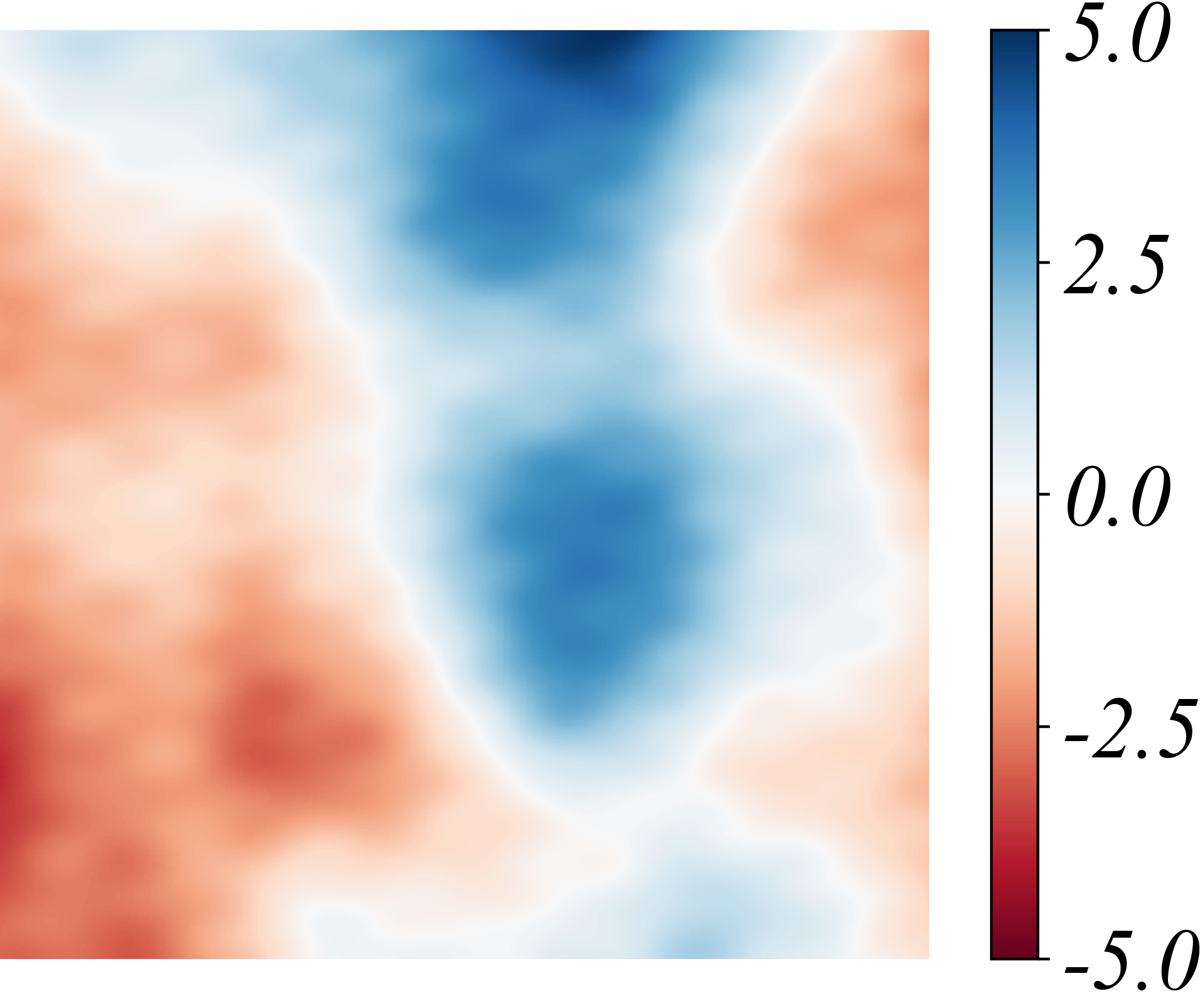}}
    \hspace{0.5em}
    \subfloat[10 Fourier]{\includegraphics[scale=0.3]{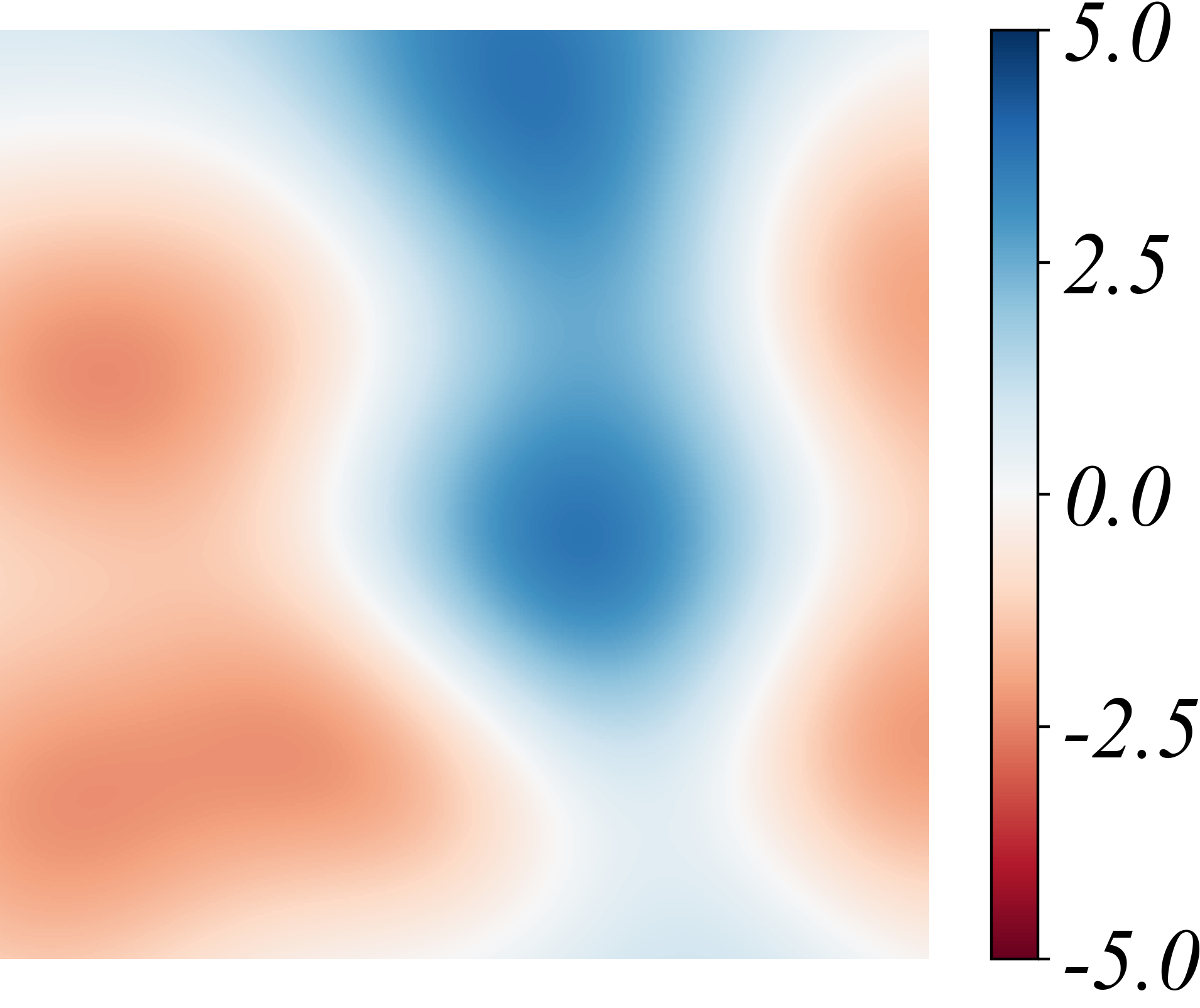}}
    \hspace{0.5em}
    \subfloat[50 Fourier]{\includegraphics[scale=0.3]{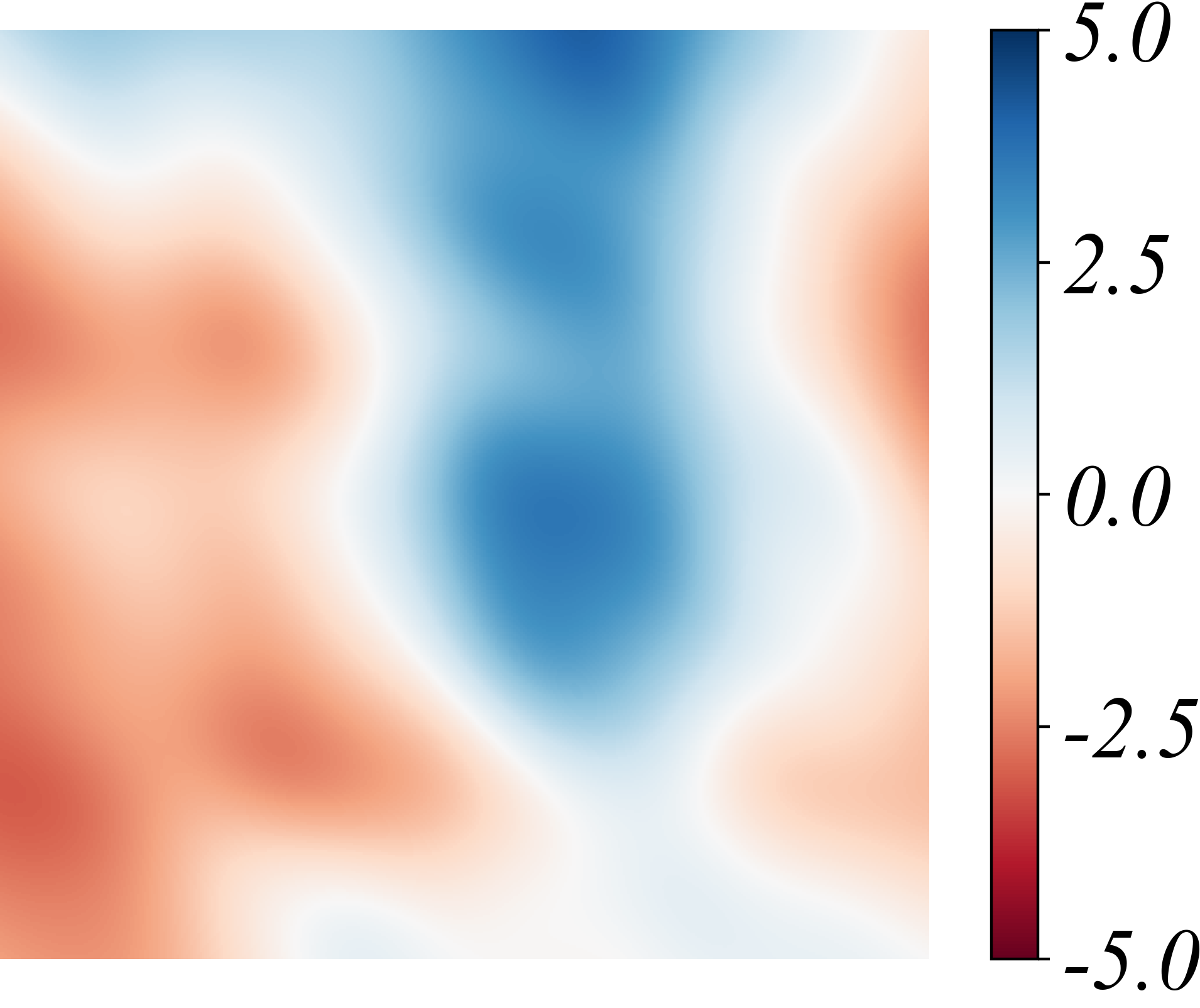}}
    \hspace{0.5em}
    \subfloat[100 Fourier]{\includegraphics[scale=0.3]{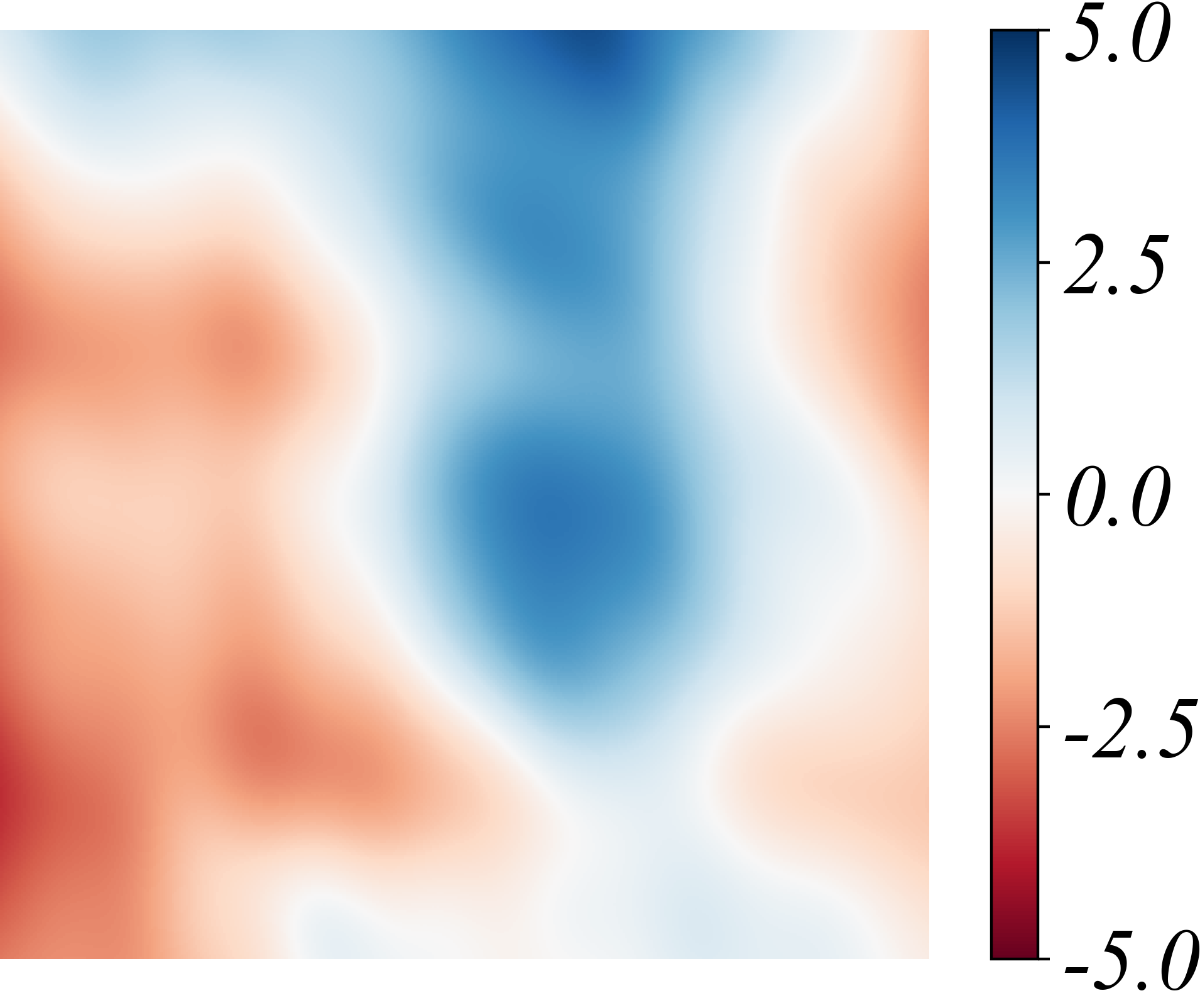}}\\
    \subfloat[ground truth]{\includegraphics[scale=0.3]{images/fourier_05/ground_truth.png}}
    \hspace{0.5em}
    \subfloat[10 pointwise]{\includegraphics[scale=0.3]{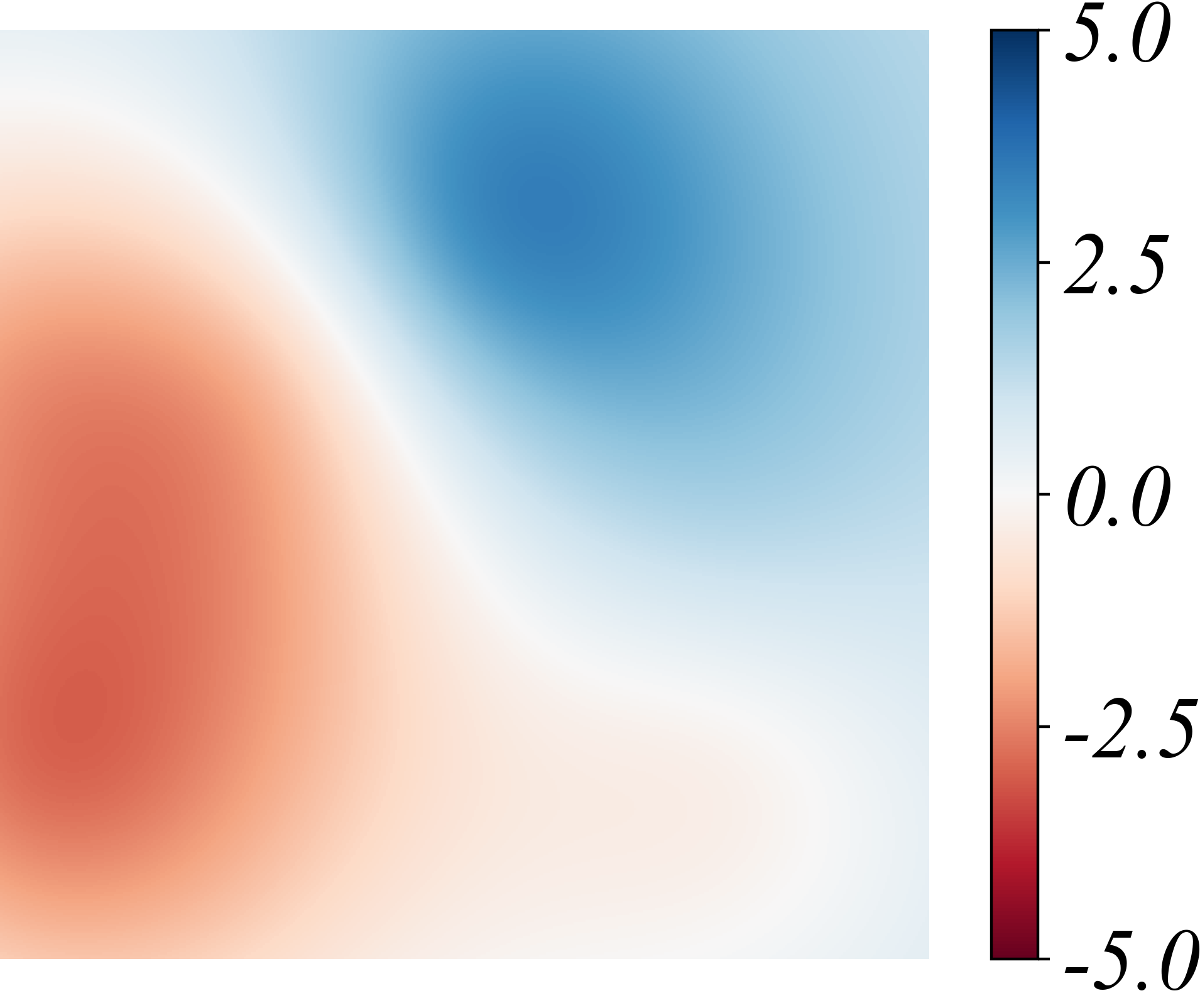}}
    \hspace{0.5em}
    \subfloat[50 pointwise]{\includegraphics[scale=0.3]{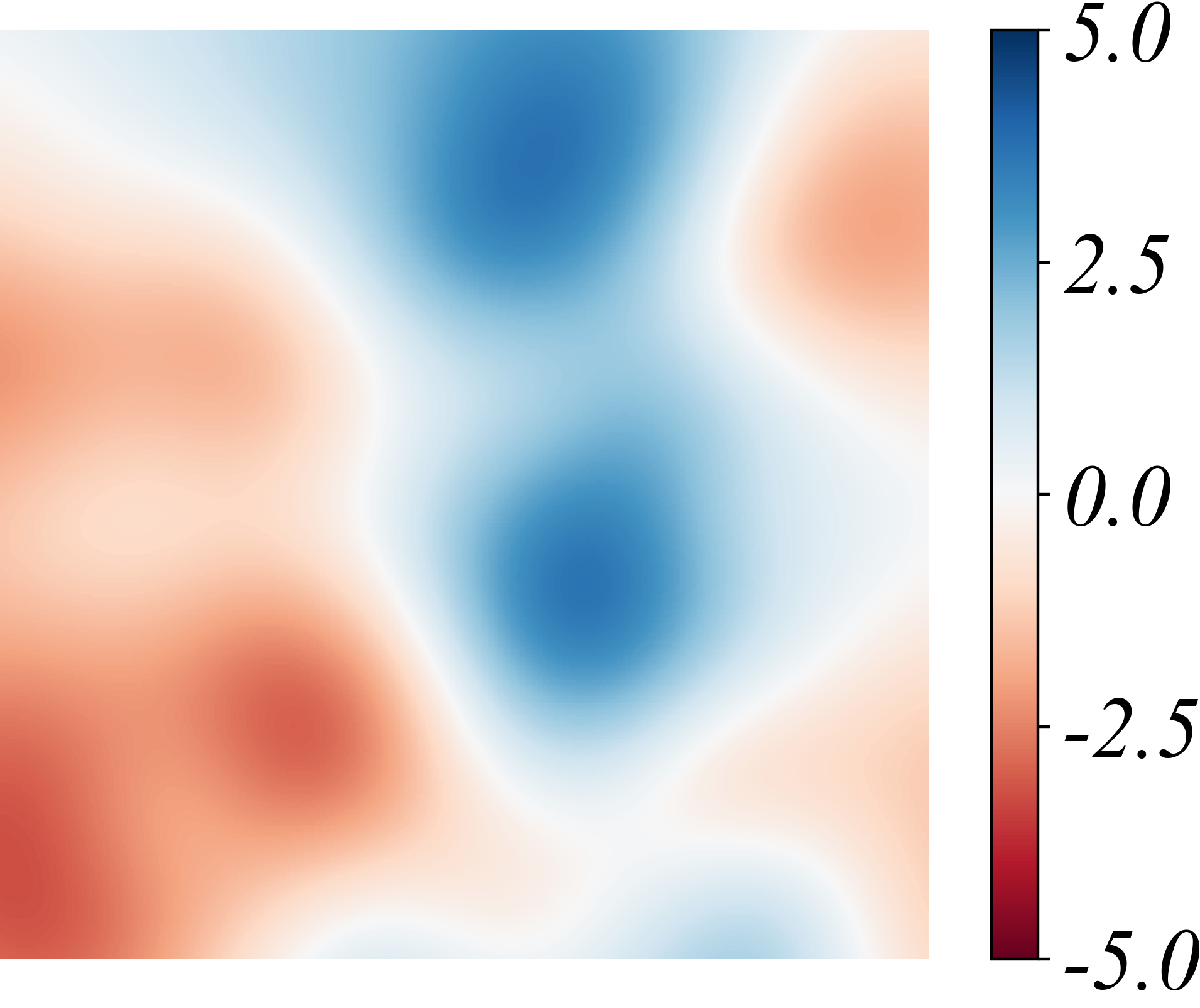}}
    \hspace{0.5em}
    \subfloat[100 pointwise]{\includegraphics[scale=0.3]{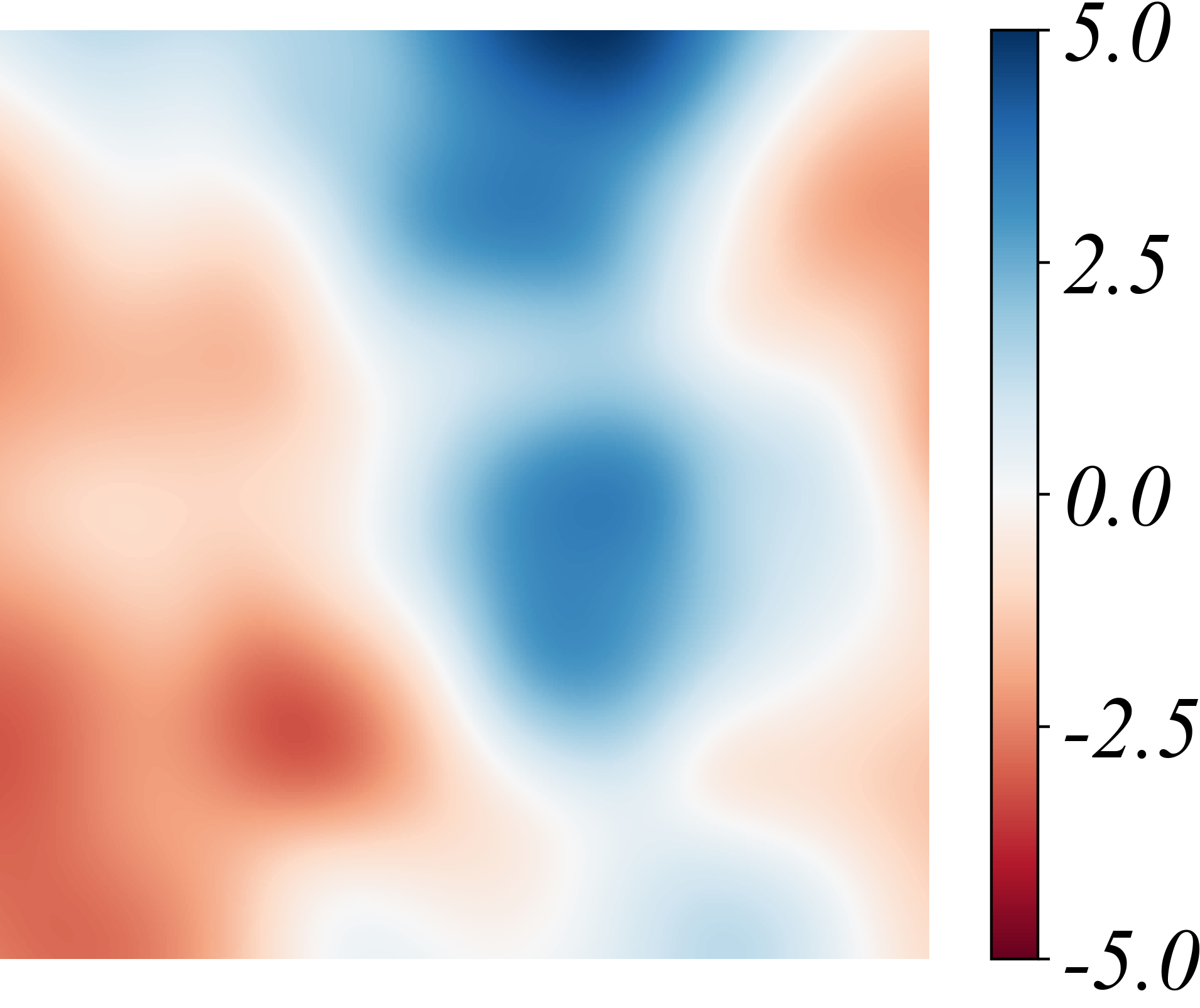}}
    \caption{Posterior mean after observation of $n=10,50,100$ Fourier coefficients (top) and after observation of $n=10, 50,100$ field values along a space-filling design (bottom). Ground truth is shown on the left. Correlation parameter (Mat\'ern 5/2) $\lambda=0.5$.}\label{fig:post_mean_fourier_05}
\end{figure}

For example, \cref{fig:post_mean_fourier_05} compares the posterior mean after observation of various Fourier coefficients, compared to observing pointwise values along a space-filling sequence. The GP model is a constant mean Mat\'ern $5/2$ random field on a square domain $\left[-1,1\right]^2$. The domain is discretized onto a $400\times 400$ grid. Note that, by nature, each Fourier coefficient involves the field values at all points of the discretization grid and thus direct computation of the posterior mean requires the full $400^2\times 400^2$ covariance matrix (which would translate to roughly 100 GB of storage). This makes this situation suitable to demonstrate the techniques presented in this section.

\begin{figure}[tbh!p]
    \centering
    \subfloat[ground truth]{\includegraphics[scale=0.3]{images/fourier_05/ground_truth.png}}
    \hspace{0.5em}
    \subfloat[10 Fourier]{\includegraphics[scale=0.3]{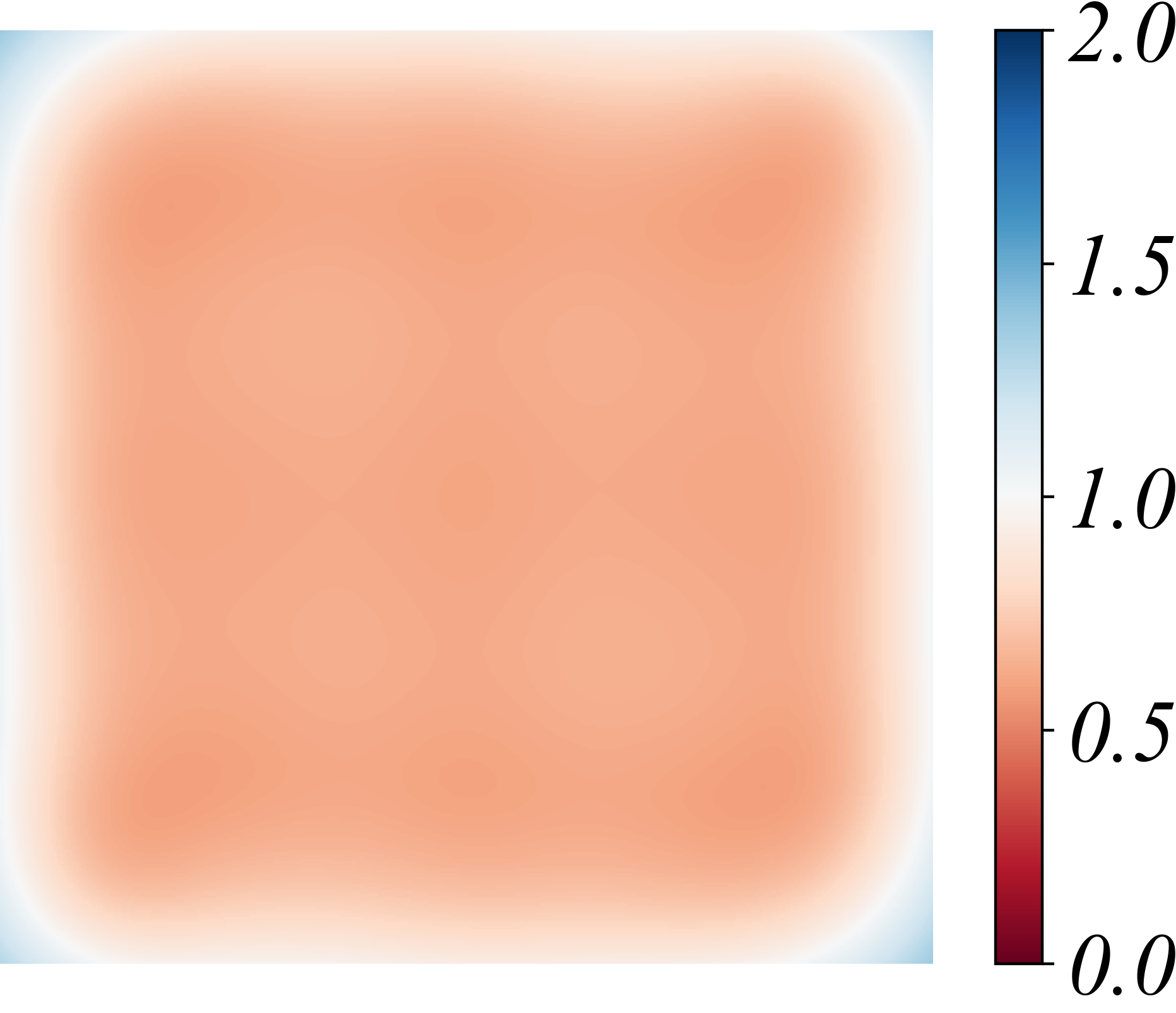}}
    \hspace{0.5em}
    \subfloat[50 Fourier]{\includegraphics[scale=0.3]{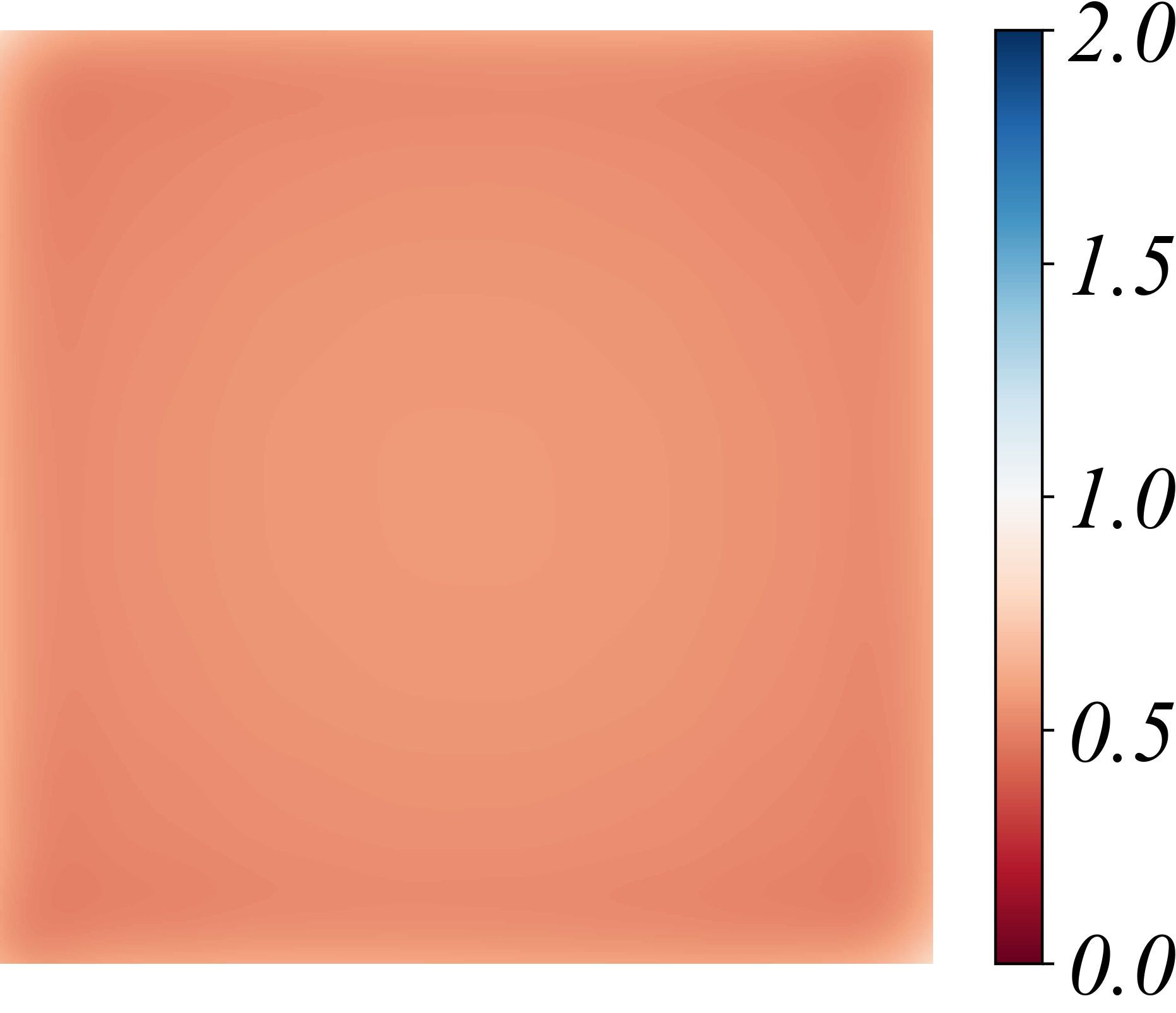}}
    \hspace{0.5em}
    \subfloat[100 Fourier]{\includegraphics[scale=0.3]{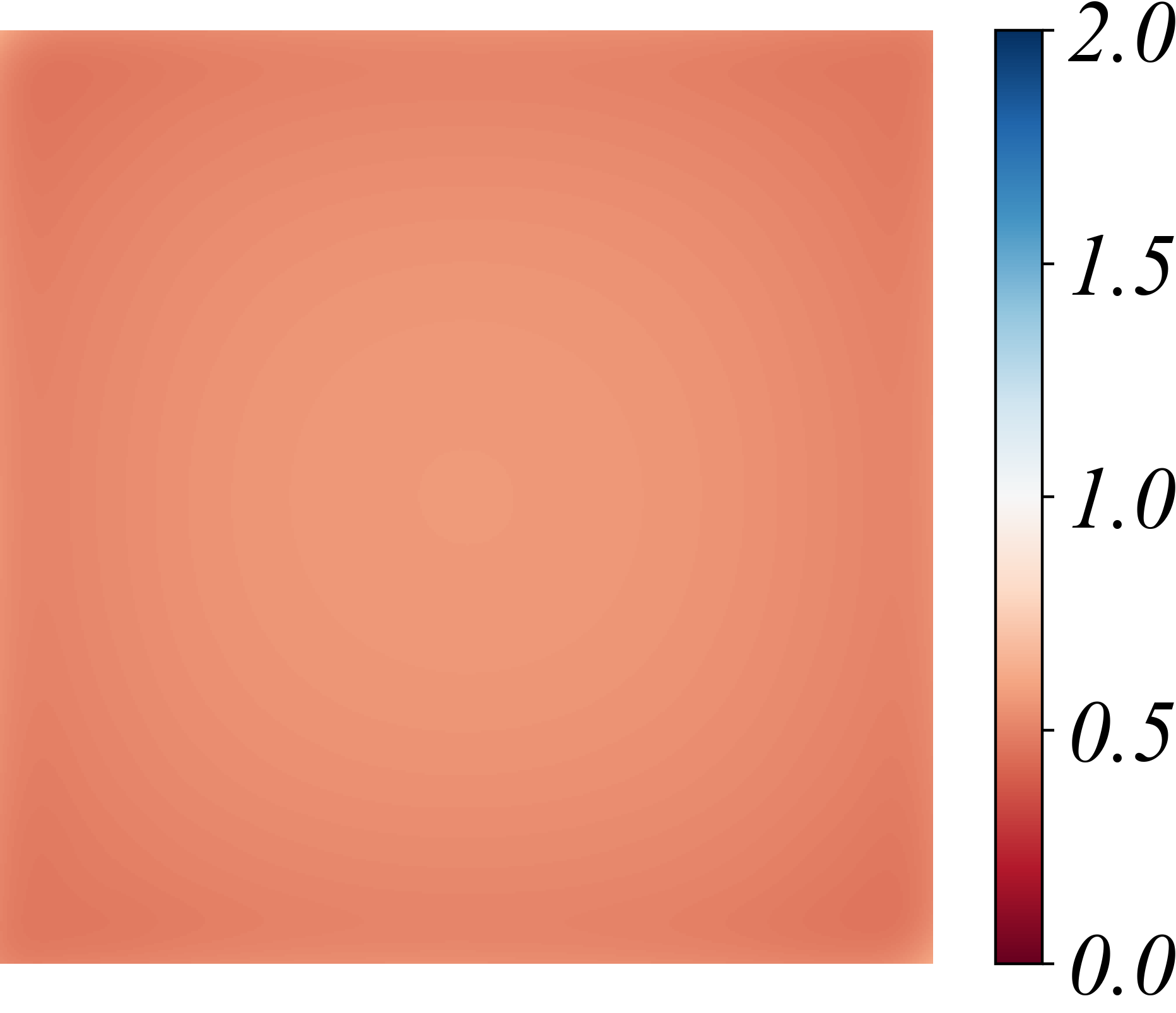}}\\
    \subfloat[ground truth]{\includegraphics[scale=0.3]{images/fourier_05/ground_truth.png}}
    \hspace{0.5em}
    \subfloat[10 pointwise]{\includegraphics[scale=0.3]{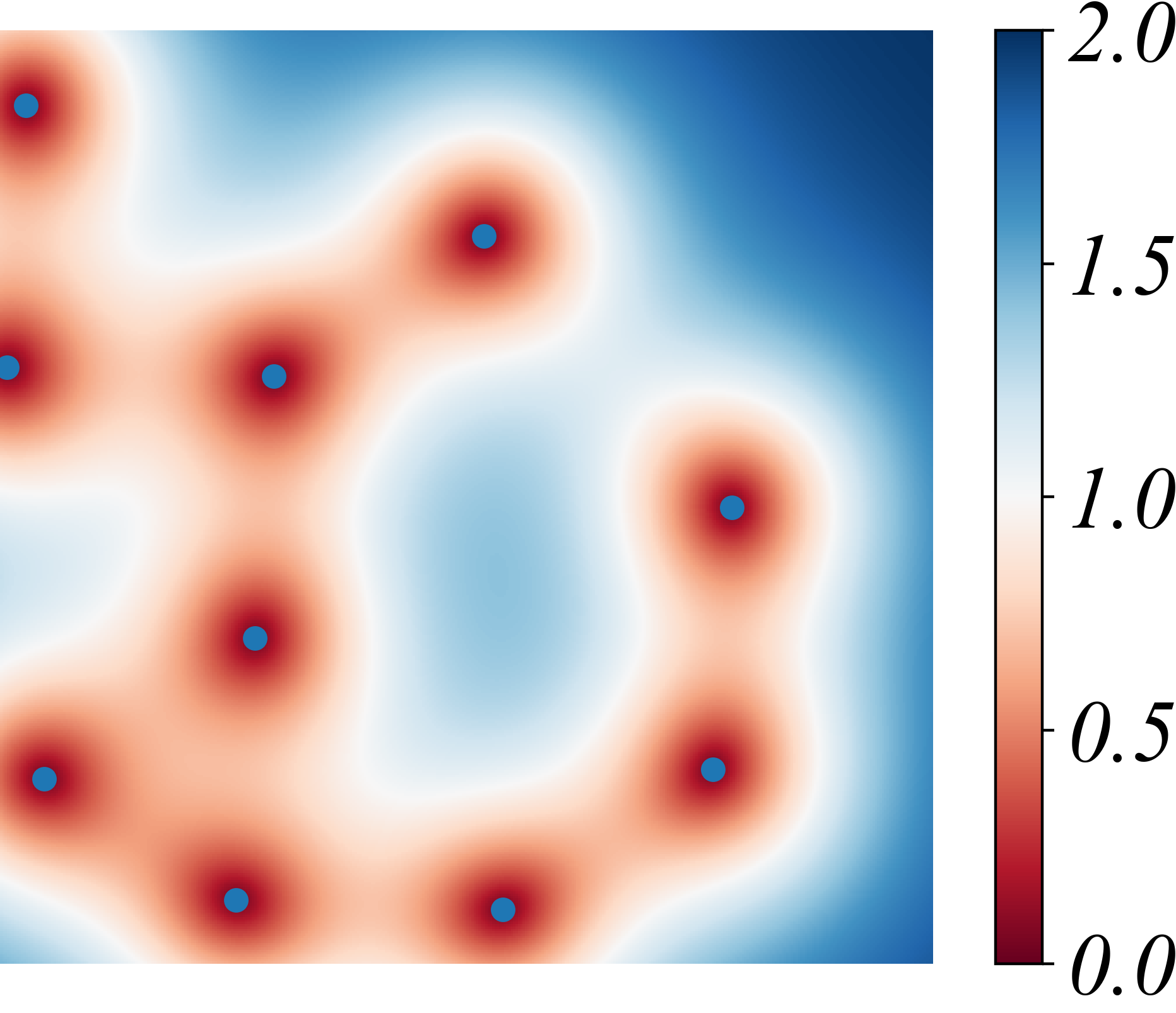}}
    \hspace{0.5em}
    \subfloat[50 pointwise]{\includegraphics[scale=0.3]{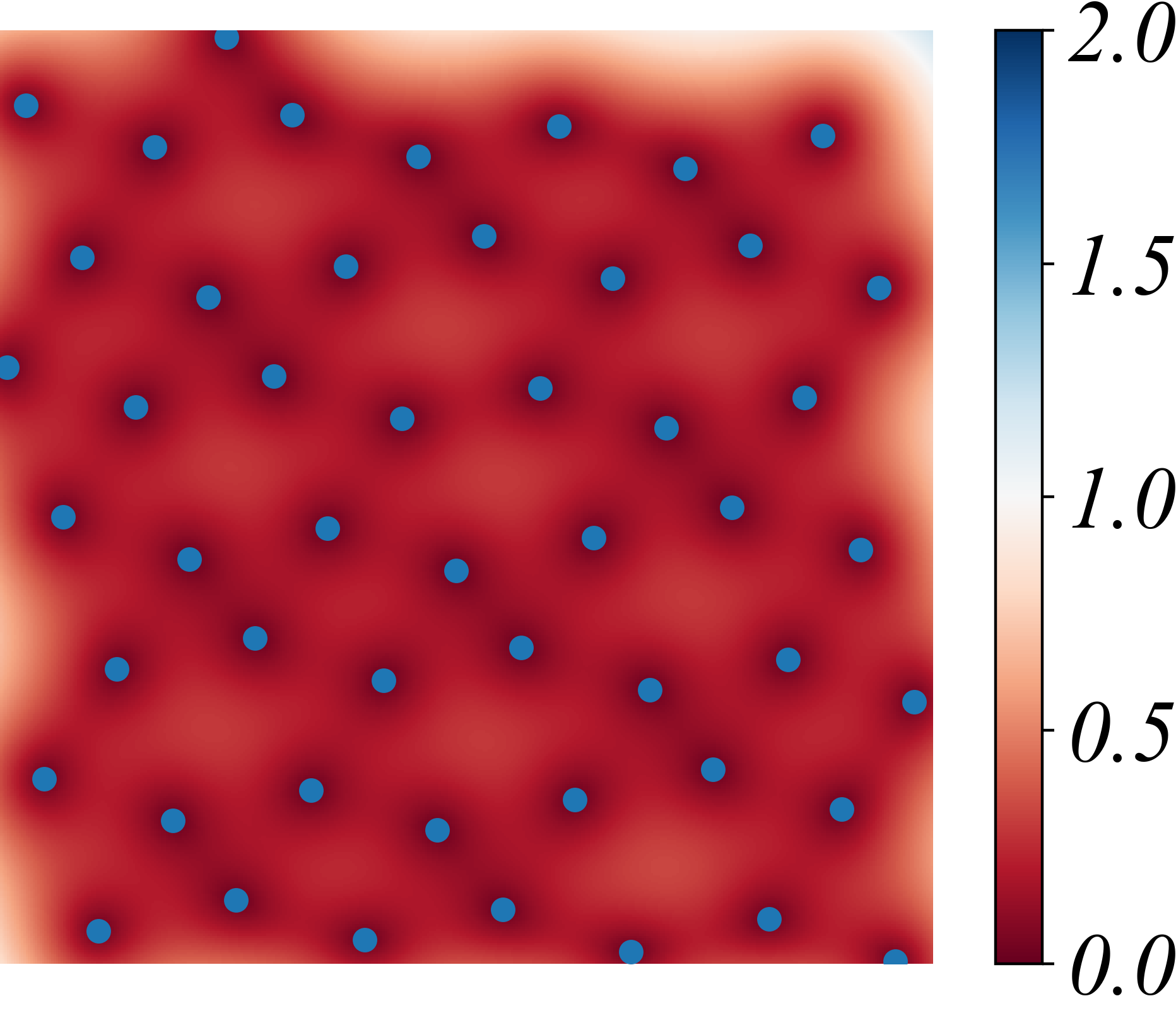}}
    \hspace{0.5em}
    \subfloat[100 pointwise]{\includegraphics[scale=0.3]{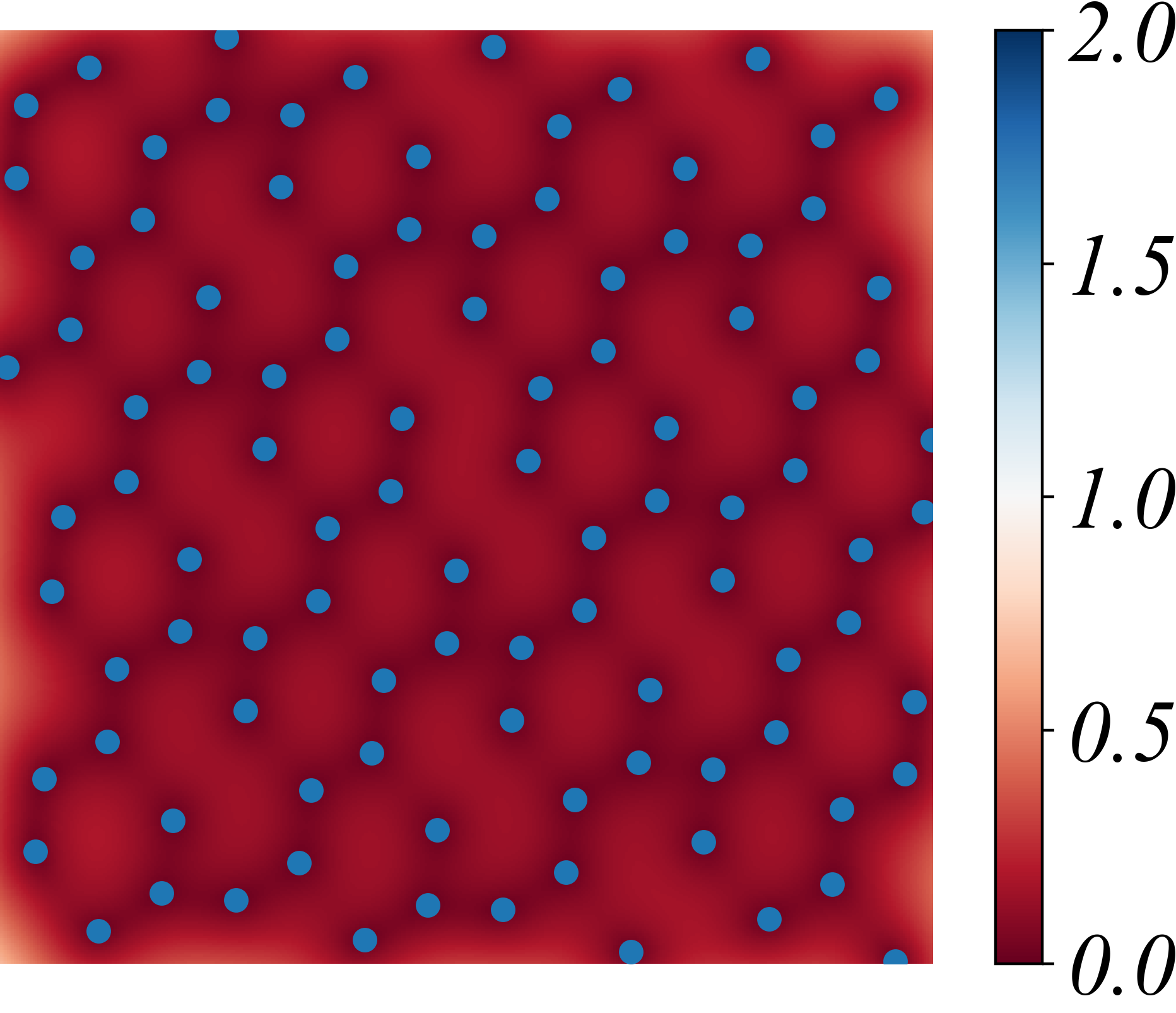}}
    \caption{Posterior standard deviation after observation of $n=10,50,100$ Fourier coefficients (top) and after observation of $n=10,50,100$ field values along a space-filling design (bottom). Ground truth is shown on the left. Correlation parameter (Mat\'ern 5/2) $\lambda=0.5$.}\label{fig:post_std_fourier_05}
\end{figure}

~\\
In this example, the Fourier coefficients are ordered by growing $l_{\infty}$ norm. One observes that Fourier coefficients provide very different information than pointwise observations and decrease uncertainty in a more spatially uniform way, as shown in \cref{fig:post_std_fourier_05}.

The extent to which Fourier coefficients provide more or less information than pointwise observations (depending on ordering) depends on the correlation length of the field's covariance function. Indeed, for long correlation lengths, the low frequency Fourier coefficients contain most of the information about the field's global behaviour, whereas for small correlation length, high frequency coefficients are needed to learn the small-scale structure. 

\section{Application: Scaling Gaussian Processes to Large-Scale Inverse Problems}\label{sec:applications}
In this section, we demonstrate how the implicit representation of the posterior covariance introduced in \cref{sec:implicit} allows scaling Gaussian processes to situations that are too large to handle using more traditional techniques. Such situations are frequently encountered in large-scale inverse problems and we will thus focus our exposition on such a problem arising in geophysics. In this setting, we demonstrate how our implicit representation allows to train prior hyperparameters in \cref{sec:hyperparams}, in \cref{sec:sampling} we demonstrate posterior sampling on large grids and finally in \cref{sec:optimal_design} address a state-of-the-art sequential experimental design problem for excursion set recovery.\\

\textbf{Example Gravimetric Inverse Problem:} We focus on the problem of reconstructing the 
mass density distribution $\density:\domain\rightarrow\mathbb{R}$ 
in some given underground domain $\domain$ from 
observations of the vertical 
component of the gravitational field at points $s_1, ..., s_{\datdim}$ on the
surface of the domain. 
Such gravimetric data are extensively used on volcanoes \citep{montesinos,represas,linde_2017} 
and can be acquired non-destrucively at comparatively low costs by 1-2 persons in 
rough terrain using gravimeters, without the need for road infrastructure or 
other installations (such as boreholes) that are often lacking on volcanoes. 
Gravimeters are sensitive to spatial variations in density, which makes them useful to understand geology, localize ancient volcano conduits and present magma chambers, and to identify regions of loose light-weight material that are prone to landslides that could in the case of volcanic islands generate tsunamis.

\begin{figure}[tbhp]
\begin{subfigure}{0.49\textwidth}
	\includegraphics[width=0.99\linewidth]{images/overview/density_3d_crop-eps-converted-to.pdf}
	\caption{}
\end{subfigure}
\begin{subfigure}{0.49\textwidth}
	\includegraphics[width=0.99\linewidth]{images/overview/data_vals_sphere_warm_crop-eps-converted-to.pdf}
	\caption{}
\end{subfigure}

\begin{subfigure}{0.49\textwidth}
	\includegraphics[width=0.99\linewidth]{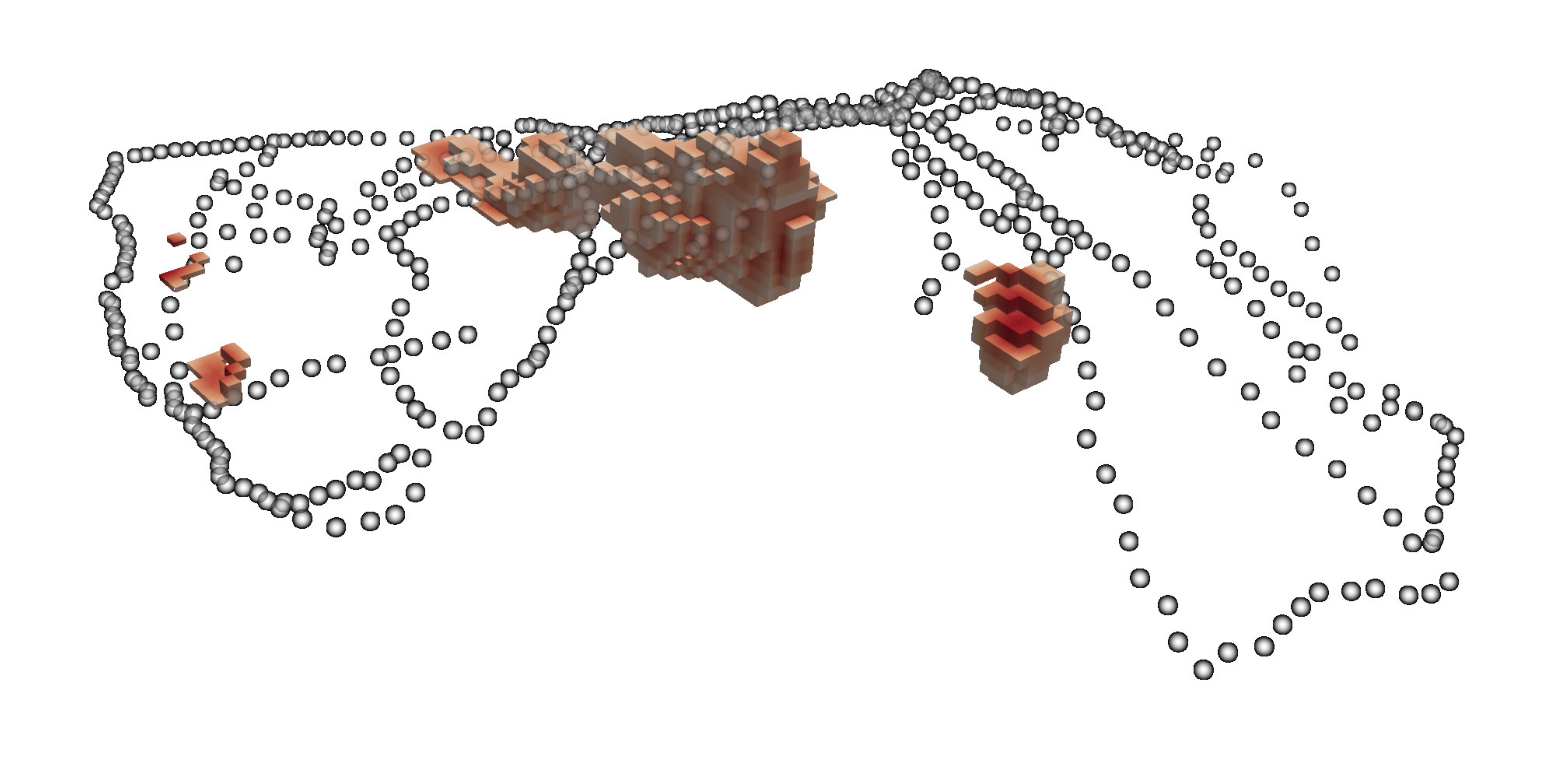}
	\caption{}
\end{subfigure}
\begin{subfigure}{0.49\textwidth}
	\includegraphics[width=0.99\linewidth]{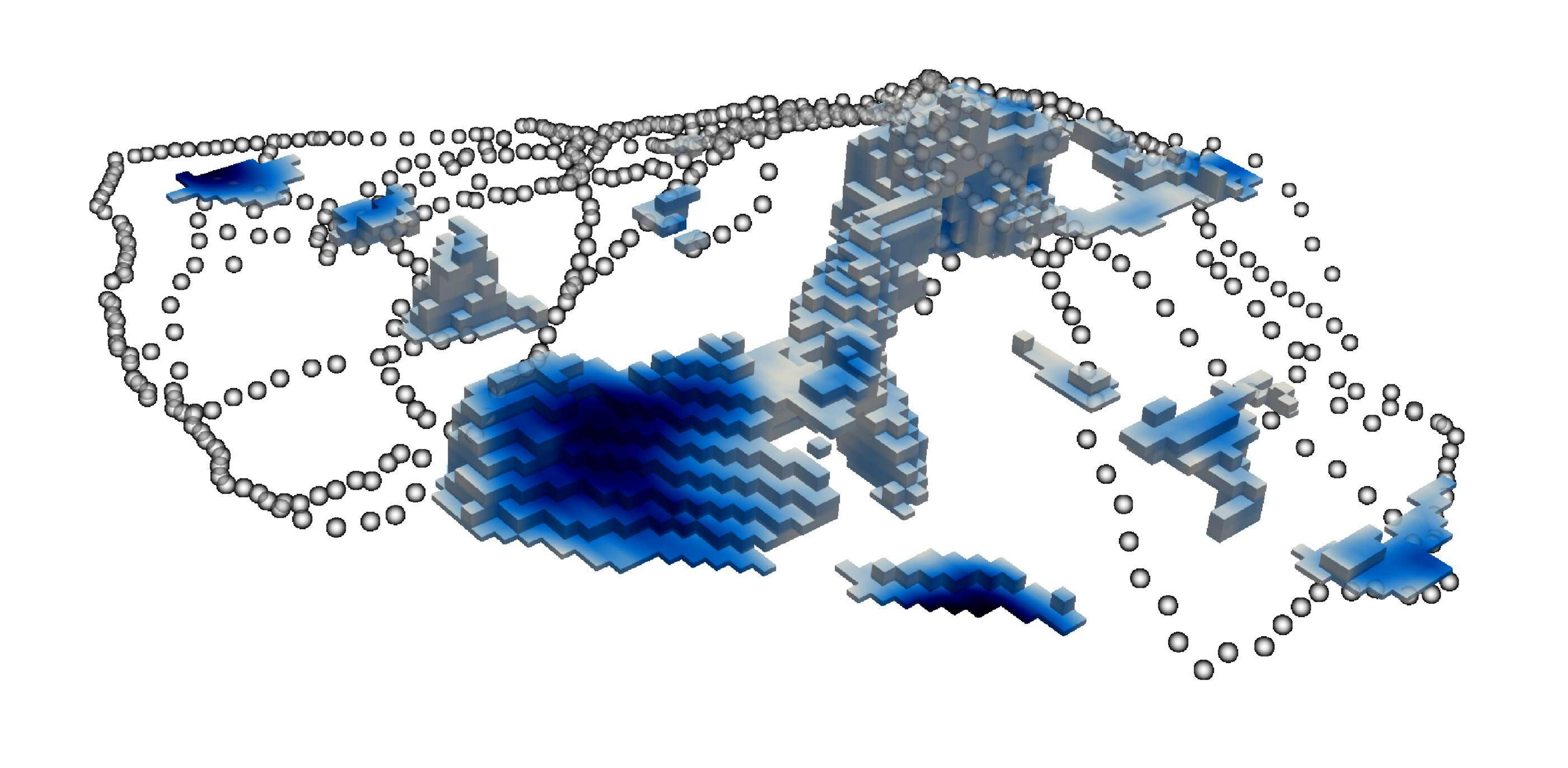}
	\caption{}
\end{subfigure}
\caption{Problem overview: (a) underground mass density (realisation from GP prior), (b) vertical intensity of the generated gravity field at selected locations, (c) high density regions and (d) low density regions. Thresholds and colorscales were chosen arbitrarily.}
\label{fig:problem_overview}
\end{figure}

As an example dataset we use gravimetric data gathered on the surface of the Strombli volcano during a field campaign in 2012 \citep{linde}. In \cref{sec:optimal_design} we will also consider the problem of recovering high (or low) density 
regions inside the volcano. \cref{fig:problem_overview} displays the main components of the problem.

The observation operator describing gravity measurements is an integral one (see \cref{sec:grav_op}), which, after discretization, fits the Bayesian inversion framework of 
\cref{sec:compute_posterior}. Indeed, this kind of problems is usually discretized on a finite grid of points $\bm{X}=\left(x_1, \dotsc, x_m\right)$, hence the available data is of the form
\begin{equation}\label{eq:data_model_grav}
    \bm{Y} = \underbar{G}\rho_{\predpts} + \bm{\epsilon},
\end{equation}
where the $\datdim \times \dimpred$ matrix $\underbar{G}$ represents the discretized version of the observation operator for the gravity field at $s_1, \dotsc, s_{\datdim}$ and we assume i.i.d Gaussian noise $\bm{\epsilon}\sim\mathcal{N}(0, \tau^2 I_{\datdim})$. The posterior may then be computed using \cref{eq:cond_mean_matrix,eq:cond_cov_matrix}. Note that the three-dimensional nature of the problem quickly makes it intractable for traditional inversion methods as the resolution of the inversion grid is refined. For example, when discretizing the Stromboli inversion domain as \citet{linde} into cubic cells of $50m$ side length, one is left with roughly $2\cdot 10^5$ cells, which translates to posterior covariance matrices of size $160~\mathrm{GB}$ (using 32 bits floating point numbers). This characteristic of gravimetric inverse problems make them well-suited to demonstrate the implicit representation framework which we introduced in \cref{sec:implicit}. In \cref{sec:optimal_design} we will show how our technique allows state-of-the-art adaptive design techniques to be applied to large real-world problems.

\subsection{Hyperparameter Optimization}\label{sec:hyperparams}
When using Gaussian process priors to solve inverse problems, one has to select the hyperparameters of the prior. There exists different approaches for optimizing hyperparameters. We here only consider maximum likelihood estimation (MLE).\\

We restrict ourselves to GP priors that have a constant prior mean $m_0\in\mathbb{R}$ and a covariance kernel $k$ that depends on a prior variance parameter $\sigma_0^2$ and other correlation parameters $\bm{\theta}_0\in\mathbb{R}^t$:
\begin{equation}\label{eq:kernel}
    k(x,y) = \sigma_0^2 r(x, y; \bm{\theta}_0),
\end{equation}
where $r(., .;\bm{\theta}_0)$ is a correlation function, such that $r(x,x;\bm{\theta}_0)=1, \forall x \in D$. The maximum likelihood estimator for the hyperparameters may then be obtained by minimizing the negative marginal log likelihood (nmll) of the data, which in the discretized setting of \cref{sec:compute_posterior} may be writen as \citep{rasmussen_williams}:
\begin{align}\label{eq:mll}
    \mathcal{L}\left(m_0, \sigma_0, \bm{\theta}_0;\bm{y}\right) &= 
    \frac{1}{2} \log \det R
    + \frac{1}{2} \Big(\bm{y} - \underbar{G} m_{\bm{X}} \Big)^T
    R^{-1} 
    \Big( \bm{y} - \underbar{G} m_{\bm{X}} \Big) + \frac{n}{2} \log 2\pi,\\
    R &:= \left(\underbar{G} K_{\bm{X} \bm{X}} \underbar{G}^T + \tau^2 \bm{I}_n\right).
\end{align}
Since only the quadratic term depends on $m_0$, we can adapt concentration identities \citep{park_concentration} to write the optimal $m_0$ as a function of the other hyperparameters:
\begin{equation}\label{eq:concentration}
\hat{m}_0^{MLE}\left(\sigma_0, \bm{\theta}_0\right) = \Big(\bm{1}_m^T \underbar{G}^T
R^{-1}  
\underbar{G} \bm{1}_m \Big)^{-1} y^T 
R^{-1} 
G \bm{1}_m,
\end{equation}
where $\bm{1}_{\dimpred}$ denotes the $\dimpred$-dimensional column vector containing only $1$'s. Here we always assume $R$ to be invertible. The remaining task is then to minimize the concentrated nmll:
\[
\left(\sigma_0, \bm{\theta}_0\right)\mapsto \mathcal{L}\left(\hat{m}_0^{MLE}\left(\sigma_0, \bm{\theta}_0\right),\sigma_0, \bm{\theta}_0\right).
\]
Note that the main computational challenge in the minimization of \cref{eq:mll} comes from the presence of the $\dimpred \times \dimpred$ matrix $K_{\bm{X}\bm{X}}$. In the following, we will only consider the case of kernels that depend on a single length scale parameter: $\bm{\theta}_0=\lambda_0\in\mathbb{R}$, though the procedure described below can in principle be adapted for multidimensional $\bm{\theta}_0$.\\

In practice, for kernels of the form \cref{eq:kernel} the prior variance $\sigma_0^2$ may be factored out of the covariance matrix (for known noise variance), so that only the prior length scale $\lambda_0$ appears in this large matrix. One then optimizes these parameters separately, using chunking (\cref{sec:chunking}) to compute matrix products. Since $\sigma_0$ only appears in an $\datdim \times \datdim$ matrix which does not need to be chunked (the data size $\datdim$ being moderate in real applications), one can use automatic differentiation libraries such as \citet{pytorch} to optimize it by gradient descent. On the other hand, there is no way to factor out $\lambda_0$ out of the large matrix $K_{\bm{X}\bm{X}}$, so we resort to a brute force approach by specifying a finite search space for it. To summarize, we proceed here in the following way:\\

\begin{enumerate}[label=(\roman*)]
\item (brute force search) Discretize the search space for the length scale by only allowing $\lambda_0 \in \Lambda_0$, where $\Lambda_0$ is a discrete set (usually equally spaced values on a reasonable search interval);
\item (gradient descent) For each possible value of $\lambda_0$, minimize the
    (concentrated) $\mathcal{L}$ over the remaining free parameter $\sigma_0$ by gradient descent.
\end{enumerate}

\begin{figure}[h!]
\begin{subfigure}{0.49\textwidth}
	\includegraphics[width=0.99\linewidth]{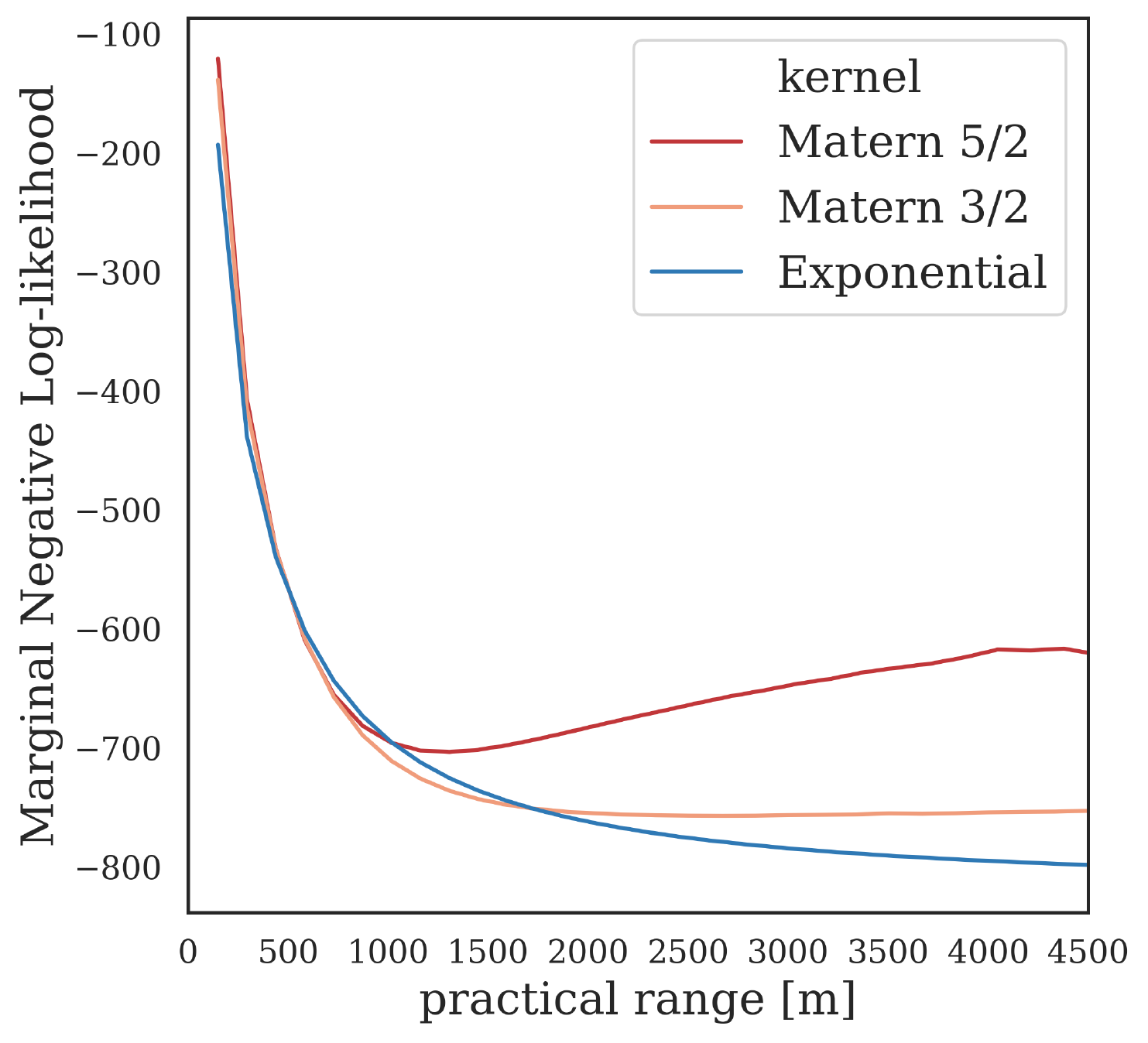}
	\caption{}
\end{subfigure}
\begin{subfigure}{0.49\textwidth}
	\includegraphics[width=0.99\linewidth]{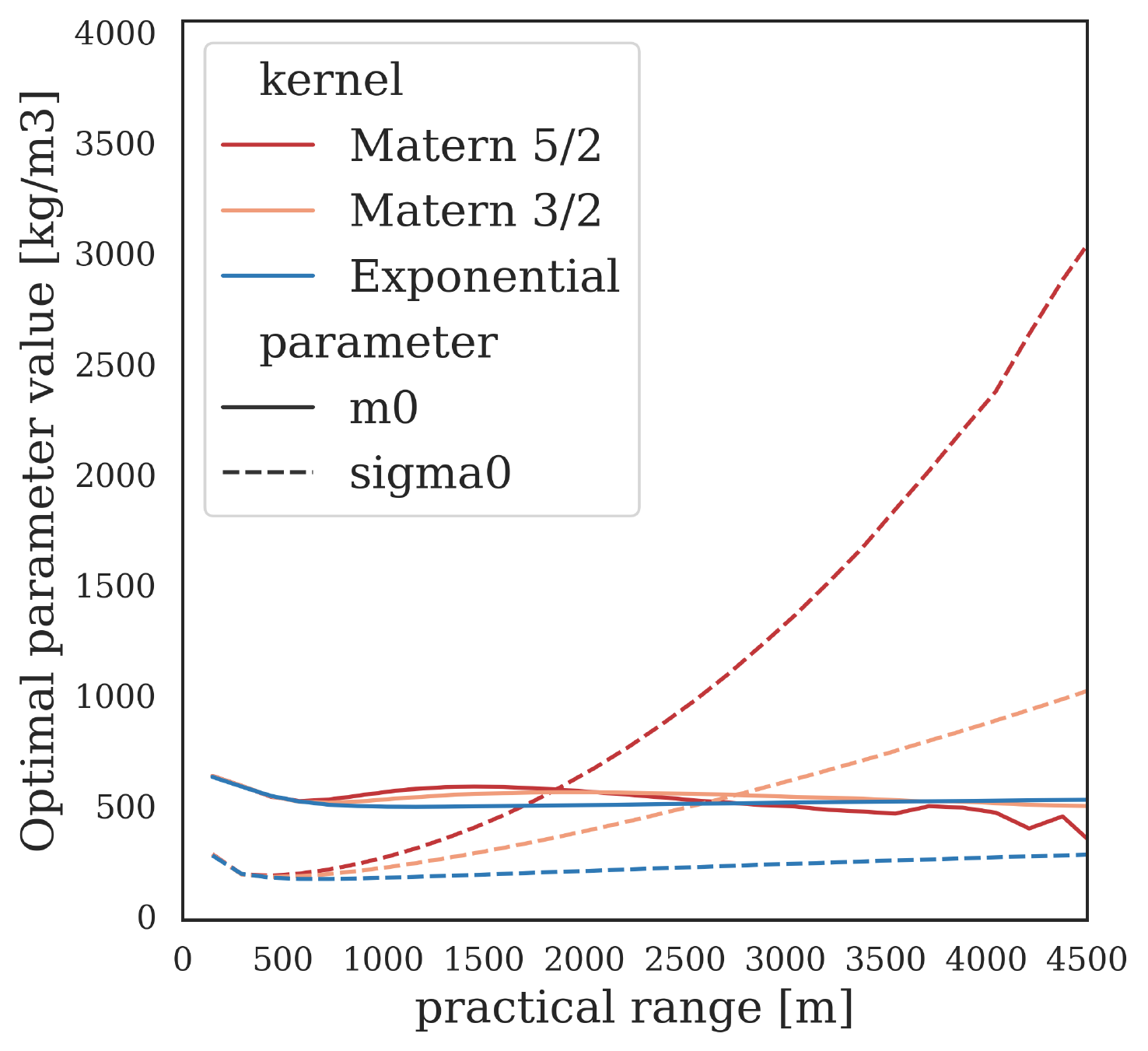}
	\caption{}
\end{subfigure}
\caption{(a) Concentrated negative marginal log-likelihood and (b) optimal hyperparameter values for different length scale parameters $\lambda_0$.}
\label{fig:hyperparam_optim}
\end{figure}

We ran the above approach on the Stromboli dataset with standard stationary kernels (Mat\'{e}rn $3/2$, Mat\'{e}rn $5/2$, exponential). In agreement with \citet{linde}, the observational noise 
standard deviation is $0.1~[mGal]$. The optimization results for different values of the length scale parameter are shown in \cref{fig:hyperparam_optim}. The best estimates of the parameter values for each kernel are shown in \cref{tab:table_hyper}. The table also shows the practical range $\bar{\lambda}$ which is defined as the distance at which the covariance falls to $5\%$ of its original value.

\begin{figure}[h!]
\centering
\begin{subfigure}{0.49\textwidth}
\includegraphics[width=0.99\linewidth]{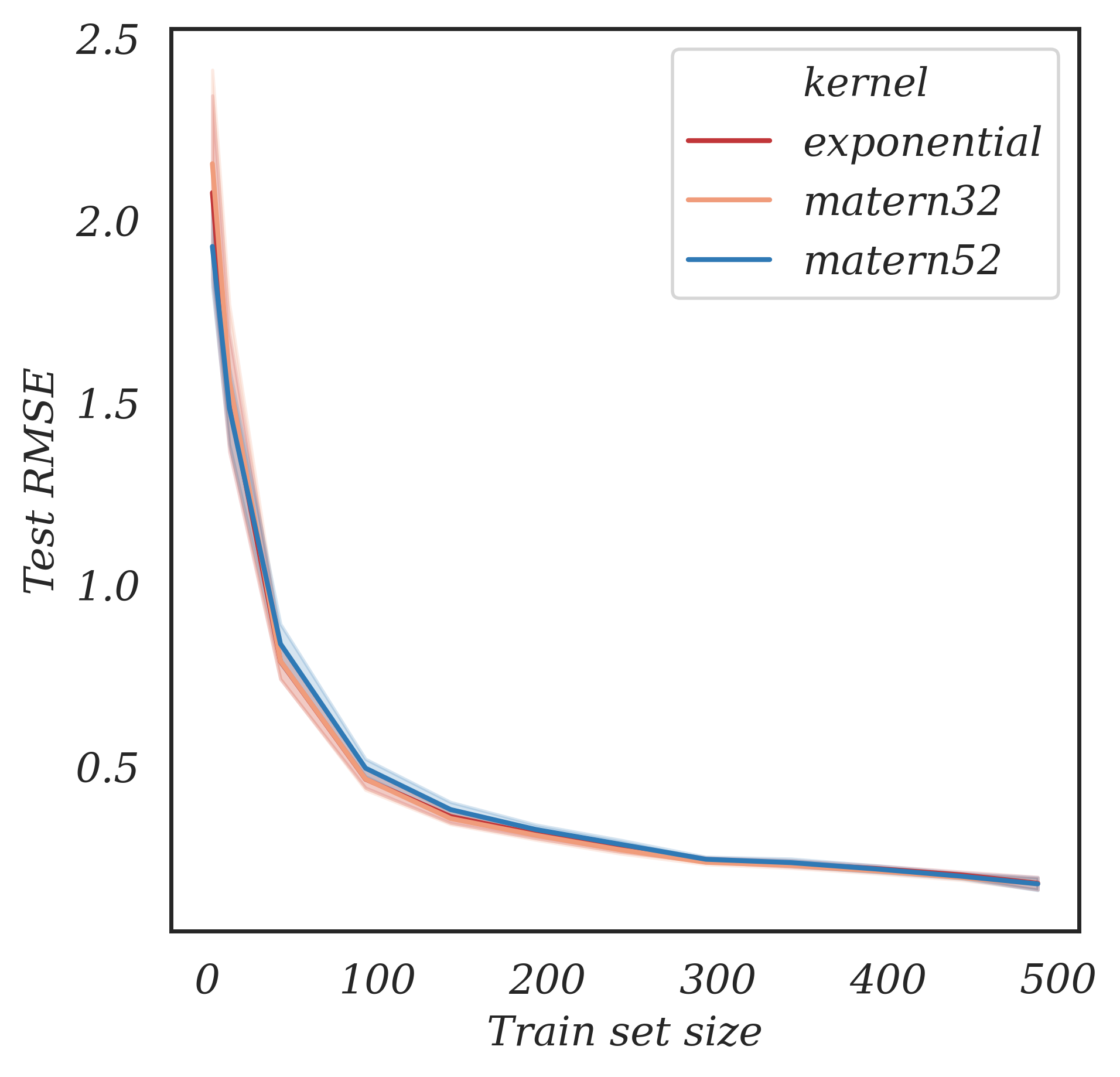}
\caption{}
\end{subfigure}
\begin{subfigure}{0.49\textwidth}
\includegraphics[width=0.99\linewidth]{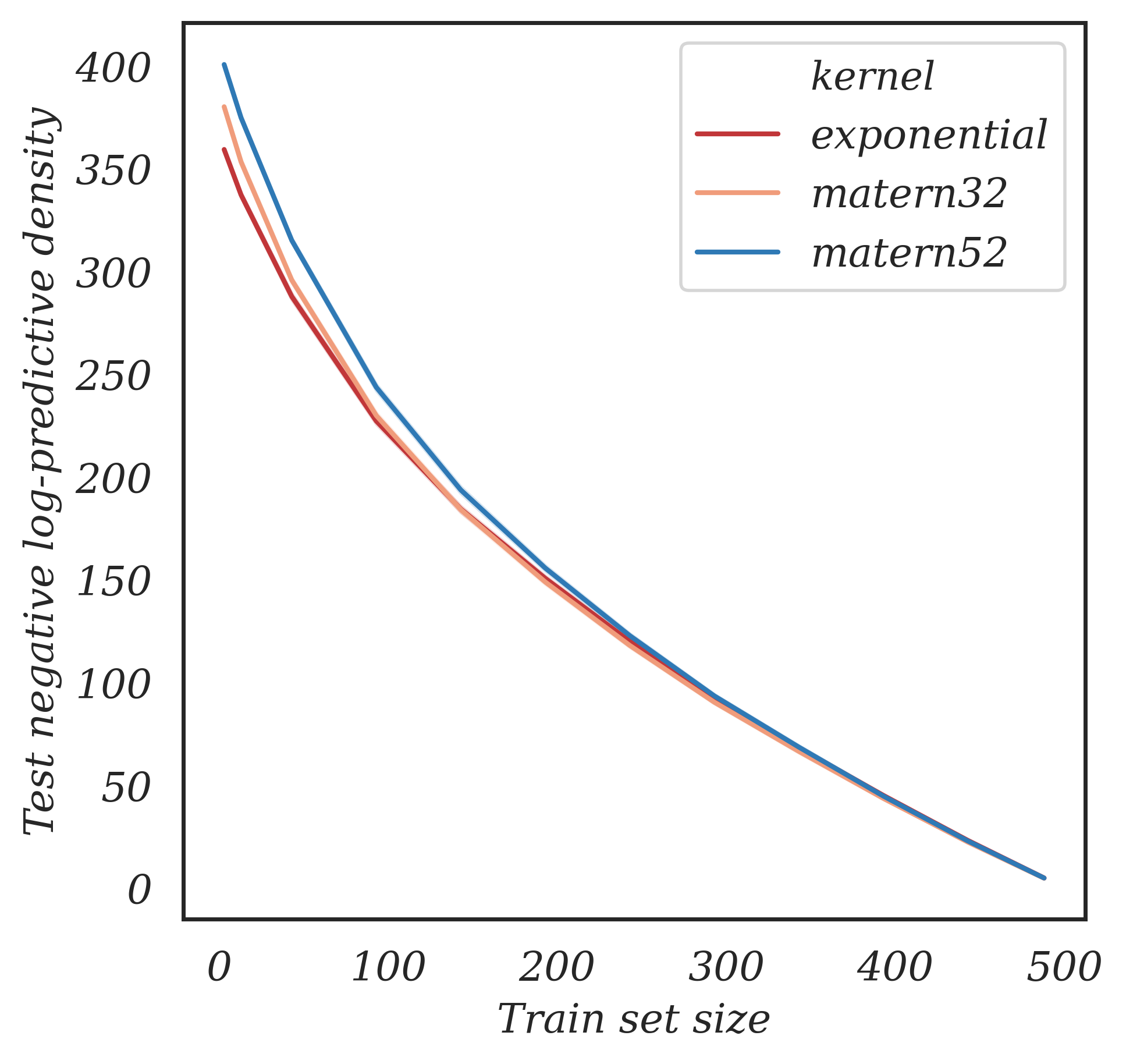}
\caption{}
\end{subfigure}
\caption{(a) Root mean squared error and (b) negative log predictive density on test set for the different models (with optimal hyperparameters). The full dataset contains 501 observations.}
\label{fig:test_set_evolution}
\end{figure}

We asses the robustness of each kernel by predicting a set of left out observations 
using the other remaining observations. \cref{fig:test_set_evolution} displays 
RMSE and negative log predictive density for different proportion of train/test splits.
\begin{table}[h!]
\begin{tabular}{ c c c c c c c c c } 
\toprule
 & & \multicolumn{4}{c}{ Hyperparameters } & & \multicolumn{2}{c}{ Metrics } \\
 \cmidrule{3-6} \cmidrule{8-9}
 
     Kernel  & &  $\lambda$  &  $\bar{\lambda}$  &  $m_0$  &  $\sigma_0$  & &  $\mathcal{L}$  & Train RMSE \\
\midrule
    Exponential & & 1925.0 & 5766.8 & 535.4 & 308.9 & & -804.4 & 0.060\\ 
    \textbf{Matérn 3/2} & & \textbf{651.6} & \textbf{1952.0} & \textbf{2139.1} & \textbf{284.65} & & \textbf{-1283.5} & \textbf{0.071}\\ 
    Matérn 5/2 & & 441.1 & 1321..3 & 2120.9 & 349.5 & & -1247.6 & 0.073\\ 
\end{tabular}
\caption{Optimal hyperparameters (Stromboli dataset) for different kernels.}
\label{tab:table_hyper}
\end{table}

Note the the above procedure is more of a quality assurance than a rigorous statistical evaluation of the model, since all datapoints were already used in the fitting of the hyperparameters. 
Due to known pathologies of the exponential kernel 
(MLE for length scale parameter going to infinity), we choose to use the Mat\'{e}rn 3/2 model for the experiments of \cref{sec:sampling} and \cref{sec:optimal_design}. The maximum likelihood estimator of the prior hyperparameters for this model are $\hat{m}_0^{MLE}=2139.1~[kg/m^3]$, $\hat{\sigma}_0^{MLE} = 284.65~[kg/m^3]$ and $\hat{\lambda}_0^{MLE}=651.6~[m]$.

\subsection{Posterior Sampling}\label{sec:sampling}
Our implicit representation also allows for efficient sampling from the posterior by
using the \textit{residual kriging} algorithm \citep{chiles_delfiner,
de_fouquet_cond_kriging}, which we here adapt to linear operator observations. Note that in order to sample a Gaussian process at $\dimpred$ sampling points, one needs to generate $\dimpred$ correlated Gaussian random variables, which involves covariance matrices of size $\dimpred^2$, leading to the same computational bottlenecks as described in \cref{sec:compute_posterior}. On the other hand, the residual kriging algorithm generates realizations from the
posterior by updating realizations of the prior, as we explain next.\\

As before, suppose we have a GP $Z$ defined on some compact Euclidean domain $D$ and assume 
$Z$ has continuous sample paths almost surely. Furthermore, say we have $\datdim$ observations 
described by linear operators $\ell_1, ..., \ell_{\datdim} \in C(D)^*$. 
Then the conditional expectation of $Z$ conditional on the $\sigma$-algebra 
$\Sigma:=\sigma\left(\ell_1\left(Z\right), ..., \ell_{\datdim}\left(Z\right)\right)$ 
is an orthogonal projection (in the $L^2$-sense \citep{williams}) of $Z$ onto $\Sigma$. 
This orthogonality can be used to decompose the conditional law of $Z$ conditional on $\Sigma$ 
into a conditional mean plus a residual. 
Indeed, if we let $Z^{'}$ be another GP with the same distribution as $Z$ and let 
$\Sigma^{'} :=\sigma\left(\ell_1\left(Z^{'}\right), ..., \ell_{\datdim}\left(Z^{'}\right)\right)$,
then we have the following equality in distribution:
\begin{align}
    Z_x\mid \Sigma &= \expec{Z_x \mid \Sigma} + \left(Z^{'}_x -
	\expec{Z^{'}_x \mid \Sigma^{'}}\right),~ \text{all }x\in D.\label{th:reskrig}
\end{align}
Compared to direct sampling of the posterior, the above approach involves two main operations: 
sampling from the prior and conditioning under operator data. When the covariance kernel is stationary and belongs to one of the usual families (Gaussian, Mat\'{e}rn), methods exist to sample from the prior on large grids \citep{turning_bands}; whereas the conditioning part may be performed using our implicit representation.

\begin{remark}
Note that in a sequential setting as in \cref{sec:seq_assimilation}, the residual kriging algorithm  
may be used to maintain an ensemble of realizations from
the posterior distribution by updating a fixed set of prior realizations at
every step in the spirit of \citet{foxy}.
\end{remark}

\subsection{Sequential Experimental Design for Excursion Set Recovery}\label{sec:optimal_design}
As a last example of application where our implicit update method provides substantial
savings, we consider a sequential data collection task involving an inverse problem.
Though sequential design criterion for inverse problems have already
been considered in the litterature \citep{opt_des_inverse}, most of them only focus on selecting observations to improve the reconstruction of the unknown parameter field, or some linear functional thereof.

We here consider a different setting. In light of recent progress in excursion set estimation 
\citep{azzimonti_uq,chevalier_uq}, we instead focus on the task of recovering
an excursion set of the unknown parameter field $\density$, 
that is, we want to learn the unknown set 
$\trueExcuSet := \lbrace x \in \domain :
\density\left(x\right)\geq\thresh\rbrace$, where $\thresh$ is some threshold. 
In the present context of Stromboli, high density areas are related to dykes (previous feeding conduits of the volcano), while low density values are related to deposits formed by paroxysmal explosive phreato-magmatic events \citep{linde}. 
To the best of our knowledge such sequential experimental design problems 
for excursion set learning in inverse problems have not been considered
elsewhere in the litterature.

\begin{remark}
For the sake of simplicity, we focus only on excursion sets above some threshold, but all the techniques presented here may be readily generalized to generalized excursion sets of the form  $\trueExcuSet := \lbrace x \in \domain :
\density\left(x\right) \in I \rbrace$ where $I$ is any finite union of intervals on the extended real line.
\end{remark}
\begin{figure}[tbhp]
\begin{subfigure}{0.31\textwidth}
	\includegraphics[width=0.99\linewidth]{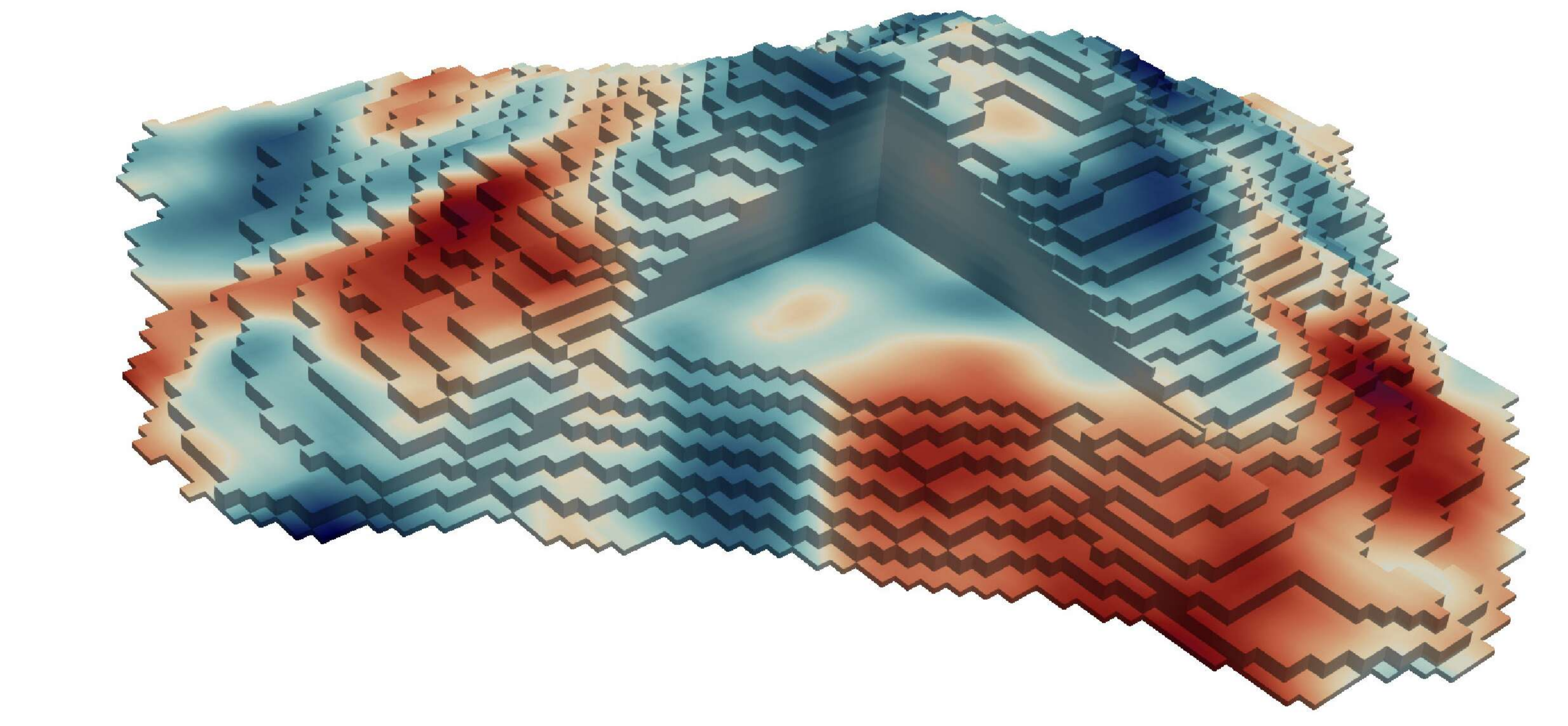}
	\caption{}
\end{subfigure}
\hspace{3mm}%
\begin{subfigure}{0.31\textwidth}
	\includegraphics[width=0.99\linewidth]{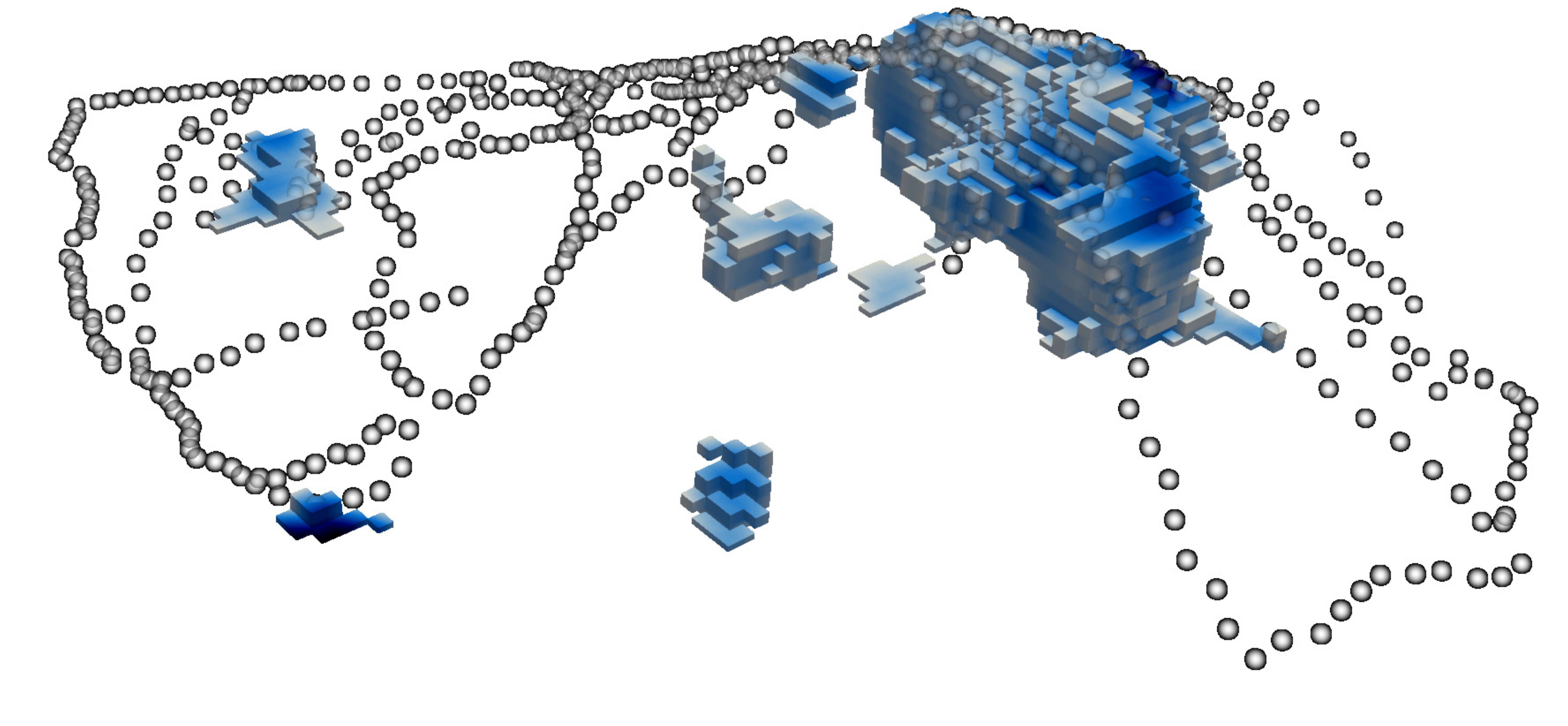}
	\caption{}
\end{subfigure}
\hspace{3mm}%
\begin{subfigure}{0.31\textwidth}
	\includegraphics[width=0.99\linewidth]{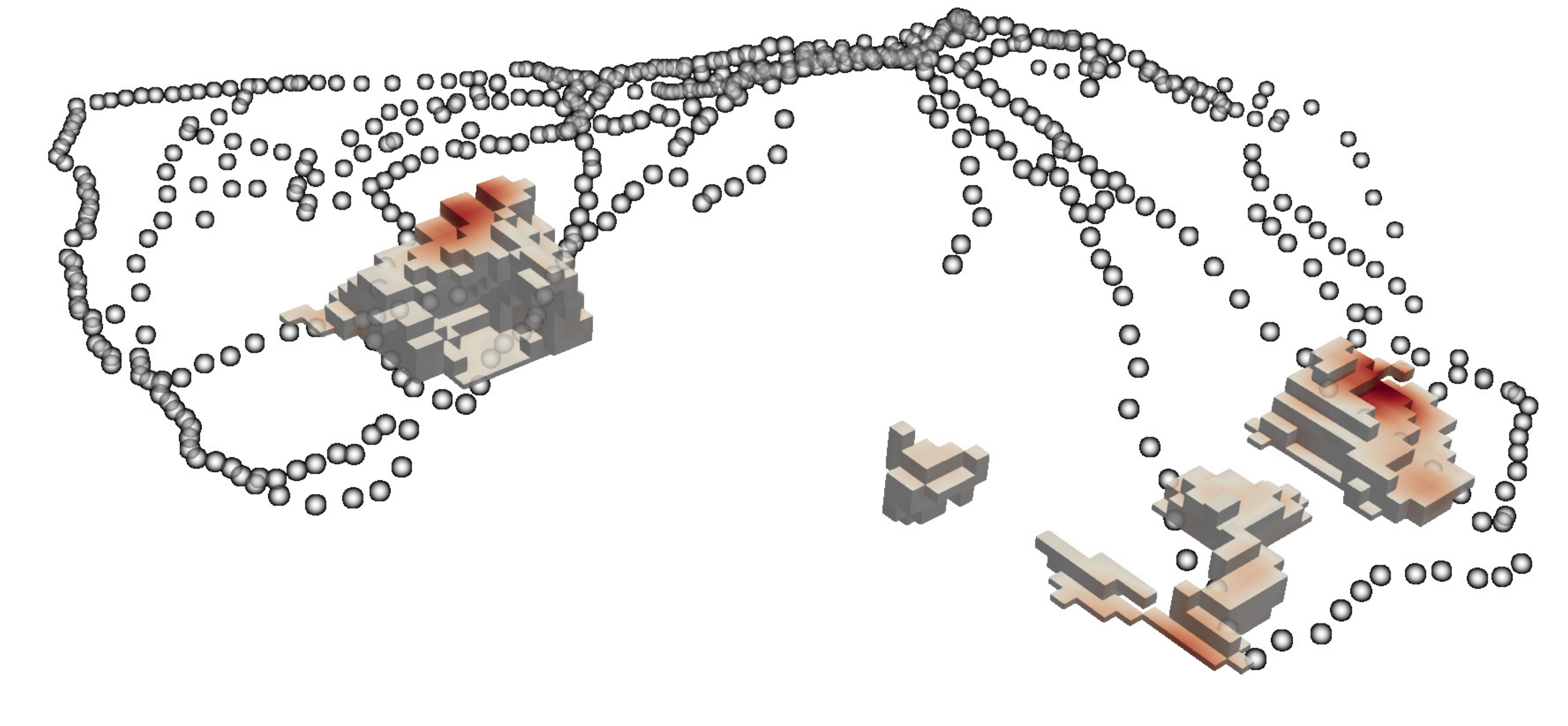}
	\caption{}
\end{subfigure}	
\\
\begin{subfigure}{0.31\textwidth}
	\includegraphics[width=0.99\linewidth]{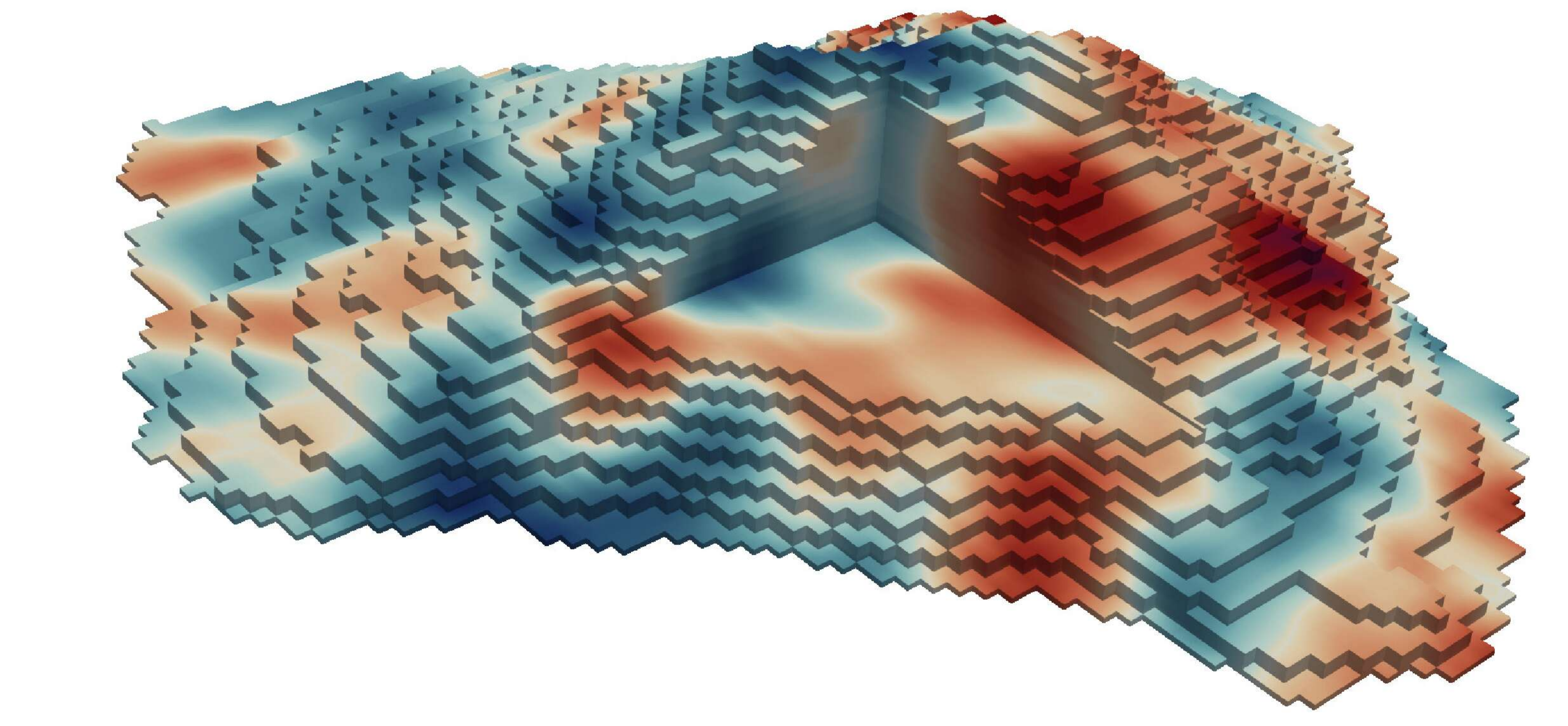}
	\caption{}
\end{subfigure}
\hspace{3mm}%
\begin{subfigure}{0.31\textwidth}
	\includegraphics[width=0.99\linewidth]{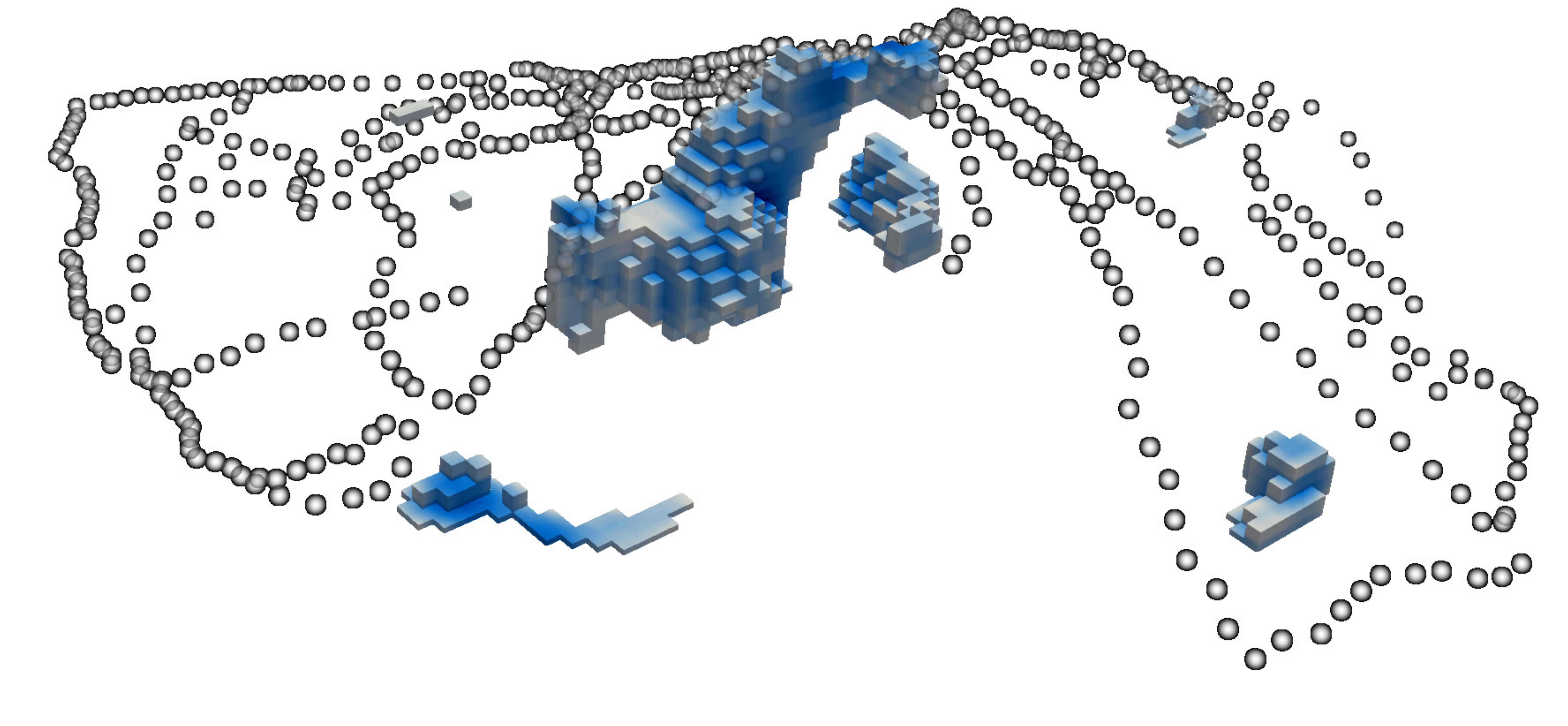}
	\caption{}
\end{subfigure}
\hspace{3mm}%
\begin{subfigure}{0.31\textwidth}
	\includegraphics[width=0.99\linewidth]{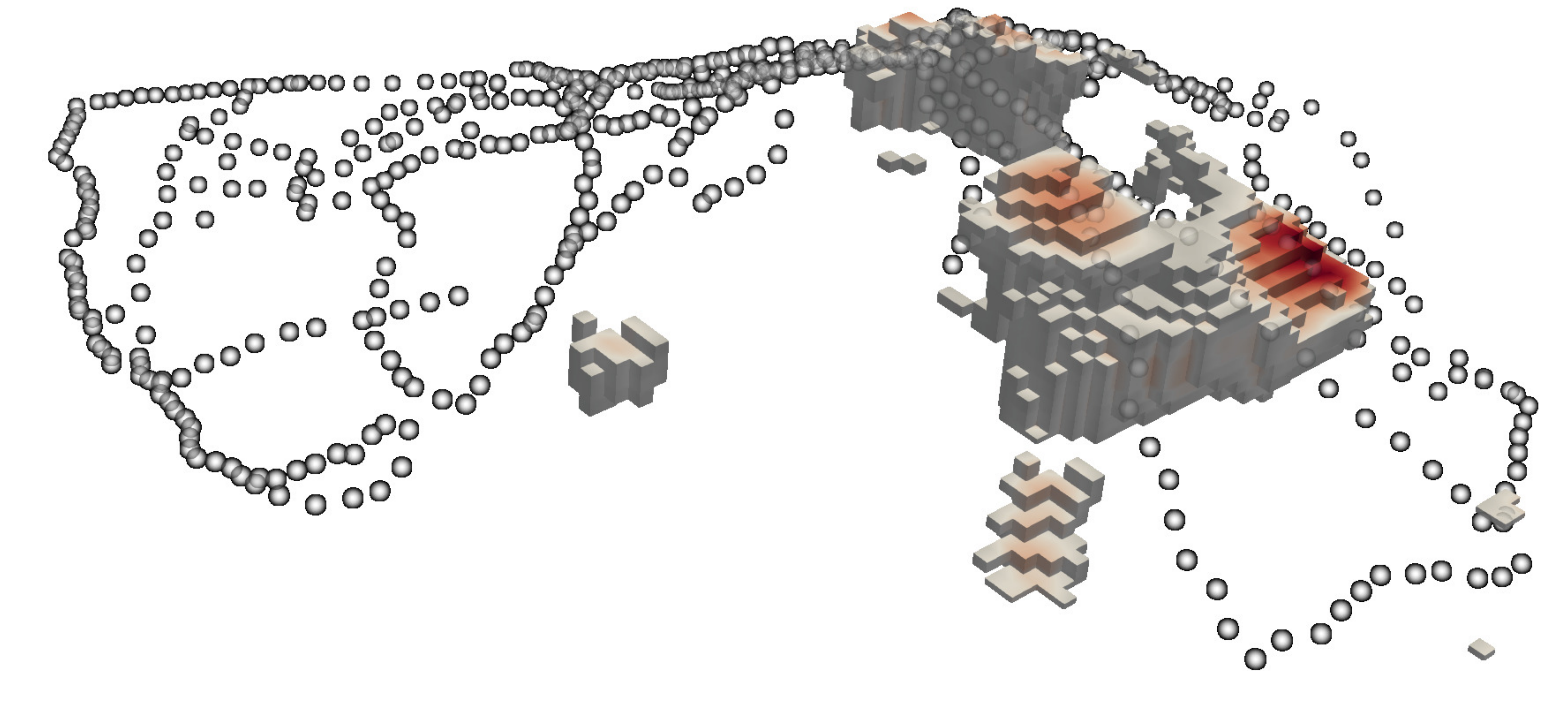}
	\caption{}
\end{subfigure}
\\
\begin{subfigure}{0.31\textwidth}
	\includegraphics[width=0.99\linewidth]{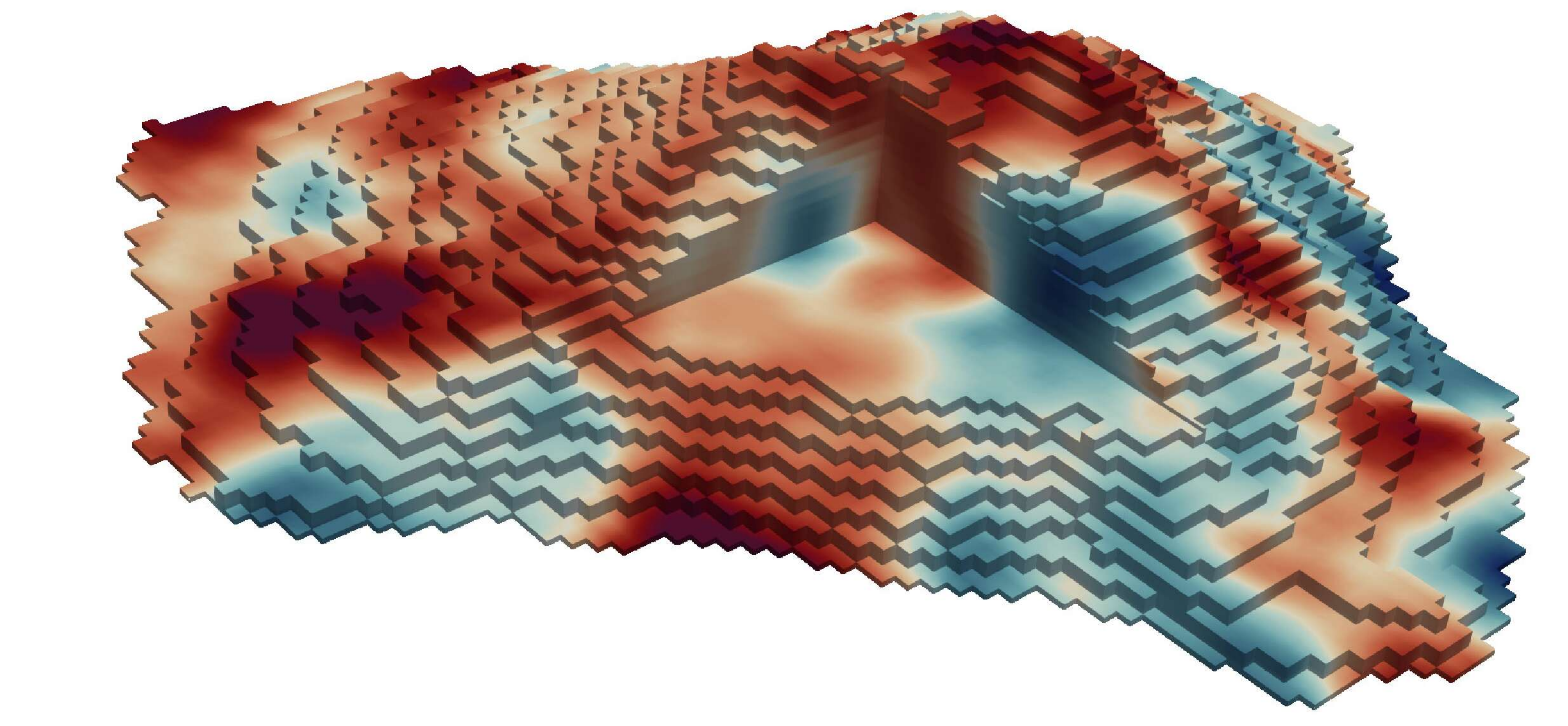}
	\caption{}
\end{subfigure}
\hspace{3mm}%
\begin{subfigure}{0.31\textwidth}
	\includegraphics[width=0.99\linewidth]{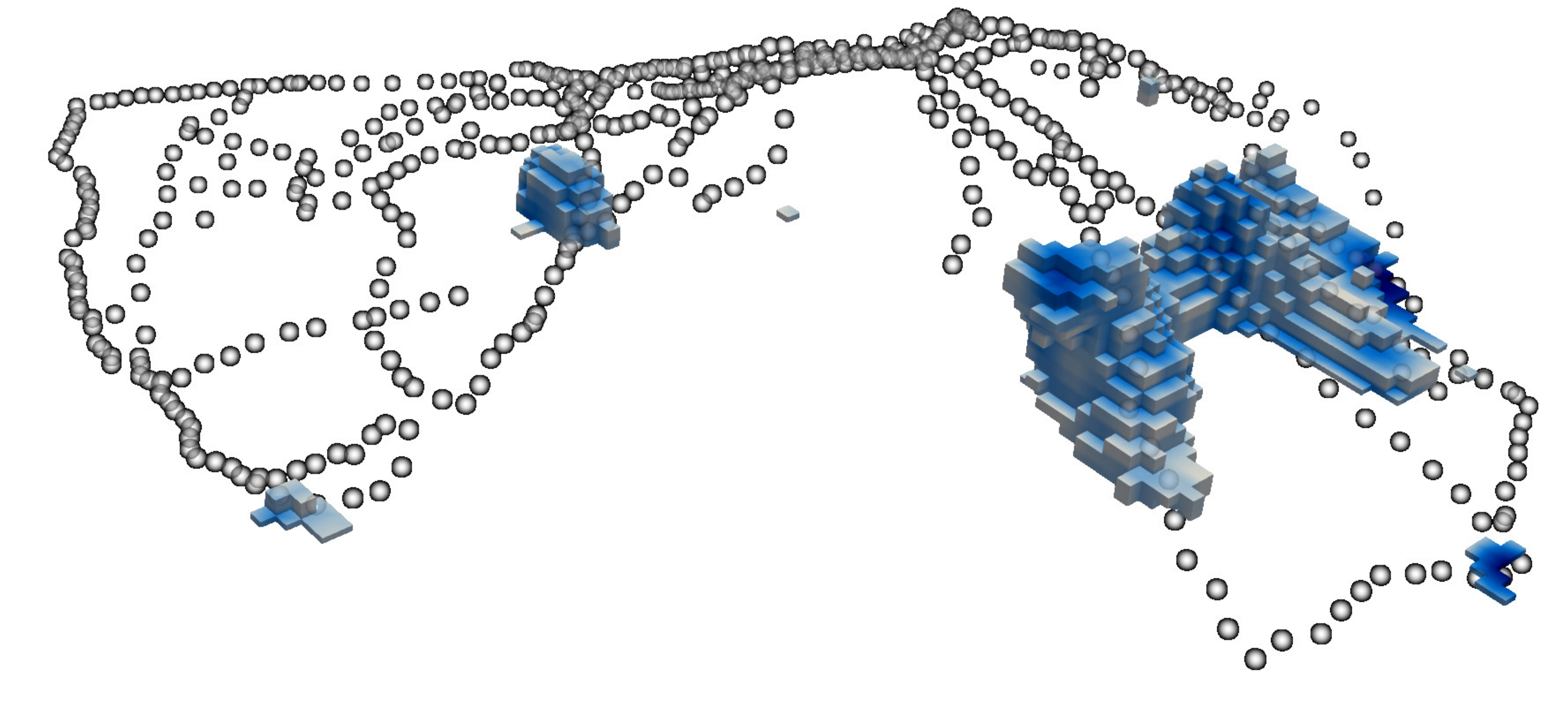}
	\caption{}
\end{subfigure}
\hspace{3mm}%
\begin{subfigure}{0.31\textwidth}
	\includegraphics[width=0.99\linewidth]{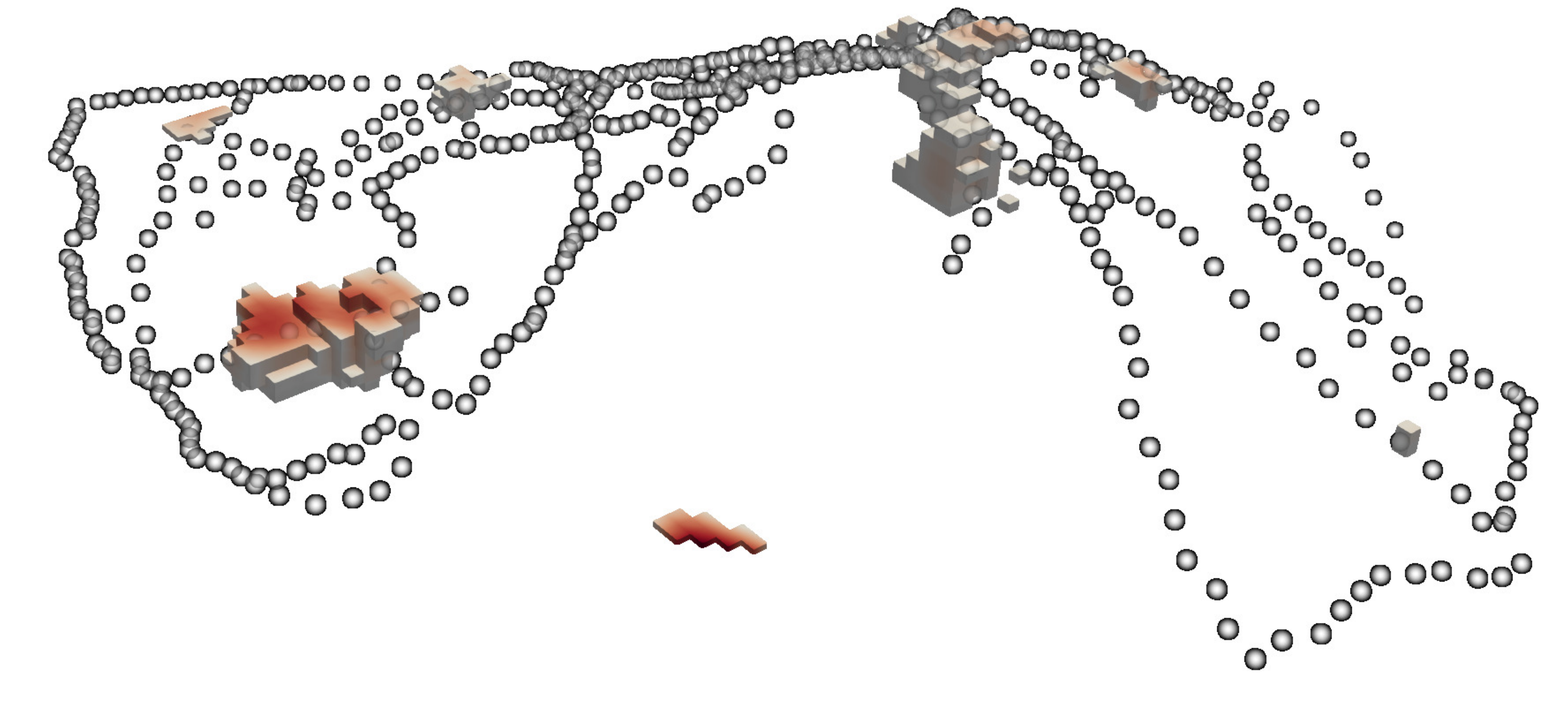}
	\caption{}
\end{subfigure}
\caption{Realizations from Mat\'{e}rn $3/2$ GP prior (hyperparameters taken from \cref{tab:table_hyper}) with corresponding excursion sets: (left to right) Underground mass density field (arbitrary colorscale), high density regions and low density regions, thresholds: $2600~[kg/m^3]$ and $1700~[kg/m^3]$.}
\label{fig:prior_overview}
\end{figure}
We here consider a sequential setting, where observations are made one at a time and at each stage we have to select which obsevation to make next in order to optimally reduce the uncertainty on our estimate of $\trueExcuSet$.
Building upon \citet{weighted_imse,bect2012sequential,azzimonti_adaptive,chevalier_2014_fast}, 
there exists several families of criteria to select the next observations. Here, we
restrict ourselves to a variant of the weighted IMSE criterion \citep{weighted_imse}. The
investigation of other state-of-the-art criteria is left for future work. We note in passing that most Bayesian sequential design criteria 
involve posterior covariances and hence tend to become 
intractable for large-scale problems. Moreover in a sequential setting, fast updates of the posterior covariance are crucial. Those
characteristics make the problem considered here particularly suited for the implicit update framework intoduced in \cref{sec:implicit}.\\

The weighted IMSE criterion selects next observations by maximizing the variance
reduction they will provide at each location, weighted by the probability
for that location to belong to the excursion set $\trueExcuSet$. Assuming that $\stage$ data collection stages have already
been performed and using the notation of \cref{sec:seq_assimilation}, 
the variant that we are considering here selects the next observation location 
by maximizing the \textit{weighted integrated variance
reduction} (wIVR):
    \begin{equation}
        \textrm{wIVR}^{\stage}(\site) = \int_{\domain}
        \left(K^{(\stage)}_{xx} - K^{(\stage+1)}_{xx}\left[\fwd_{\site}\right]\right) p_{n}(x) dx,
    \end{equation}
where $s$ is some potential observation location, $K^{(n+1)}$ denotes the conditional covariance after including a gravimetric observation made at $s$ (this quantity is independent of the
observed data) and $\fwd_{s}$ is the forward operator corresponding to this observation. Also, here $p_{\stage}$ denotes the 
\textit{coverage function} at stage $\stage$ (we refer the reader to  \cref{sec:bayesian_set_estimation} for more details on Bayesian set estimation). After discretization, applying \cref{th:seq_update}
turns this criterion into:
    \begin{equation}\label{eq:weighted_ivr}
        \sum_{x\in \predpts} \covN_{x\predpts} \fwd_{s}^T\left(\fwd_{s} 
            \covN_{\predpts \predpts} \fwd_{s
    }^T + \tau^2 \id\right)^{-1}\fwd_{s}\covN_{\predpts x} p_{\stage}(x),
    \end{equation}
where we have assumed that all measurements are affected by $\mathcal{N}(0, \tau^2)$ distributed noise.\\

Note that for large-scale problems, the wIVR criterion in the form 
given in \cref{eq:weighted_ivr} becomes intractable for traditional methods 
because of the presence of the full 
posterior covariance matrix $\covN_{\predpts\predpts}$ in the parenthesis. 
The implicit representation presented in \cref{sec:implicit} can be used 
to overcome this difficulty. Indeed, the criterion can be evaluated 
using the posterior covariance multiplication routine \cref{th:sequential} 
(where the small dimension $q$ is now equal to the number of candidate observations considered 
at a time, here $1$ but batch acquisition scenarios could also be tackled). 
New observations can be seamlessly integrated along the way by updating the representation
using \cref{alg:update_interm}.\\

\textbf{Experiments and Results:}
We now study how the wIVR criterion can help to reduce the uncertainty on excursion sets within the Stromboli volcano. We here focus on recovering the volume of the excursion set instead of its precise location.
To the best of our knowledge, in the existing literature such sequential design criteria for excursion set recovery have only been applied to small-scale inverse problems and have not been scaled to larger, more realistic problems where the dimensions at play prevent direct access to the posterior covariance matrix.\\

\begin{figure}[h!]
\centering
  \includegraphics[width=0.67\linewidth,height=0.6\linewidth]{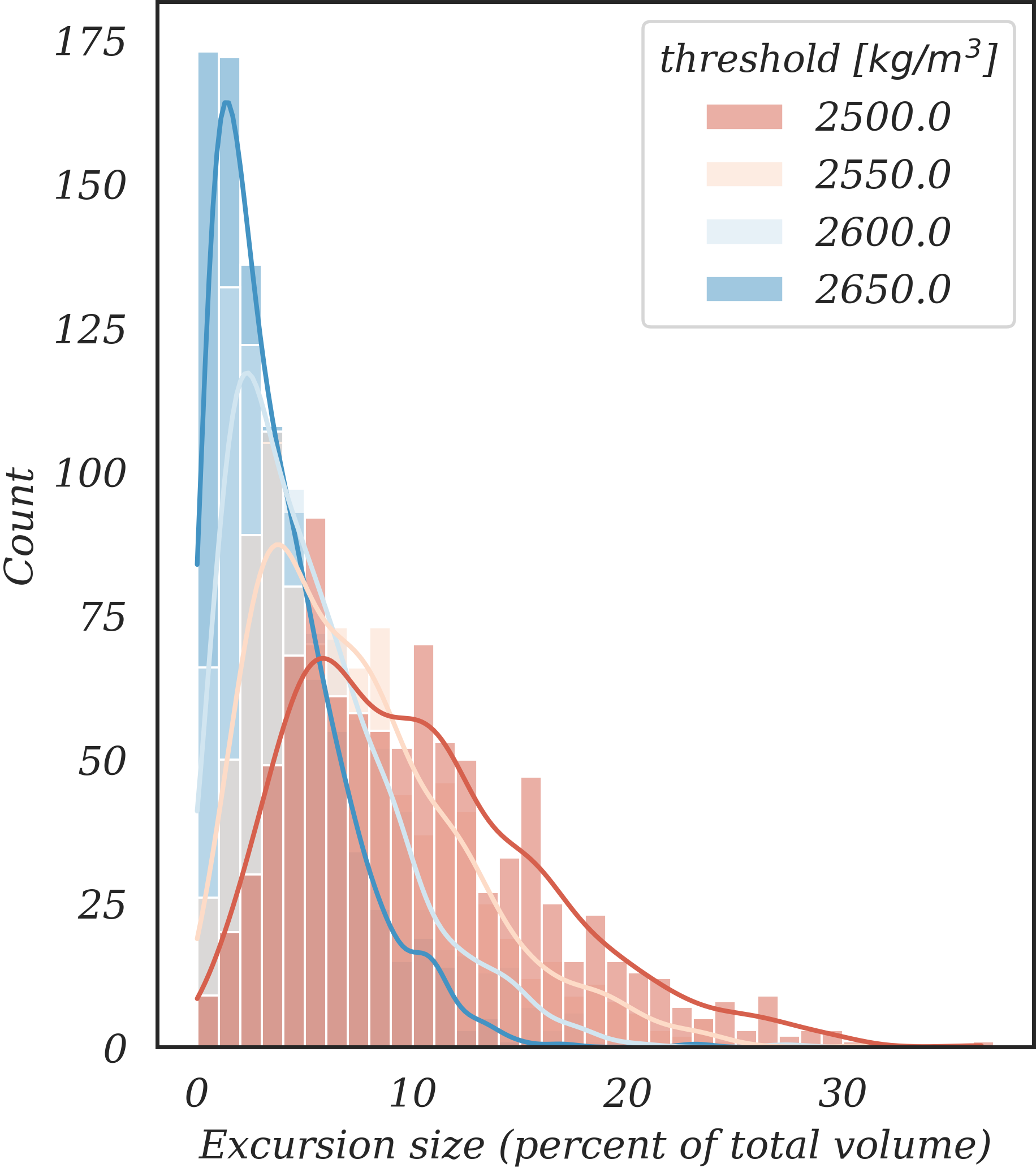}
  \caption{Distribution of excursion set volume under the prior for different thresholds. Size is expressed as a percentage of the volume of the inversion domain.}\label{fig:excu_volume_histogram}
\end{figure}

In the following experiments, we use the Stromboli volcano inverse problem and work with a discretization into cubic cells of $50~[m]$ side length. We use a Mat\'{e}rn $3/2$ GP prior with hyperparameters trained on real data (\cref{tab:table_hyper}) to generate semi-realistic ground truths for the experiments. We then simulate numerically the data collection process by computing the response that results from the considered ground truth and adding random observational noise. When computing sequential designs for excursion set estimation, the threshold that defines the excursion set can have a large impact on the accuracy of the estimate. Indeed, different thresholds will produce excursion sets of different sizes, which may be easier or harder to estimate depending on the set estimator used. For the present problem, \cref{fig:excu_volume_histogram} shows the distribution of the excursion volume under the considered prior for different excursion thresholds.

It turns out that the estimator used in our experiments (Vorob'ev expectation) behaves differently depending on the size of the excursion set to estimate. Indeed, the Vorob'ev expectation tends to produce a smoothed version of the true excursion set, which in our situation results in a higher fraction of false positives for larger sets. Thus, we consider two scenarios: a \textit{large} scenario where the generated excursion sets have a mean size of $10\%$ of the total inversion volume and a \textit{small} scenario where the excursion sets have a mean size of $5\%$ of the total inversion volume.
One should note that those percentages are in broad accordance with the usual size of excursion sets that are of interest in geology. The chosen thresholds are $2500~[kg/m^3]$ for the \textit{large} excursions and $2600~[kg/m^3]$ for the \textit{small} ones.
\begin{figure}[h!]
\centering
 \begin{subfigure}{0.48\textwidth}
 	\includegraphics[width=0.99\linewidth]{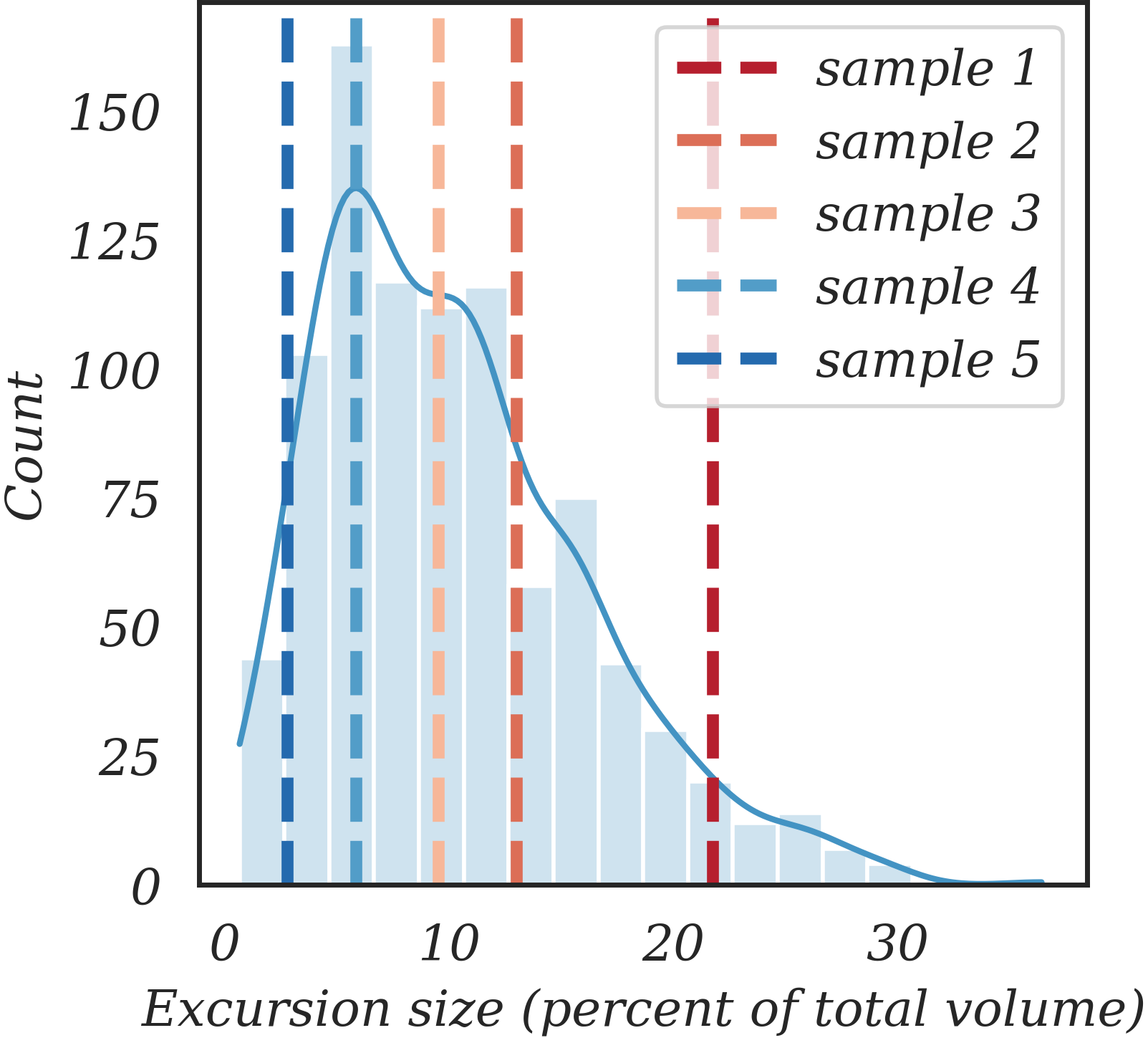}
	\caption{(\textit{large} scenario) threshold: 2500 ${[}kg/m^3{]}$}
	\end{subfigure} 
  \hfill
\begin{subfigure}{0.48\textwidth}
	\includegraphics[width=0.99\linewidth]{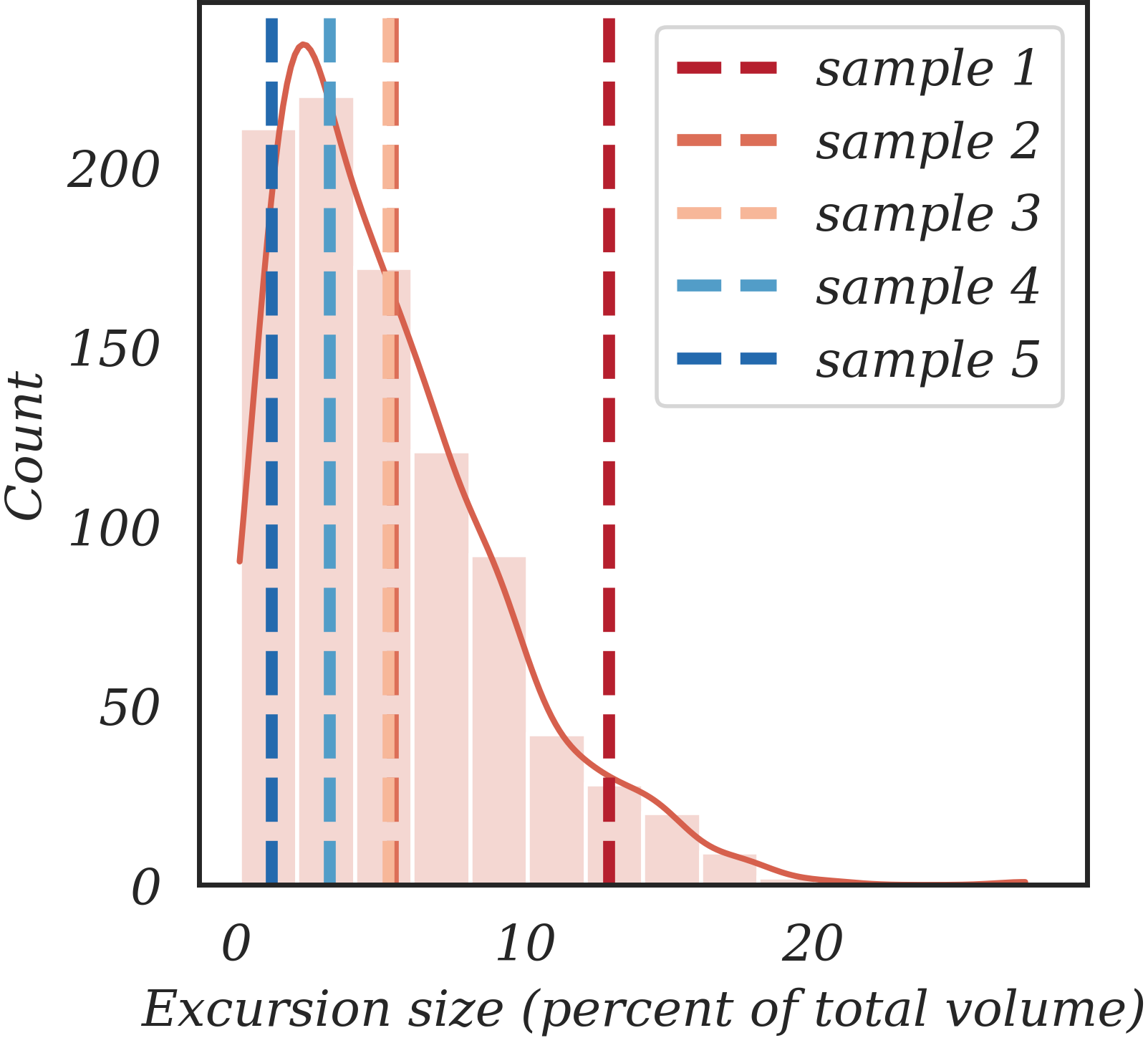}
	\caption{(\textit{small} scenario) threshold: $2600 {[}kg/m^3{]}$}
\end{subfigure}
\caption{Distribution of excursion volume (with kernel density estimate) under the prior for the two considered thresholds, together with excursion volumes for each ground truth.}\label{fig:excu_volume_scenarios}
 \end{figure}

\begin{figure}[h]
    \centering
\begin{subfigure}{0.75\textwidth}
	\includegraphics[width=0.99\linewidth]{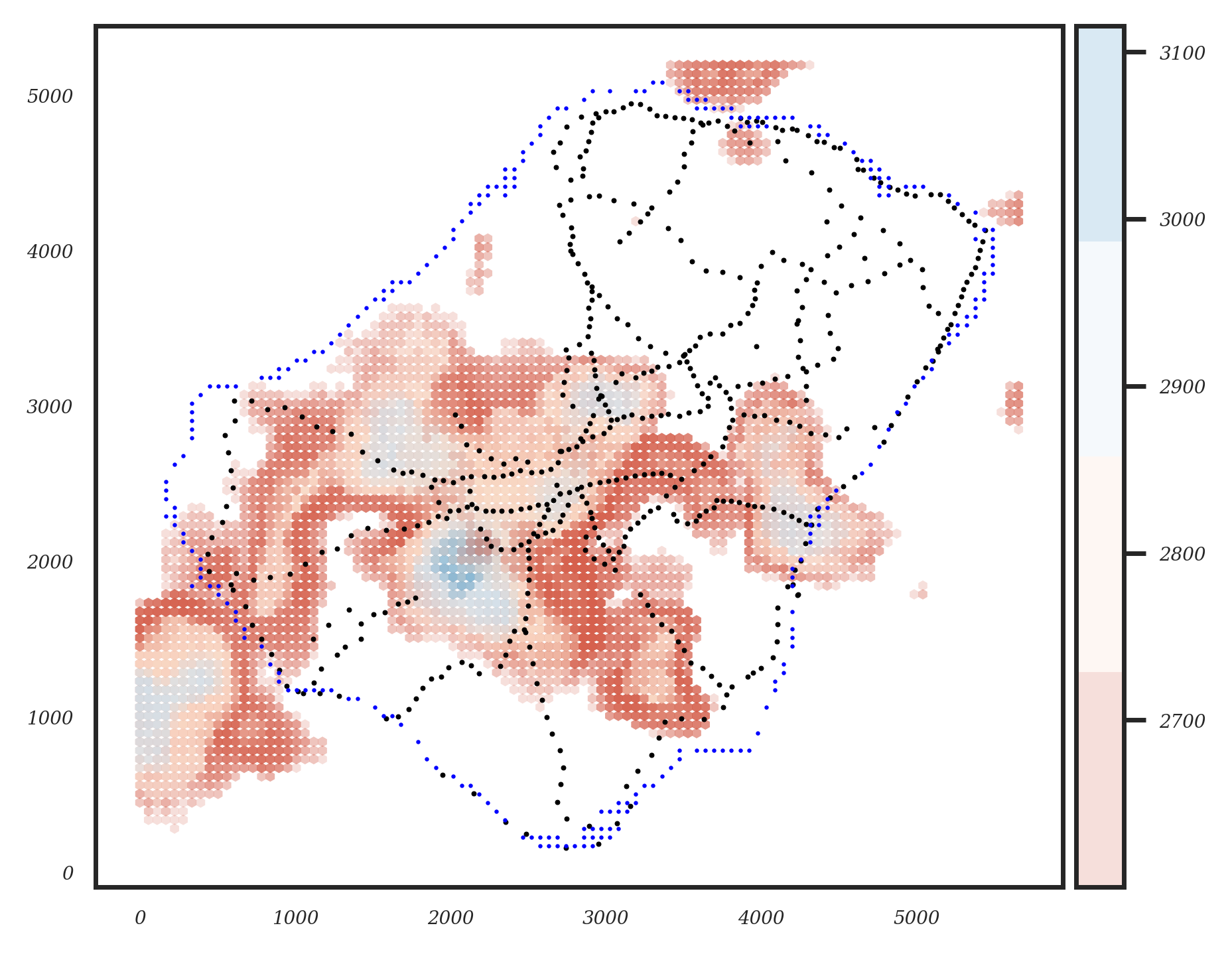}
	\caption{xy projection}
\end{subfigure}\\
\begin{subfigure}{0.49\textwidth}
	\includegraphics[width=0.99\linewidth]{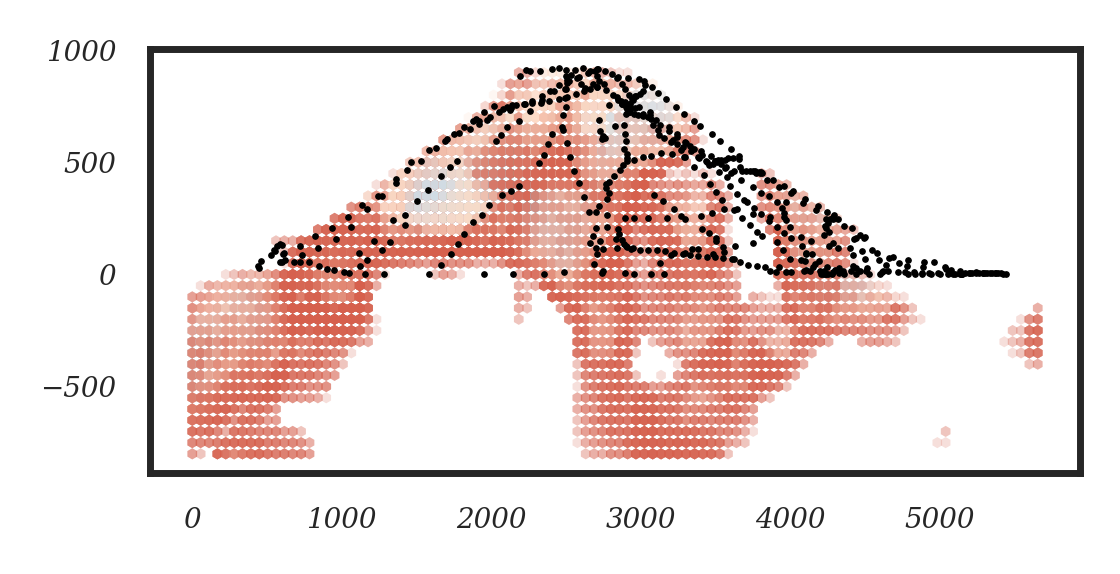}
	\caption{xz projection}
\end{subfigure}
\begin{subfigure}{0.49\textwidth}
	\includegraphics[width=0.99\linewidth]{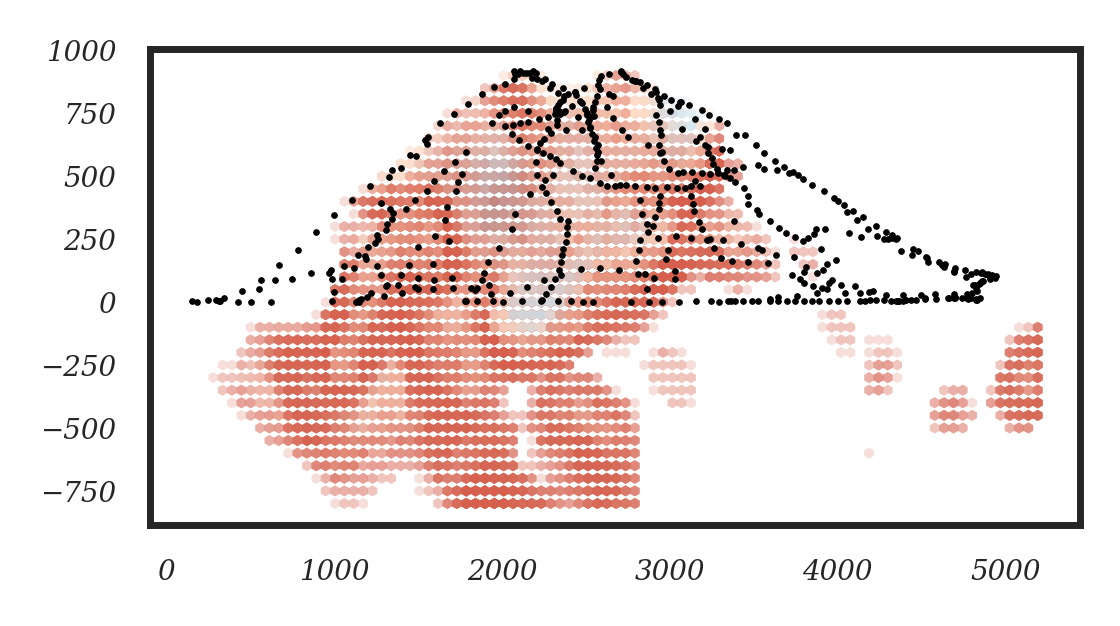}
	\caption{yz projection}
\end{subfigure}
\caption{Projection of the excursion set (small scenario) for the first ground truth. Island boundary denoted in blue, observation location from previous field campaign denoted by black dots. Distances are displayed in [m] and density in [kg/$m^3$].}
\label{fig:ground_truth_excu_profile}
\end{figure}

The experiments are run on five different ground truths, which are samples from a Mat\'{e}rn $3/2$ GP prior (see previous paragraphs). The samples were chosen such that their excursion set for the \textit{large} scenario have volumes that correspond to the $5\%$, $27.5\%$, $50\%$, $72.5\%$ and $95\%$ quantiles of the prior excursion volume distribution for the corresponding threshold. \cref{fig:excu_volume_scenarios} shows the prior excursion volume distribution together with the volumes of the five different samples used for the experiments. \cref{fig:ground_truth_excu_profile} shows a profile of the excursion set (small scenario) for one of the five samples used in the experiments. The data collection location from the 2012 field campaign \citep{linde} are denoted by black dots. The island boundary is denoted by blue dots. Note that, for the sake of realism, in the experiments we only allow data collection at locations that are situated on the island (data acquired on a boat would have larger errors); meaning that parts of the excursion set that are outside of the island will be harder to recover.

\begin{figure}[t]
\centering
\begin{subfigure}[b]{0.48\textwidth}
	\includegraphics[width=0.99\linewidth]{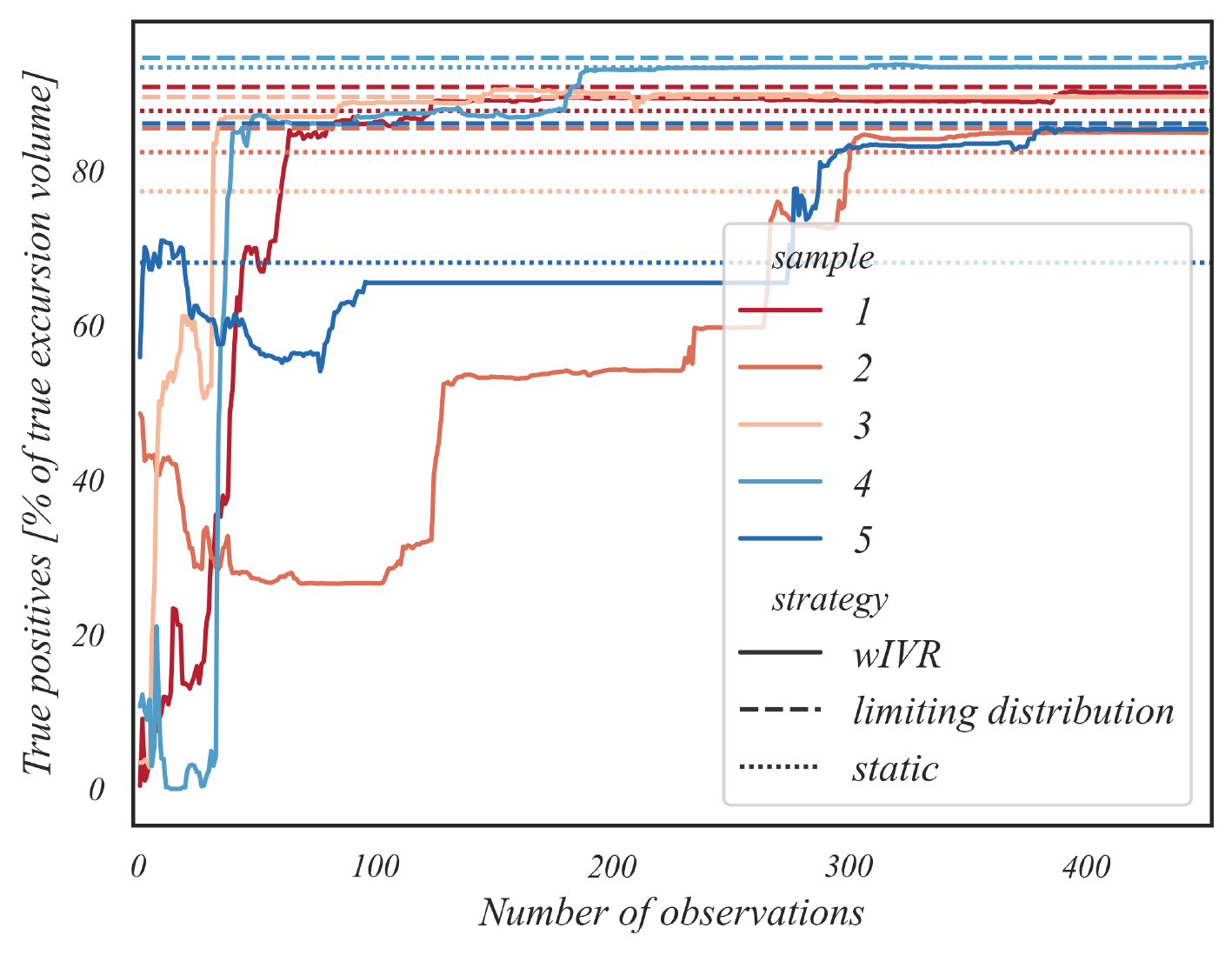}
	\caption{True positives}
\end{subfigure}
\hspace{3mm}%
\begin{subfigure}[b]{0.48\textwidth}
	\includegraphics[width=0.99\linewidth]{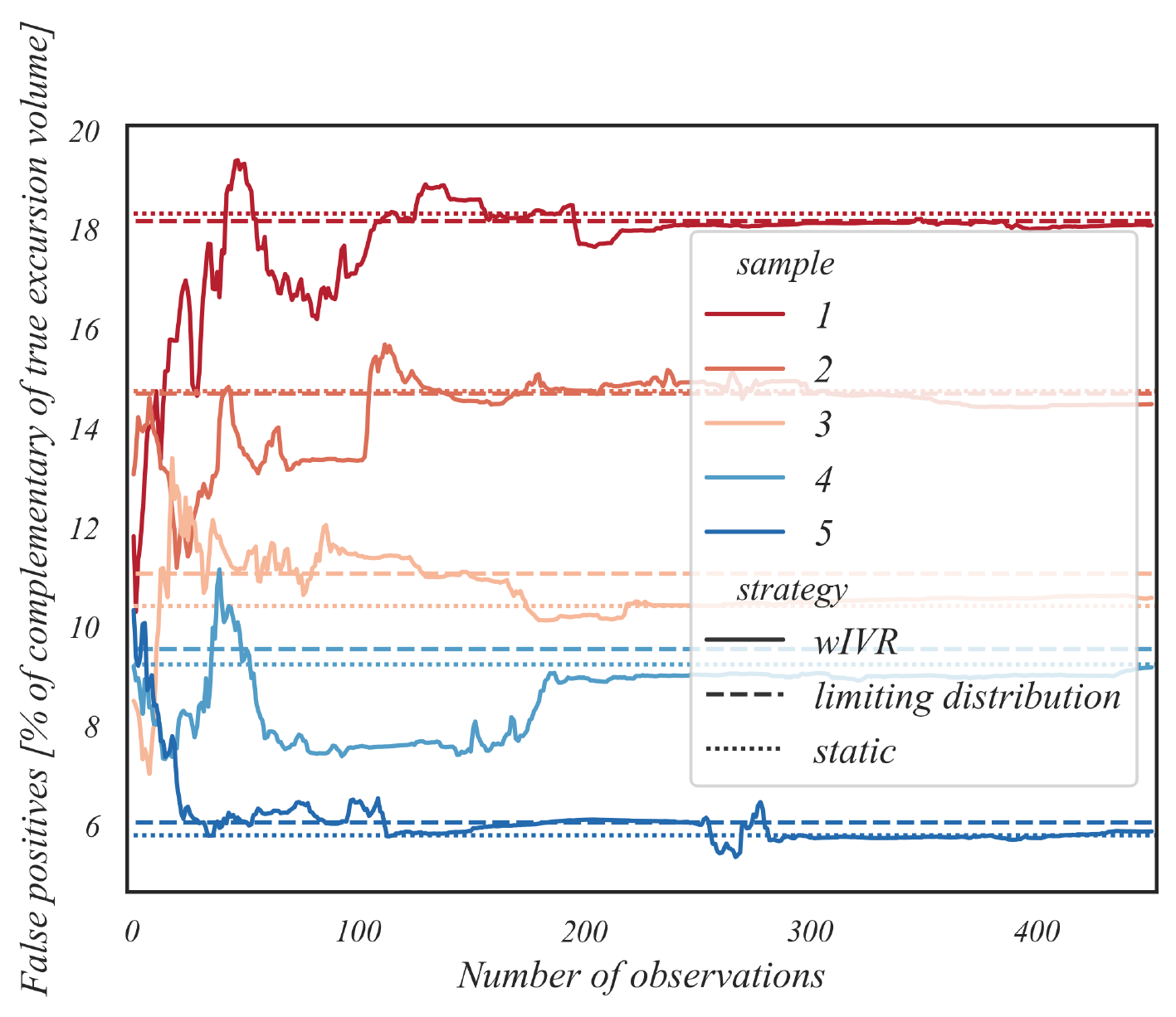}
	\caption{False positives}
\end{subfigure}
  \caption{Evolution of true and false positives for the \textit{large} scenario as a function of the number of observations.}\label{fig:detection_evolution_big}
\end{figure}

\begin{figure}[tbhp]
\begin{subfigure}[b]{0.48\textwidth}
	\includegraphics[width=0.99\linewidth]{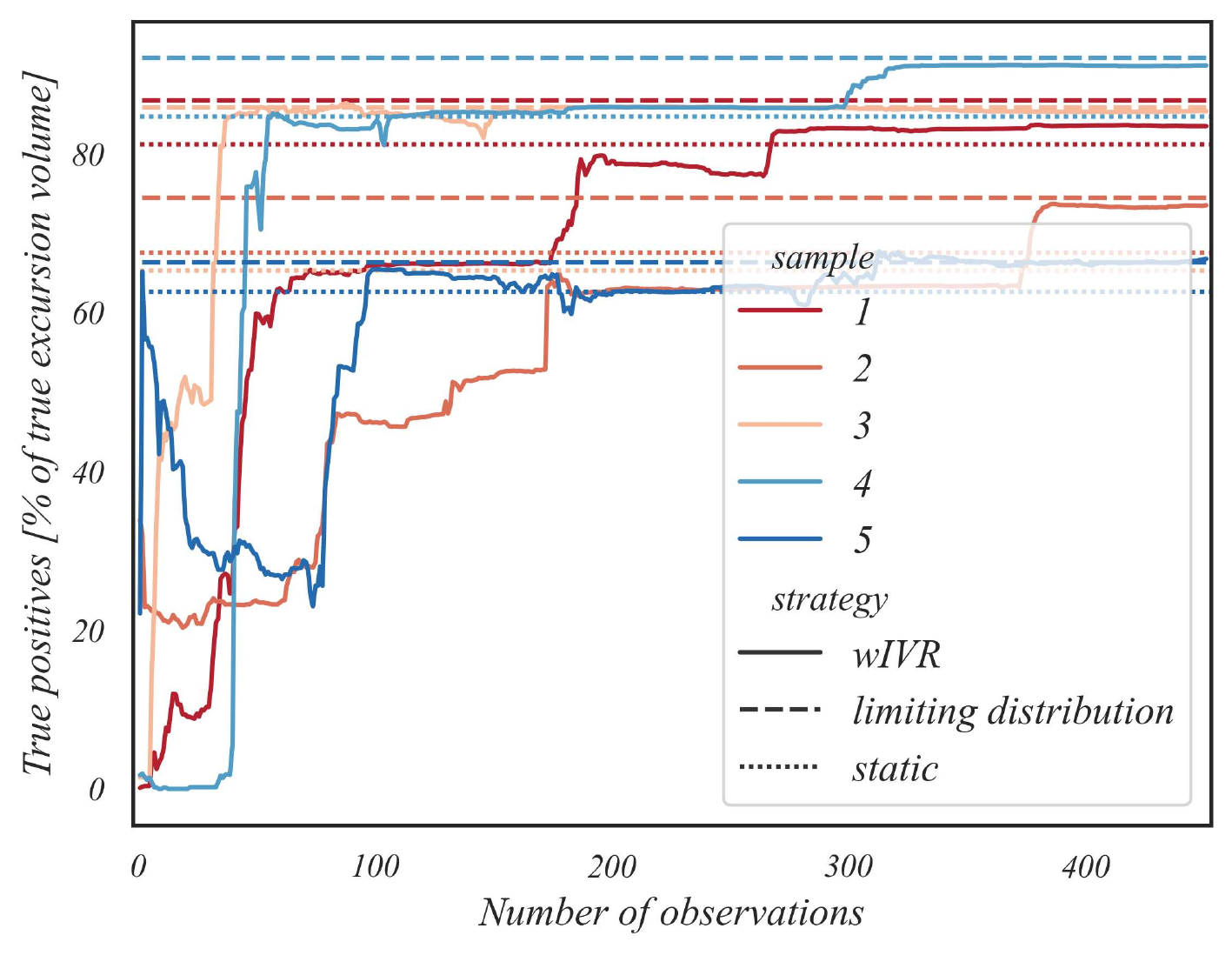}
	\caption{True positives}
\end{subfigure}
\begin{subfigure}[b]{0.48\textwidth}
	\includegraphics[width=0.99\linewidth]{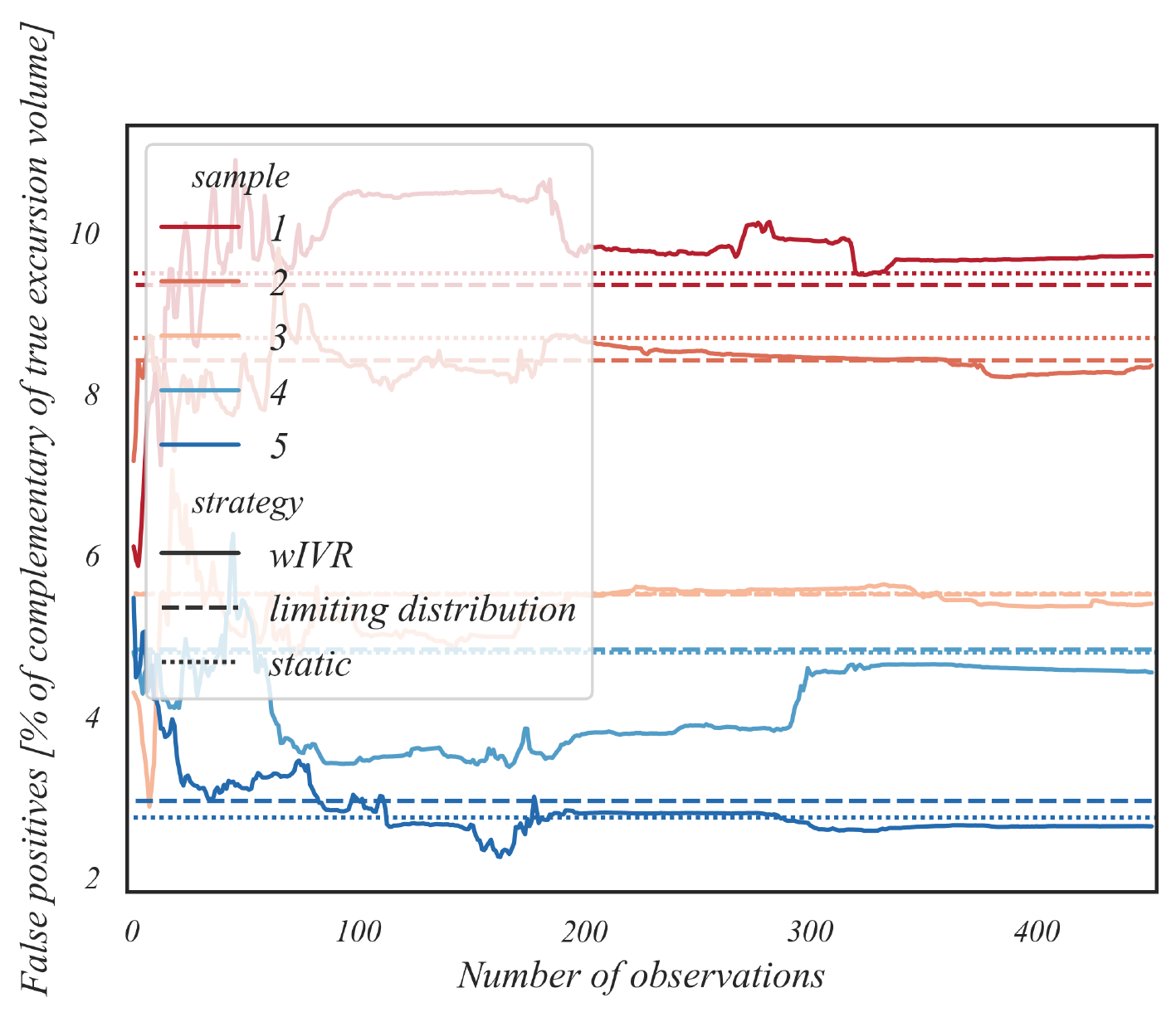}
	\caption{False positives}
\end{subfigure}
  \caption{Evolution of true and false positives for the \textit{small} scenario as a function of the number of observations.}\label{fig:detection_evolution_small}
\end{figure}

Experiments are run by starting at a fixed starting point on the volcano surface, and then sequentially choosing the next observation locations on the volcano surface according to the wIVR criterion. Datapoints are collected one at a time. We here only consider myopic optimization, that is, at each stage, we select the next observation
site $\nextSite$ according to:
\begin{equation*}
    \nextSite = \argmin_{\site\in\candidateSites}
    \textrm{wIVR}^{\stage}(\site),
\end{equation*}
where ties are broken arbitrarily. Here $\candidateSites$ is a set of
candidates among which to pick the next observation location. In our
experiments, we fix $\candidateSites$ to consist of all surface points within a ball of radius $150$ meters around the last observation location. 
Results are summarized in \cref{fig:detection_evolution_big,fig:detection_evolution_small}, which shows the evolution of the fraction of true positives and false positives as a function of the number of observations gathered.

We see that in the \textit{large} scenario (\cref{fig:detection_evolution_big}) the wIVR criterion is able to correctly detect 70 to 80\% of the excursion set (in volume) for each ground truth after 450 observations. For the \textit{small} scenario (\cref{fig:detection_evolution_small}) the amount of true positives reached after 450 observations is similar, though two ground truths are harder to detect.\\

Note that in \cref{fig:detection_evolution_big,fig:detection_evolution_small} the fraction of false negatives is expressed as a percentage of the volume of the complementary of the true excursion set $D \setminus \Gamma^*$. We see that the average percentage of false positives after $450$ observations tends to lie between $5$ and $15\%$, with smaller excursion sets yielding fewer false positives. While the Vorob'ev expectation is not designed to minimize the amount of false positives, there exists \textit{conservative set estimators} \citep{azzimonti_adaptive} that specialize on this task. We identify the extension of such estimators to inverse problems as a promising venue for new research.

In both figures we also plot the fraction of true positives and false positives that result from the data collection plan that was used in \citet{linde}. Here only the situation at the end of the data collection process is shown. We see that for some of the ground truths the wIVR criterion is able to outperform static designs by around 10\%. Note that there are ground truths where it performs similarly to a static design. We believe this is due to the fact that for certain ground truths most of the information about the excursion set can be gathered by spreading the observations across the volcano, which is the case for the static design that also considers where it is practical and safe to measure.

\begin{figure}[h]
\centering
\begin{subfigure}{0.7\textwidth}
	\includegraphics[width=0.99\linewidth]{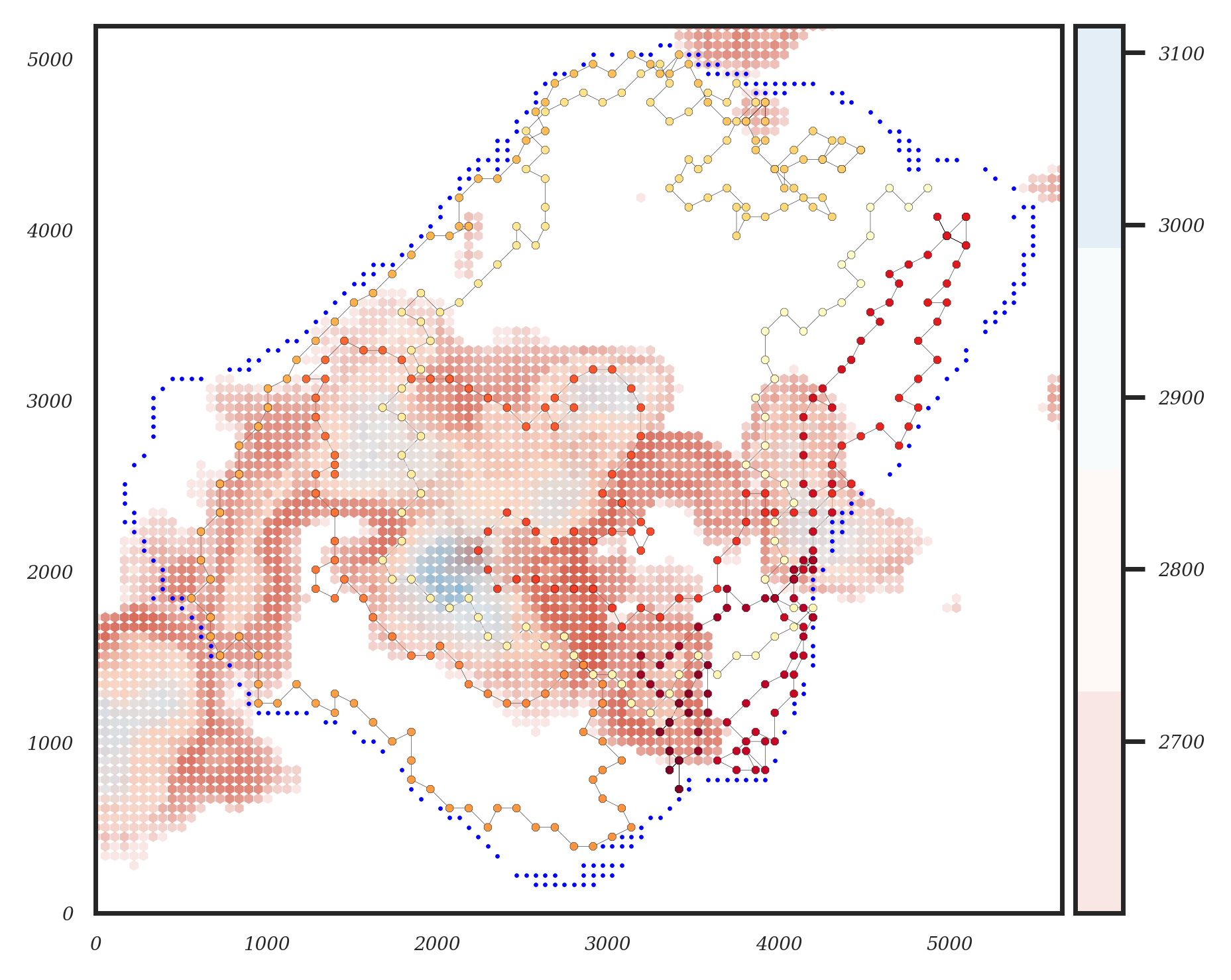}
	\caption{xy projection}
\end{subfigure}\\
\begin{subfigure}{0.49\textwidth}
	\includegraphics[width=0.99\linewidth]{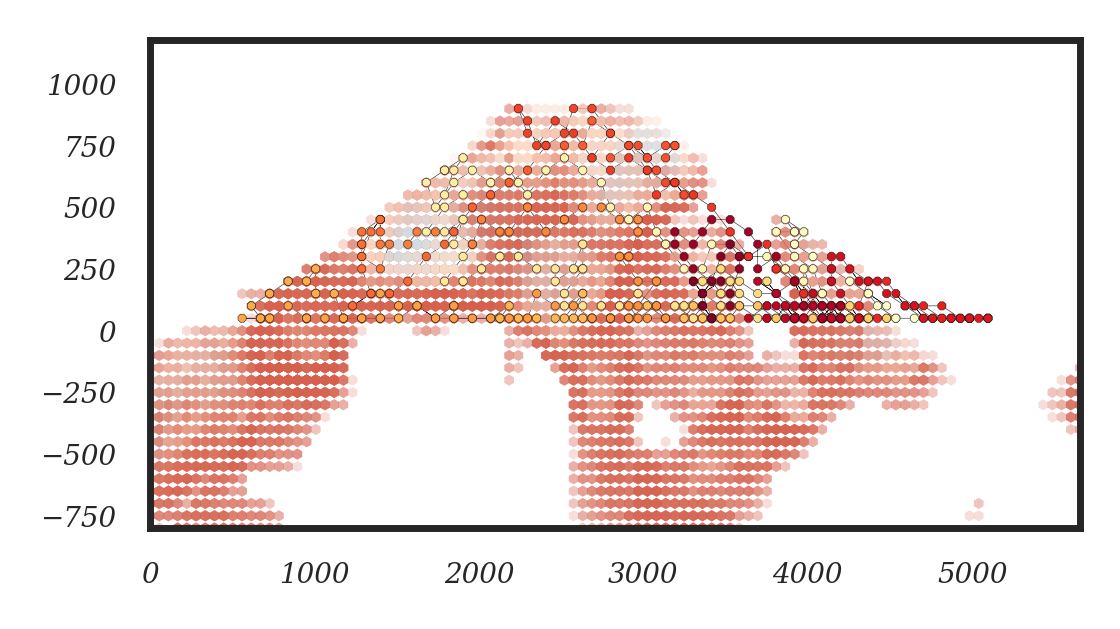}
	\caption{xz projection}
\end{subfigure}
\begin{subfigure}{0.49\textwidth}
	\includegraphics[width=0.99\linewidth]{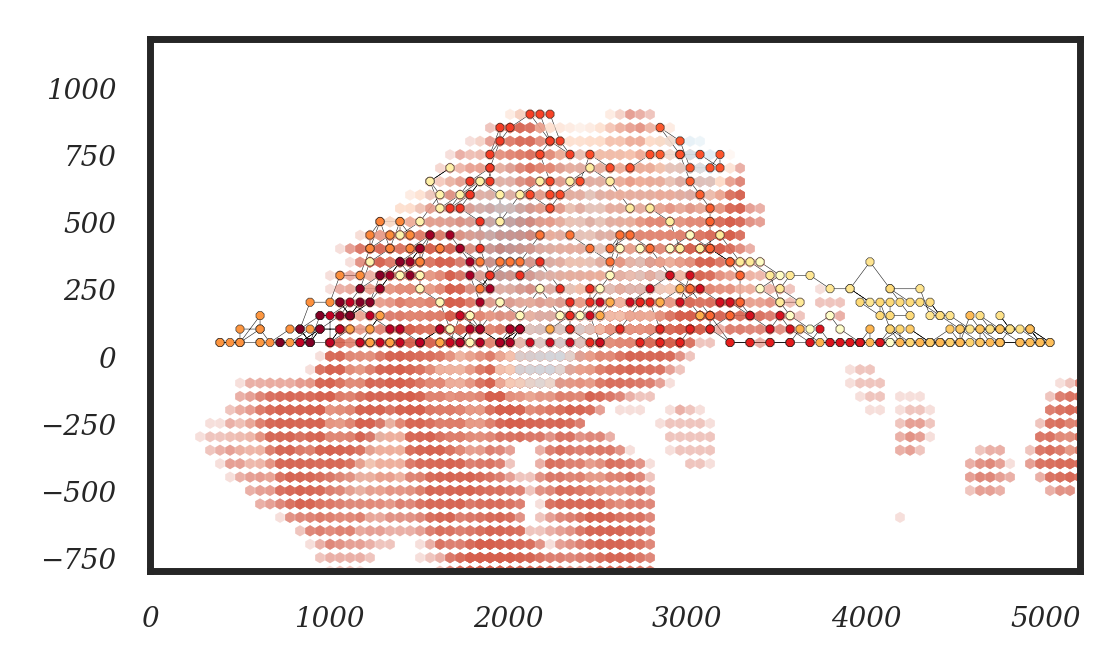}
	\caption{yz projection}
\end{subfigure}
\caption{Projection of the true excursion set (small scenario) and visited locations (wIVR strategy) for the first ground truth. Island boundary is shown in blue. Distances are displayed in [m] and density in [kg/$m^3$].}
\label{fig:excu_slice_visited}
\end{figure}

\textbf{Limiting Distribution:} The dashed horizontal lines in  \cref{fig:detection_evolution_big,fig:detection_evolution_small}  show the detection percentage that can be achieved using the \textit{limiting distribution}. We define the \textit{limiting distribution} as the posterior distribution one were to obtain 
if one had gathered data at all allowed locations (everywhere on the volcano surface). This distribution may be approximated by gathering data at all points of a given (fine grained) discretization of the surface. In general, this is hard to compute since it requires ingestion of a very large amount of data, but thanks to our implicit representation (\cref{sec:implicit}) we can get access to this object, thereby, allowing new forms of uncertainty quantification.\\

In a sense, the limiting distribution represents \textit{the best we can hope for} when covering the volcano with this type of measurements (gravimetric). It gives a measure of the residual uncertainty inherent to the type of observations used
(gravimetric). Indeed, it is known that a given density field is not
identifiable from gravimetric data alone (see \citet{blakely_1995} for example).
Even if gravity data will never allow for a perfect reconstruction of the excursion
set, we can use the limiting distribution to compare the performance of
different sequential design criteria and strategies. It also provides a mean of quantifying the remaining uncertainty under the chosen class of models.
A sensible performance metric
is then the number of observations that a given criterion needs to approach the
minimal level of residual uncertainty which is given by the limiting
distributions.\\

%
As a last remark, we stress that the above results and the corresponding reconstruction qualities are tied to an estimator, in our case the Vorob'ev expectation. If one were to use another estimator for the excursion set, those results could change significantly.\\

\textbf{Posterior Volume Distribution:} Thanks to our extension of the residual kriging algorithm to inverse problems (see \cref{sec:sampling}), we are able to sample from the posterior at the end of the data collection process. This opens new venues for uncertainty quantification in inverse problems. For example, we can use sampling to estimate the posterior distribution of the excursion volume and estimate the residual uncertainty on the size of the excursion set. 

\begin{figure}[h!]
\centering
\begin{subfigure}{0.49\textwidth}	\includegraphics[width=0.99\linewidth,height=0.76\linewidth]{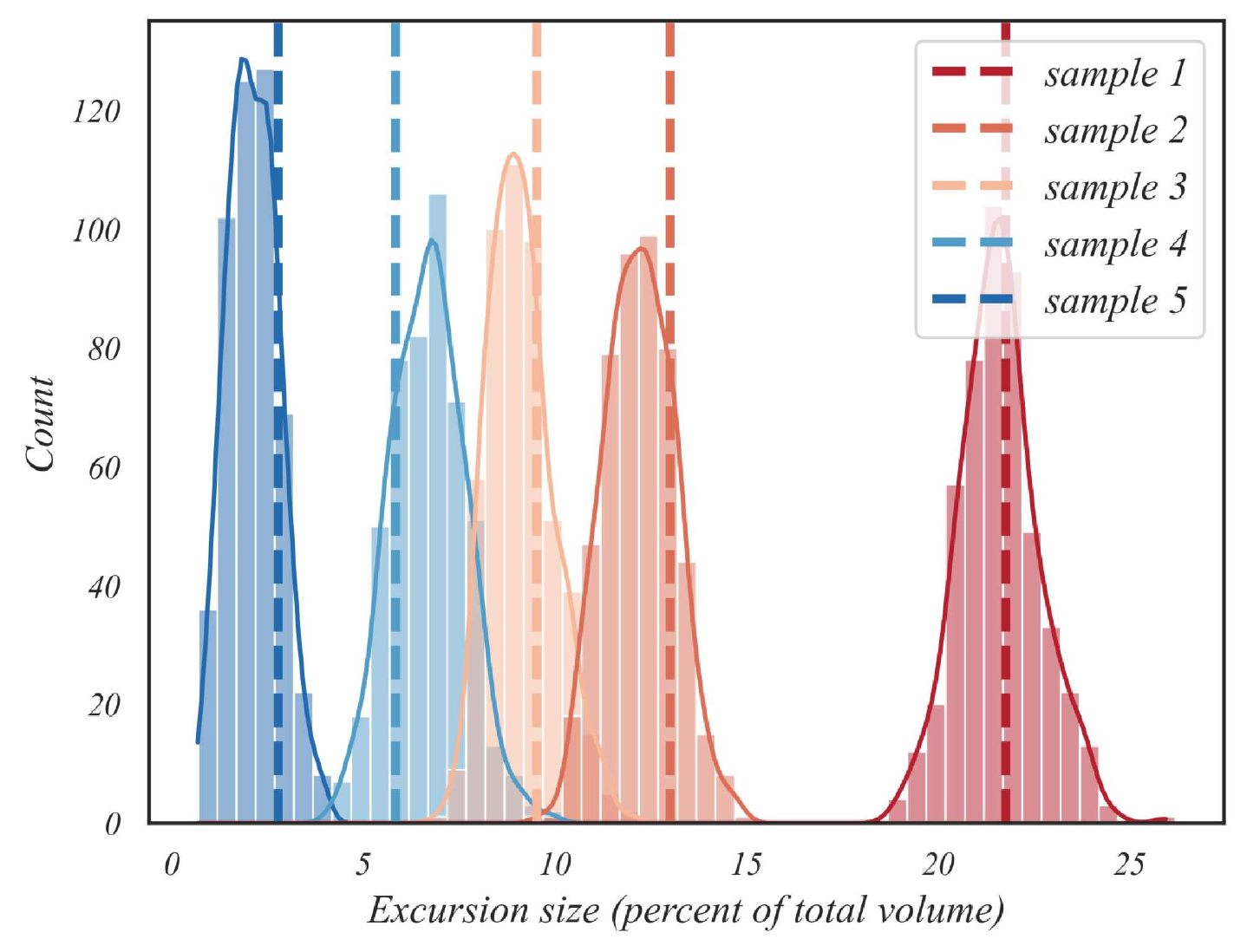}
	\caption{(\textit{large} scenario) threshold: $2500~{[}kg/m^3{]}$}
\end{subfigure}
\begin{subfigure}{0.49\textwidth}	\includegraphics[width=0.99\linewidth,height=0.76\linewidth]{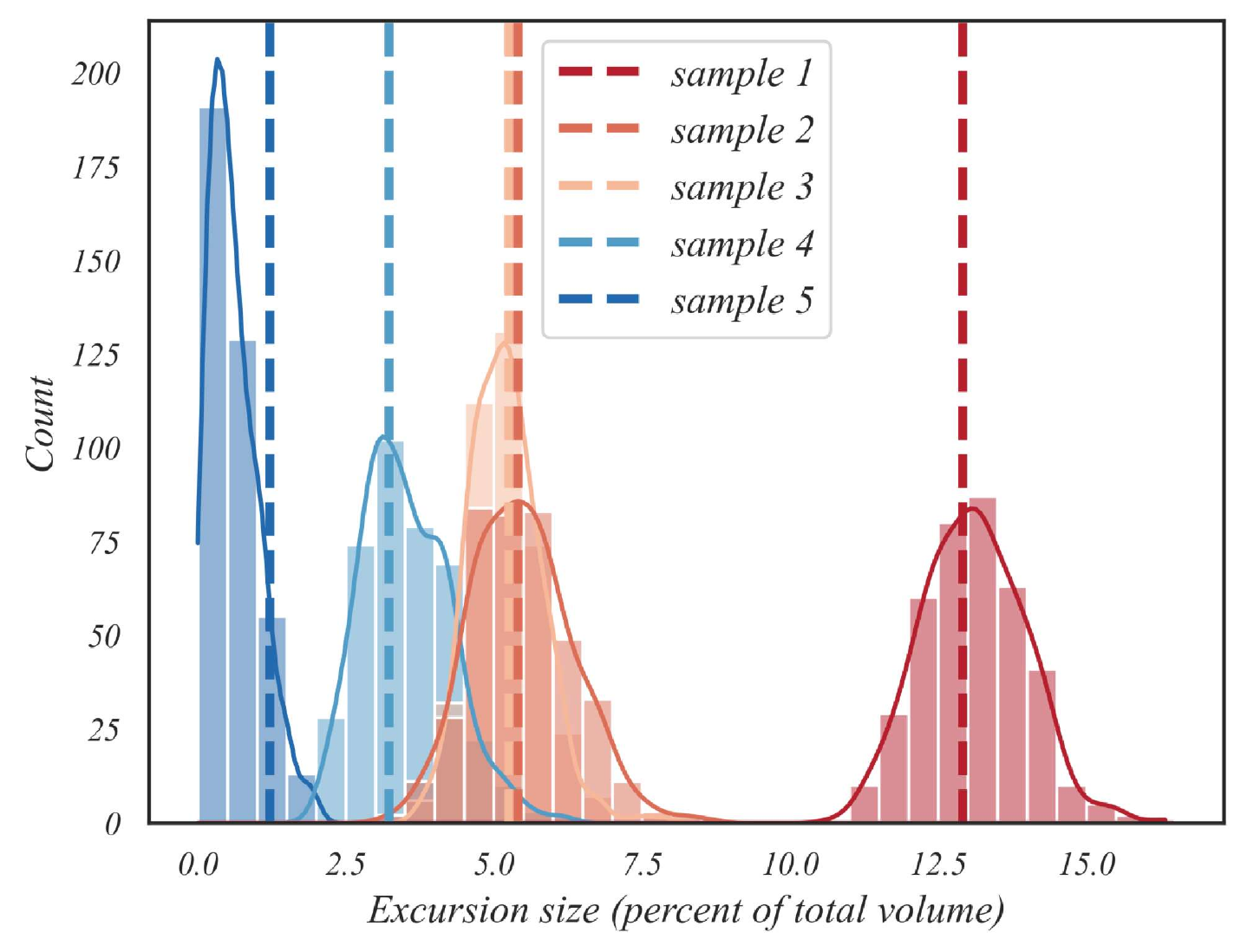}
  	\caption{(\textit{small} scenario) threshold: $2600~{[}kg/m^3{]}$}
\end{subfigure}
  \caption{Empirical posterior distribution (after 450 observations) of the excursion volume for each ground truth. True volumes are denoted by vertical lines.}\label{fig:excu_volume_posterior}
 \end{figure}
 
\cref{fig:excu_volume_posterior} shows the empirical posterior distribution of the excursion volume for each of the ground truths considered in the preceding experiments. When compared to the prior distribution, \cref{fig:excu_volume_scenarios}, one sees that the wIVR criterion is capable of significantly reducing the uncertainty on the excursion volume. This shows that though the location of the excursion set can only be recovered with limited accuracy, as shown in \cref{fig:detection_evolution_big,fig:detection_evolution_small}, the excursion volume can be estimated quite well. This is surprising given that the criterion used (wIVR) is a very crude one and was not designed for that task. On the other hand, there exist  more refined criterion, like the so-called \textit{SUR strategies} (sequential uncertainty reduction) \citep{SUR_2014,SUR_2019}, among which some were specifically engineered to reduce the uncertainty on the excursion volume \citep{SUR_2012}. Even though those criterion are more computationally challenging than the wIVR one, expecially in the considered framework, we identify their application to large Bayesian inverse problem as a promising avenue for future research.

\section{Conclusion and Perspectives}
Leveraging new results about sequential 
disintegrations of Gaussian measures \citep{travelletti_technical}, 
we have introduced 
an implicit almost matrix free representation of the posterior
covariance of a GP and have demonstrated fast update of the posterior
covariance on large grids under general linear functional observations. Our
method allows streamline updating and fast extraction of posterior covariance
information even when the matrices are larger than the available computing memory.
Using our novel implicit representation, we have shown how targeted design
criteria for excursion set recovery may be extended to inverse problems
discretized on large grids. We also demonstrated UQ on such problems using
posterior sampling via residual kriging. Our results suggest that using
the considered design criteria allows reaching close-to-minimal levels of residual
uncertainty using a moderate number of observations and also exhibit significant reduction of uncertainty on the excursion volume. The GP priors used in this work are meant as a proof of concept and future work should address the pitfalls of such priors, such as lack of positiveness of the realisations and lack of expressivity. Other promising research avenues include extension to
multivariate excursions \cite{fossum2020learning} and inclusion of more
sophisticated estimators such as conservative estimates
\cite{azzimonti_adaptive}. On the dynamic programming side, extending the myopic optimization of the criterion to finite 
horizon optimization in order to provide optimized data collection trajectories is an obvious next step which could have significant impact on the geophysics community. Also, including location dependent observation costs such as accessibility 
in the design criterion could help provide more realistic observation plans.

\subsubsection*{Acknowledgements}
This work was funded by the Swiss National Science Foundation (SNF) through project no.~178858.

\bibliography{bibliography}

\begin{thebibliography}{}

\bibitem[Attia et~al., 2018]{opt_des_inverse}
Attia, A., Alexanderian, A., and Saibaba, A. (2018).
\newblock Goal-oriented optimal design of experiments for large-scale
  {B}ayesian linear inverse problems.
\newblock {\em Inverse {P}roblems}, 34.

\bibitem[Azzimonti et~al., 2016]{azzimonti_uq}
Azzimonti, D., Bect, J., Chevalier, C., and Ginsbourger, D. (2016).
\newblock Quantifying uncertainties on excursion sets under a {G}aussian random
  field prior.
\newblock {\em SIAM/ASA Journal on Uncertainty Quantification}, 4(1):850--874.

\bibitem[Azzimonti et~al., 2019]{azzimonti_adaptive}
Azzimonti, D., Ginsbourger, D., Chevalier, C., Bect, J., and Richet, Y. (2019).
\newblock Adaptive design of experiments for conservative estimation of
  excursion sets.
\newblock {\em Technometrics}, 0(0):1--14.

\bibitem[Banerjee and Das~Gupta, 1977]{banerjee}
Banerjee, B. and Das~Gupta, S. (1977).
\newblock Gravitational attraction of a rectangular parallelepiped.
\newblock {\em Geophysics}, 42(5):1053--1055.

\bibitem[Barnes and Watson, 1992]{update_barnes}
Barnes, R.~J. and Watson, A. (1992).
\newblock Efficient updating of kriging estimates and variances.
\newblock {\em Mathematical Geology}, 24(1):129--133.

\bibitem[Bect et~al., 2019]{SUR_2019}
Bect, J., Bachoc, F., and Ginsbourger, D. (2019).
\newblock A supermartingale approach to gaussian process based sequential
  design of experiments.
\newblock {\em Bernoulli}, 25(4A):2883--2919.

\bibitem[Bect et~al., 2012a]{bect2012sequential}
Bect, J., Ginsbourger, D., Li, L., Picheny, V., and Vazquez, E. (2012a).
\newblock Sequential design of computer experiments for the estimation of a
  probability of failure.
\newblock {\em Statistics and Computing}, 22(3):773--793.

\bibitem[Bect et~al., 2012b]{SUR_2012}
Bect, J., Ginsbourger, D., Li, L., Picheny, V., and Vazquez, E. (2012b).
\newblock Sequential design of computer experiments for the estimation of a
  probability of failure.
\newblock {\em Statistics and Computing}, 22(3):773--793.

\bibitem[Blakely, 1995]{blakely_1995}
Blakely, R.~J. (1995).
\newblock {\em Potential Theory in Gravity and Magnetic Applications}.
\newblock Cambridge University Press.

\bibitem[Chevalier et~al., 2014a]{chevalier_2014_fast}
Chevalier, C., Bect, J., Ginsbourger, D., Vazquez, E., Picheny, V., and Richet,
  Y. (2014a).
\newblock Fast parallel kriging-based stepwise uncertainty reduction with
  application to the identification of an excursion set.
\newblock {\em Technometrics}, 56(4):455--465.

\bibitem[Chevalier et~al., 2014b]{SUR_2014}
Chevalier, C., Bect, J., Ginsbourger, D., Vazquez, E., Picheny, V., and Richet,
  Y. (2014b).
\newblock Fast parallel kriging-based stepwise uncertainty reduction with
  application to the identification of an excursion set.
\newblock {\em Technometrics}, 56(4):455--465.

\bibitem[Chevalier et~al., 2015]{foxy}
Chevalier, C., David, G., and Emery, X. (2015).
\newblock Fast update of conditional simulation ensembles.
\newblock {\em Mathematical Geosciences}, 47:771--789.

\bibitem[Chevalier et~al., 2013]{chevalier_uq}
Chevalier, C., Ginsbourger, D., Bect, J., and Molchanov, I. (2013).
\newblock Estimating and quantifying uncertainties on level sets using the
  {V}orob’ev expectation and deviation with {G}aussian process models.
\newblock In {\em mODa 10--Advances in Model-Oriented Design and Analysis},
  pages 35--43. Springer.

\bibitem[Chevalier et~al., 2014c]{update_chevalier}
Chevalier, C., Ginsbourger, D., and Emery, X. (2014c).
\newblock Corrected kriging update formulae for batch-sequential data
  assimilation.
\newblock In Pardo-Ig\'{u}zquiza, E., Guardiola-Albert, C., Heredia, J.,
  Moreno-Merino, L., Dur\'{a}n, J., and Vargas-Guzm\'{a}n, J., editors, {\em
  Mathematics of Planet Earth. Lecture Notes in Earth System Sciences}.
  Springer, Berlin, Heidelberg.

\bibitem[Chil\`{e}s and Delfiner, 2012]{chiles_delfiner}
Chil\`{e}s, J.-P. and Delfiner, P. (2012).
\newblock {\em Geostatistics: {M}odeling {S}patial {U}ncertainty}.
\newblock Wiley, 2 edition.

\bibitem[Dashti and Stuart, 2016]{stuart_dashti}
Dashti, M. and Stuart, A.~M. (2016).
\newblock The {B}ayesian approach to inverse problems.
\newblock {\em Handbook of Uncertainty Quantification}, pages 1--118.

\bibitem[de~Fouquet, 1994]{de_fouquet_cond_kriging}
de~Fouquet, C. (1994).
\newblock Reminders on the conditioning kriging.
\newblock In Armstrong, M. and Dowd, P.~A., editors, {\em Geostatistical
  Simulations}, pages 131--145, Dordrecht. Springer Netherlands.

\bibitem[Emery, 2009]{update_emery}
Emery, X. (2009).
\newblock The kriging update equations and their application to the selection
  of neighboring data.
\newblock {\em Computational Geosciences}, 13(3):269--280.

\bibitem[Fossum et~al., 2021]{fossum2020learning}
Fossum, T.~O., Travelletti, C., Eidsvik, J., Ginsbourger, D., and Rajan, K.
  (2021).
\newblock {Learning excursion sets of vector-valued Gaussian random fields for
  autonomous ocean sampling}.
\newblock {\em The Annals of Applied Statistics}, 15(2):597 -- 618.

\bibitem[Gao et~al., 1996]{update_gao}
Gao, H., Wang, J., and Zhao, P. (1996).
\newblock The updated kriging variance and optimal sample design.
\newblock {\em Mathematical Geology}, 28(3):295--313.

\bibitem[Hairer et~al., 2005]{hairer_stuart}
Hairer, M., Stuart, A.~M., Voss, J., and Wiberg, P. (2005).
\newblock {Analysis of SPDEs arising in path sampling. Part I: The Gaussian
  case}.
\newblock {\em Communications in Mathematical Sciences}, 3(4):587 -- 603.

\bibitem[Hamelijnck et~al., 2021]{hamelijnck2021spatiotemporal}
Hamelijnck, O., Wilkinson, W.~J., Loppi, N.~A., Solin, A., and Damoulas, T.
  (2021).
\newblock Spatio-temporal variational gaussian processes.
\newblock In Beygelzimer, A., Dauphin, Y., Liang, P., and Vaughan, J.~W.,
  editors, {\em Advances in Neural Information Processing Systems}, volume~34.
  Curran Associates, Inc.

\bibitem[Hansen, 2010]{hansen_discrete}
Hansen, P.~C. (2010).
\newblock {\em Discrete Inverse Problems}.
\newblock Society for Industrial and Applied Mathematics.

\bibitem[Hendriks et~al., 2018]{hendriks2018evaluating}
Hendriks, J.~N., Jidling, C., Wills, A., and Schön, T.~B. (2018).
\newblock Evaluating the squared-exponential covariance function in gaussian
  processes with integral observations.

\bibitem[Hensman et~al., 2013]{hensman_big}
Hensman, J., Fusi, N., and Lawrence, N.~D. (2013).
\newblock Gaussian processes for big data.
\newblock In Nicholson, A.~E. and Smyth, P., editors, {\em Proceedings of the
  Twenty-Ninth Conference on Uncertainty in Artificial Intelligence, {UAI}
  2013, Bellevue, WA, USA, August 11-15, 2013}. {AUAI} Press.

\bibitem[Jidling et~al., 2019]{jidling2019deep}
Jidling, C., Hendriks, J., Schön, T.~B., and Wills, A. (2019).
\newblock Deep kernel learning for integral measurements.

\bibitem[Jidling et~al., 2017]{jidling2017linearly}
Jidling, C., Wahlstr\"{o}m, N., Wills, A., and Sch\"{o}n, T.~B. (2017).
\newblock Linearly constrained gaussian processes.
\newblock In Guyon, I., Luxburg, U.~V., Bengio, S., Wallach, H., Fergus, R.,
  Vishwanathan, S., and Garnett, R., editors, {\em Advances in Neural
  Information Processing Systems}, volume~30. Curran Associates, Inc.

\bibitem[Kitanidis, 2015]{kitanidis_large_kalman}
Kitanidis, P.~K. (2015).
\newblock Compressed state {K}alman filter for large systems.
\newblock {\em Advances in Water Resources}, 76:120 -- 126.

\bibitem[Klebanov et~al., 2020]{klebanov_sullivan}
Klebanov, I., Sprungk, B., and Sullivan, T.~J. (2020).
\newblock The linear conditional expectation in hilbert space.

\bibitem[Linde et~al., 2014]{linde}
Linde, N., Baron, L., Ricci, T., Finizola, A., Revil, A., Muccini, F., Cocchi,
  L., and Carmisciano, C. (2014).
\newblock 3-{D} density structure and geological evolution of {S}tromboli
  volcano ({A}eolian {I}slands, {I}taly) inferred from land-based and
  sea-surface gravity data.
\newblock {\em Journal of {V}olcanology and {G}eothermal {R}esearch},
  273:58--69.

\bibitem[Linde et~al., 2017]{linde_2017}
Linde, N., Ricci, t., Baron, L., A., S., and G., B. (2017).
\newblock {The 3-D structure of the Somma-Vesuvius volcanic complex (Italy)
  inferred from new and historic gravimetric data}.
\newblock {\em Scientific Reports}, 7(1):8434.

\bibitem[Mandel, 2006]{mandel_efficient_ensemble_kalman}
Mandel, J. (2006).
\newblock Efficient implementation of the ensemble {K}alman filter.
\newblock Technical Report 231, University of Colorado at Denver and Health
  Sciences Center.

\bibitem[Mandelbaum, 1984]{mandelbaum}
Mandelbaum, A. (1984).
\newblock Linear estimators and measurable linear transformations on a hilbert
  space.
\newblock {\em Zeitschrift f\"ur Wahrscheinlichkeitstheorie und Verwandte
  Gebiete}, 65(3):385–397.

\bibitem[Mantoglou and Wilson, 1982]{turning_bands}
Mantoglou, A. and Wilson, J.~L. (1982).
\newblock The turning bands method for simulation of random fields using line
  generation by a spectral method.
\newblock {\em Water Resources Research}, 18(5):1379--1394.

\bibitem[Molchanov, 2005]{molchanov2005theory}
Molchanov, I. (2005).
\newblock {\em Theory of random sets}.
\newblock Springer.

\bibitem[{Montesinos} et~al., 2006]{montesinos}
{Montesinos}, F.~G., {Arnoso}, J., {Benavent}, M., and {Vieira}, R. (2006).
\newblock {The crustal structure of El Hierro (Canary Islands) from 3-D gravity
  inversion}.
\newblock {\em Journal of Volcanology and Geothermal Research},
  150(1-3):283--299.

\bibitem[Owhadi and Scovel, 2015]{owhadi_scovel}
Owhadi, H. and Scovel, C. (2015).
\newblock Conditioning gaussian measure on hilbert space.

\bibitem[Park and Baek, 2001]{park_concentration}
Park, J.-S. and Baek, J. (2001).
\newblock Efficient computation of maximum likelihood estimators in a spatial
  linear model with power exponential covariogram.
\newblock {\em Computers \& Geosciences}, 27(1):1--7.

\bibitem[Paszke et~al., 2019]{pytorch}
Paszke, A., Gross, S., Massa, F., Lerer, A., Bradbury, J., Chanan, G., Killeen,
  T., Lin, Z., Gimelshein, N., Antiga, L., Desmaison, A., Kopf, A., Yang, E.,
  DeVito, Z., Raison, M., Tejani, A., Chilamkurthy, S., Steiner, B., Fang, L.,
  Bai, J., and Chintala, S. (2019).
\newblock Pytorch: An imperative style, high-performance deep learning library.
\newblock In Wallach, H., Larochelle, H., Beygelzimer, A., d\textquotesingle
  Alch\'{e}-Buc, F., Fox, E., and Garnett, R., editors, {\em Advances in Neural
  Information Processing Systems 32}, pages 8024--8035. Curran Associates, Inc.

\bibitem[Picheny et~al., 2010]{weighted_imse}
Picheny, V., Ginsbourger, D., Roustant, O., Haftka, R., and Kim, N. (2010).
\newblock Adaptive designs of experiments for accurate approximation of a
  target region.
\newblock {\em Journal of Mechanical Design}, 132:071008.

\bibitem[Rajput and Cambanis, 1972]{rajput_gp_vs_measures}
Rajput, B.~S. and Cambanis, S. (1972).
\newblock {G}aussian processes and {G}aussian measures.
\newblock {\em Ann. Math. Statist.}, 43(6):1944--1952.

\bibitem[Rasmussen and Williams, 2006]{rasmussen_williams}
Rasmussen, C.~E. and Williams, C. K.~I. (2006).
\newblock {\em {G}aussian Processes for Machine Learning}.
\newblock The MIT Press.

\bibitem[Represas et~al., 2012]{represas}
Represas, P., Catal\~ao, J.~a., Montesinos, F.~G., Madeira, J., Mata, J.~a.,
  Antunes, C., and Moreira, M. (2012).
\newblock {Constraints on the structure of Maio Island (Cape Verde) by a
  three-dimensional gravity model: imaging partially exhumed magma chambers}.
\newblock {\em Geophysical Journal International}, 190(2):931--940.

\bibitem[Robbins, 1944]{robbins}
Robbins, H.~E. (1944).
\newblock {On the Measure of a Random Set}.
\newblock {\em The Annals of Mathematical Statistics}, 15(1):70 -- 74.

\bibitem[S{\"a}rkk{\"a}, 2011]{sarkka2011linear}
S{\"a}rkk{\"a}, S. (2011).
\newblock Linear operators and stochastic partial differential equations in
  {G}aussian process regression.
\newblock In {\em International Conference on Artificial Neural Networks},
  pages 151--158. Springer.

\bibitem[S{\"{a}}rkk{\"{a}} et~al., 2013]{sarkka_2013}
S{\"{a}}rkk{\"{a}}, S., Solin, A., and Hartikainen, J. (2013).
\newblock Spatiotemporal learning via infinite-dimensional bayesian filtering
  and smoothing: {A} look at gaussian process regression through kalman
  filtering.
\newblock {\em {IEEE} Signal Process. Mag.}, 30(4):51--61.

\bibitem[Solak et~al., 2003]{solak2003derivative}
Solak, E., Murray-Smith, R., Leithead, W.~E., Leith, D.~J., and Rasmussen,
  C.~E. (2003).
\newblock Derivative observations in {G}aussian process models of dynamic
  systems.
\newblock In {\em Advances in neural information processing systems}, pages
  1057--1064.

\bibitem[Solin et~al., 2015]{solin_2015}
Solin, A., Kok, M., Wahlström, N., Schön, T., and Särkkä, S. (2015).
\newblock Modeling and interpolation of the ambient magnetic field by gaussian
  processes.
\newblock {\em IEEE Transactions on Robotics}, PP.

\bibitem[Stuart, 2010]{stuart_2010}
Stuart, A.~M. (2010).
\newblock Inverse problems: A {B}ayesian perspective.
\newblock {\em Acta Numerica}, 19:451--559.

\bibitem[Tarantola and Valette, 1982]{tarantola}
Tarantola, A. and Valette, B. (1982).
\newblock Generalized nonlinear inverse problems solved using the least squares
  criterion.
\newblock {\em Reviews of Geophysics}, 20(2):219--232.

\bibitem[Tarieladze and Vakhania, 2007]{TARIELADZE2007851}
Tarieladze, V. and Vakhania, N. (2007).
\newblock Disintegration of {G}aussian measures and average-case optimal
  algorithms.
\newblock {\em Journal of Complexity}, 23(4):851 -- 866.
\newblock Festschrift for the 60th Birthday of Henryk Woźniakowski.

\bibitem[Travelletti and Ginsbourger, 2022]{travelletti_technical}
Travelletti, C. and Ginsbourger, D. (2022).
\newblock Disintegration of gaussian measures for sequential assimilation of
  linear operator data.
\newblock {\em arXiv}, abs/2207.13581.

\bibitem[Wagner et~al., 2021]{WAGNER2021110141}
Wagner, P.-R., Marelli, S., and Sudret, B. (2021).
\newblock Bayesian model inversion using stochastic spectral embedding.
\newblock {\em Journal of Computational Physics}, 436:110141.

\bibitem[Wang et~al., 2019]{exact_million}
Wang, K., Pleiss, G., Gardner, J., Tyree, S., Weinberger, K.~Q., and Wilson,
  A.~G. (2019).
\newblock Exact {G}aussian processes on a million data points.
\newblock In Wallach, H., Larochelle, H., Beygelzimer, A., d'~Alch\'{e}-Buc,
  F., Fox, E., and Garnett, R., editors, {\em Advances in {N}eural
  {I}nformation {P}rocessing {S}ystems 32}, pages 14622--14632. Curran
  Associates, Inc.

\bibitem[Williams, 1991]{williams}
Williams, D. (1991).
\newblock {\em Probability with Martingales}.
\newblock Cambridge University Press.

\bibitem[Wilson et~al., 2020]{wilson_large_sampling}
Wilson, J.~T., Borovitskiy, V., Terenin, A., Mostowsky, P., and Deisenroth, M.
  (2020).
\newblock Efficiently sampling functions from {G}aussian process posteriors.
\newblock {\em ArXiv}, abs/2002.09309.

\end{thebibliography}

\newpage
\appendix

\section{Forward operator for Gravimetric Inversion}\label{sec:grav_op}
Given some subsurface density $\density:\domain\rightarrow\mathbb{R}$ inside 
a domain $D\subset \mathbb{R}^3$ and some location
$\site$ outside the domain, the vertical component of the gravitational field at
$\site$ is given by:
\begin{align}
    \mathfrak{\fwd}_{\site}\left[\density\right] &= \int_{\domain} \density(x)
    g(x, \site) dx,
\end{align}
with Green kernel
\begin{equation}
    g(x, \site) = \frac{x^{(3)} - \site^{(3)}}{\Vert x - \site \Vert ^3},
\end{equation}
where $x^{(3)}$ denotes the vertical component of $x$.

We discretize the domain $D$ into $m$ identic cubic cells $D=\cup_{i=1}^m
D_i$ with centroids $\predpts=\left(X_1, \dotsc, X_{\dimpred}\right)$ and assume 
the mass density to be constant over each cell, so the field $\density$ may be
approximated by the vector $\density_{\predpts}$.
The vertical component of the gravitational field at $\site$ is
then given by:
\small
\begin{equation*}
    \int_{\cup_{i=1}^m D_i} g(x, s)
    \rho(x) dx \approx \sum_{i=1}^m \Bigg( \int_{D_i} g(x, s) dx\Bigg)
    \density_{X_i}:=\fwd_{\site}\density_{\predpts}.
\end{equation*}
\normalsize

Integrals of Green kernels over cuboids may be computed using the \textit{Banerjee formula} \citep{banerjee}.

\begin{theorem*}[Banerjee]
The vertical gravity field at point $(x_0, y_0, z_0)$ generated by a prism with corners $(x_h, x_l$, $y_h, y_l, ...)$ of uniform mass density $\rho$ is given by:
\begin{align*}
g_z &= \frac{1}{2} \gamma_N \rho \Bigg[ x~\log \Big( \frac{\sqrt{x^2 + y^2 + z^2} + y}{\sqrt{x^2 + y^2 + z^2} - y} \Big)\\
&\phantom{\frac{1}{2} G \rho \Bigg[}
+ y~\log \Big( \frac{\sqrt{x^2 + y^2 + z^2} + x}{\sqrt{x^2 + y^2 + z^2} - x} \Big)\\
&\phantom{\frac{1}{2} G \rho \Bigg[}
-2z~\arctan \Big( \frac{xy}{z \sqrt{x^2 + y^2 + z^2}} \Big)
\Bigg] 
\Bigg\rvert_{x_l - x_0}^{x_h - x_0}\Bigg\rvert_{y_l - y_0}^{y_h - y_0}\Bigg\rvert_{z_l - z_0}^{z_h - z_0}
\end{align*}
\end{theorem*}

\section{Bayesian Inversion and Bayesian Set Estimation}\label{sec:bayesian_set_estimation}
Given a generic Bayesian linear inverse problem with unknown function $\rho:D\rightarrow \mathbb{R}$ and prior $Z$, there exists several approaches to approximate the excursion set $\trueExcuSet = \lbrace x \in \domain :
\density\left(x\right)\geq\thresh\rbrace$ using the posterior. For example, a naive estimate for $\trueExcuSet$ may be obtained using the  \textit{plug-in estimator}:
\begin{equation*}
    \hat{\excuSet}_{\text{plug-in}} := \lbrace x \in \domain:~\tilde{m}_x \geq \thresh\rbrace,
\end{equation*}
where $\tilde{m}_x$ denotes the posterior mean function of the GP prior. In this work, we will focus on recently developed more sophisticated approaches to Bayesian excursion set estimation \citep{azzimonti_uq,chevalier_uq} based on the theory of random sets \citep{molchanov2005theory}. We here briefly recall some theory taken from the aforementioned source.\\

In the following, let $\tilde{Z}$ denote a random field on $\domain$ that is distributed according to the posterior distribution. Then, the posterior distribution of the field gives rise to a \textit{random closed set} (RACS):
\begin{equation}
    \excuSet := \lbrace x \in \domain:~\tilde{Z}_x \geq \thresh\rbrace.
\end{equation}
One can then consider the probability for any point in the domain to belong to that random set. This is captured by the \textbf{coverage function}:
\begin{align*}
    p_{\excuSet}:&\domain\rightarrow [0,1]\\
    &x\mapsto \mathbb{P}\left[x\in\excuSet\right].
\end{align*}
The coverage function allows us to define a parametric family of set estimates for $\excuSet$, the \textbf{Vorob'ev quantiles}:
\begin{equation}
    Q_{\alpha} := \lbrace x \in \domain : p_{\excuSet}(x) \geq \alpha \rbrace.
\end{equation}
The family of quantiles $Q_{\alpha}$ gives us a way to estimate $\excuSet$ by controlling the (pointwise) probability $\alpha$ that the members of our estimate lie in $\excuSet$. There exists several approaches for choosing $\alpha$. One could for example choose it so as to produce conservative estimates of the excursion set \citep{azzimonti_adaptive}. Another approach is to choose it such that the volume of the resulting quantile is equal to the expected volume of the excursion set. This gives rise to the \textbf{Vorob'ev expectation}.

\begin{definition}{(Vorob'ev Expectation)}
The Vorob'ev expectation is the quantile $Q_{\alpha_V}$ with threshold $\alpha_V$ chosen such that
\begin{equation*}
    \mu(Q_{\alpha}) \leq \mathbb{E}[\vol(\Gamma)] \leq \mu(Q_{\alpha_V}),~ \forall \alpha > \alpha_V,
\end{equation*}
where $\vol(\cdot)$ denotes the volume under the Lebesgue measure on $\mathbb{R}^d$.
\end{definition}
Note that the expected excursion volume may be computed using Robbins's theorem, which states that under suitable conditions:
\begin{equation*}
    \bar{V}_{\excuSet} := \mathbb{E}[\lambda(\excuSet)] = \int_{\domain} p_{\Gamma}(x) dx.
\end{equation*}
We refer the reader to \citet{robbins} and \citet{molchanov2005theory} for more details.

\begin{figure}[h]
\centering
\subfloat{\includegraphics[scale=0.45]{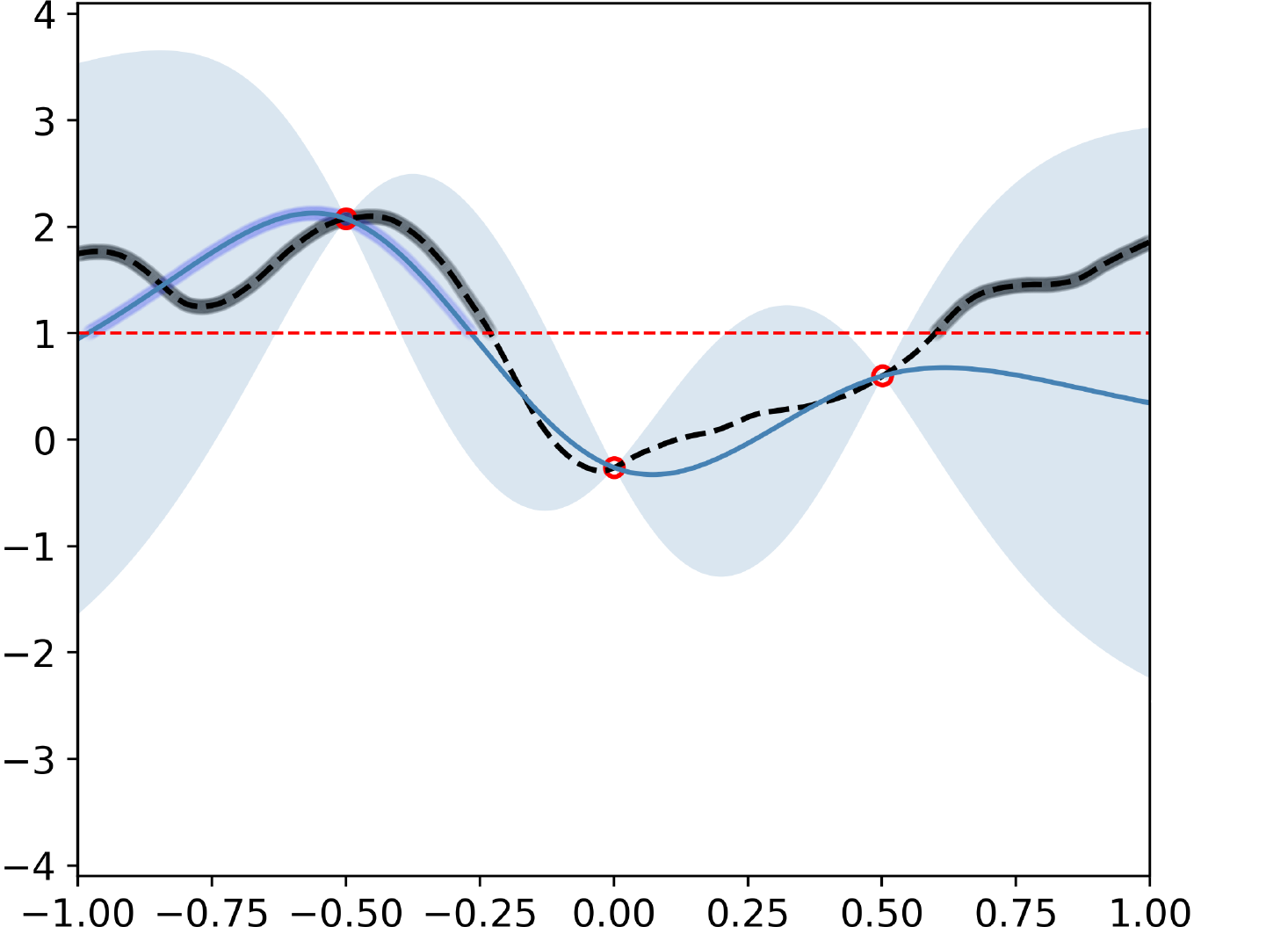}}
\subfloat{\includegraphics[scale=0.45]{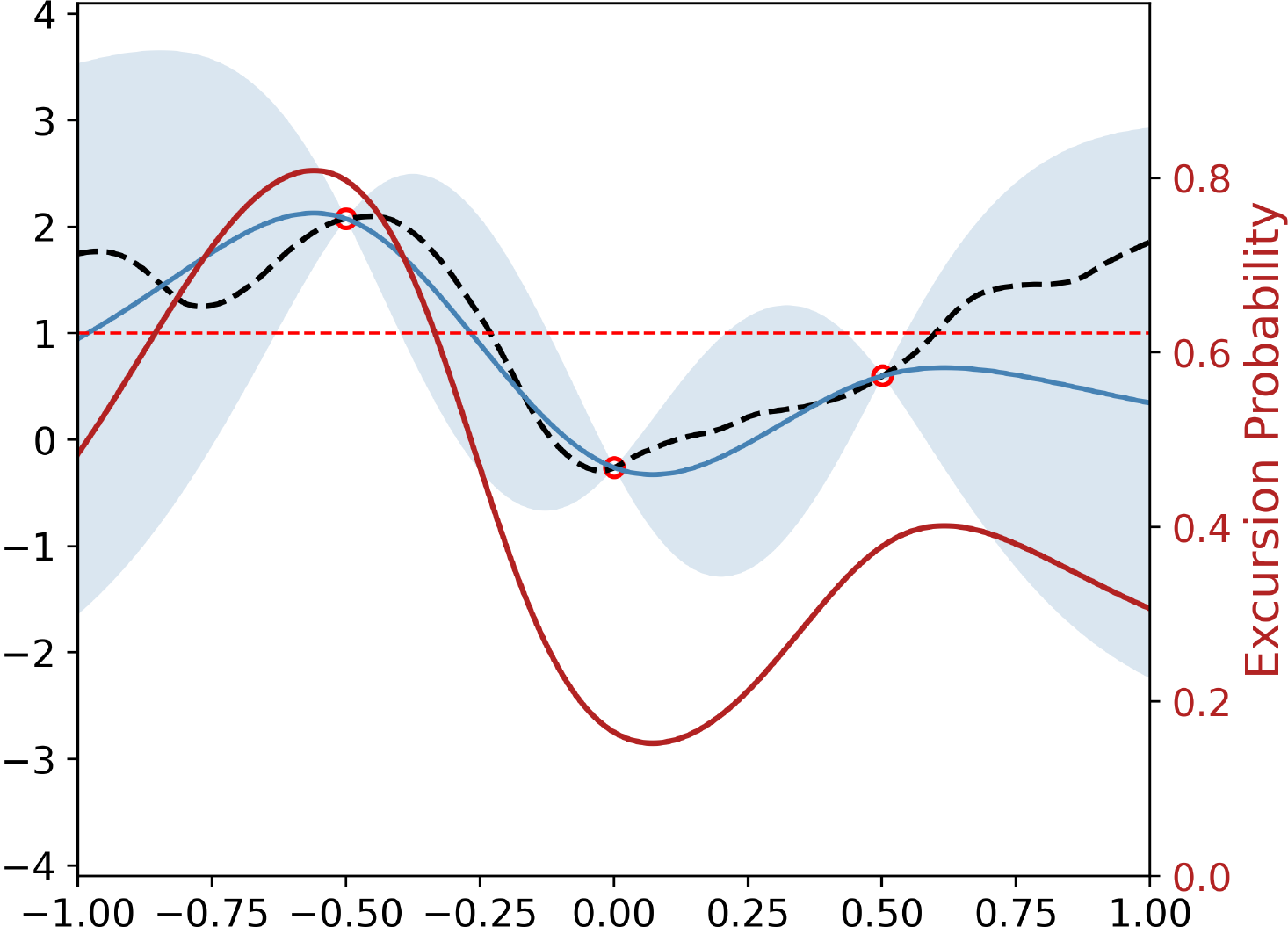}}\\
\subfloat{\includegraphics[scale=0.45]{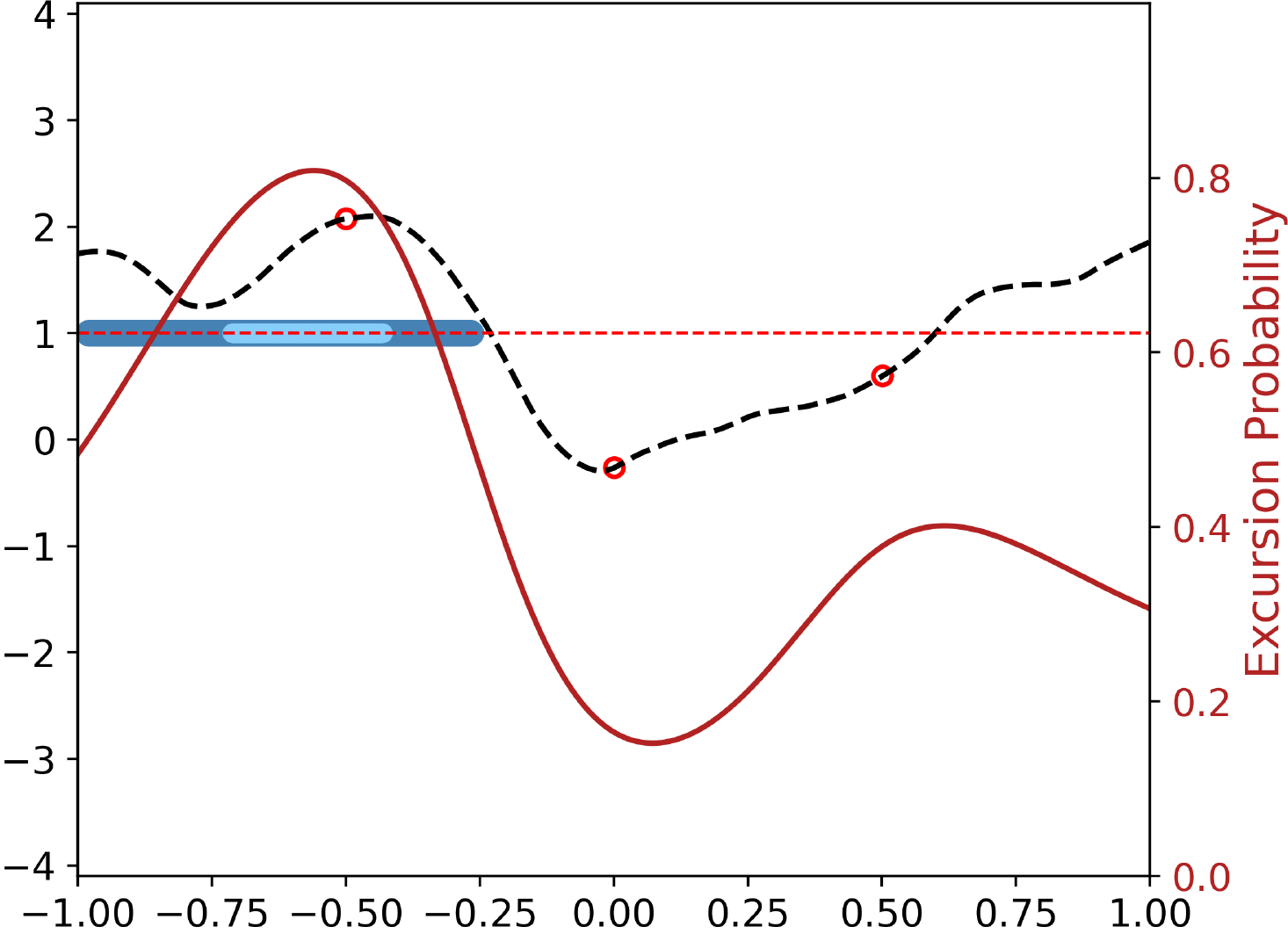}}
\subfloat{\includegraphics[scale=0.45]{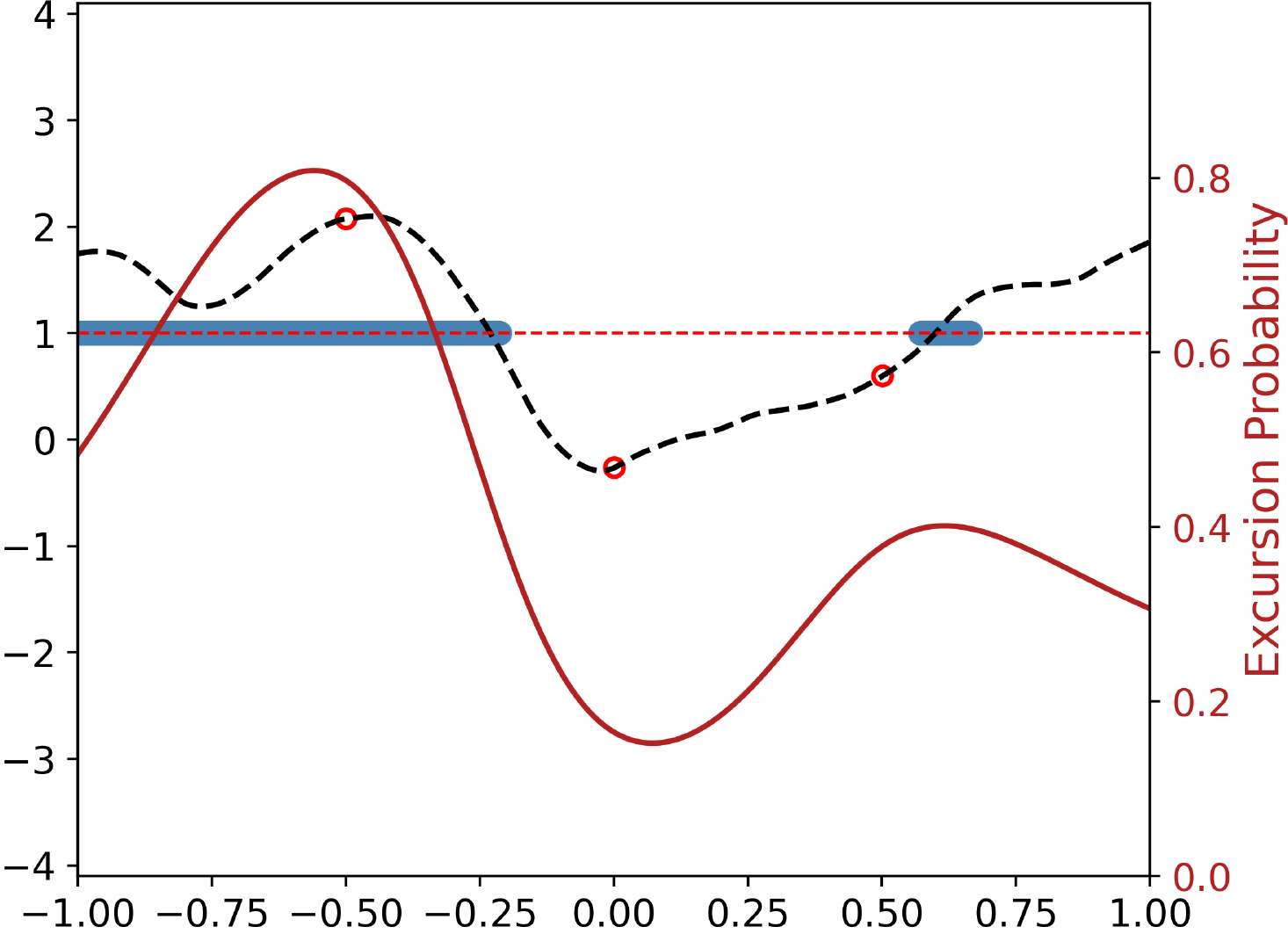}}
\caption{Bayesian set estimation for one-dimensional example. Excursion threshold in red. Posterior after 3 pointwise observations is considered. For left to right, top to bottom: i) True function (black), posterior and 2$\sigma$ confidence region (blue). True excursion region is highlighted in black and plug-in estimate is highlighted in blue. ii) Posterior excursion probability / coverage function (dark red). ii) Estimated excursion regions using Vorob'ev quantiles at level $\alpha=0.5$ (dark blue) and $\alpha=0.75$ (light blue). iv) Estimated excursion region using Vorob'ev expectation, Vorob'ev threshold is $\alpha_V=0.4$.
}
\label{fig:coverage_and_vorobev_1d}
\end{figure}

To illustrate the various Bayesian set estimation concepts introduced here, we apply them 
to a simple one-dimensional inverse problem, 
where one wants to estimate the excursion set above $1.0$ 
of a function $f:[-1, 1]\rightarrow \mathbb{R}$ after 3 pointwise evaluations 
of the function have been observed. Results are shown in \cref{fig:coverage_and_vorobev_1d}

\newpage

\section{Proofs}
\begin{proof}{(\cref{th:sequential})}
    We proceed by induction. The case $n=1$ follows from \cref{eq:cond_cov_matrix}. The induction step is directly given by \cref{th:seq_update}.
\end{proof}

\begin{proof}{(\cref{lemma:mult})}
    The product is computed using Algorithm 1. It involves multiplication
    of $A$ with the prior covariance, which costs
    $\mathcal{O}\left(\dimpred^2\dimA\right)$ and multiplication with all the
    previous intermediate matrices, which contribute
    $\mathcal{O}\left(\dimpred\dimChunk\dimA\right)$ and
    $\mathcal{O}\left(\dimChunk^2\dimA\right)$ respectively, at each stage.
\end{proof}

\begin{proof}{(\cref{lemma:repr_cost})}
The cost of computing the $\somestage$-th pushforward $\pushfwdI$ is 
$\mathcal{O}\left(\dimpred^2\dimChunk + \somestage(\dimpred \dimChunk^2
 + \dimChunk^3)\right)$. Summing this cost for all stages
 $\somestage=1,\dotsc,\stage$ then gives 
 $\mathcal{O}\left(\dimpred^2\datdimTot + \dimpred \datdimTot^2 +
    \datdim^2\dimChunk\right)$. To that cost, one should add the cost of
    computing $\invopI^{-1}$, which costs
    $\mathcal{O}\left(\dimChunk^3\right)$ at each stage, yielding a 
    $\mathcal{O}\left(\datdimTot \dimChunk^2\right)$ contribution to the total
    cost, which is dominated by $\datdimTot^2\dimChunk$ since $\dimChunk<\datdimTot$.
\end{proof}

\section{Supplementary Experimental Results}\label{sec:supplementary_exp}
We here include more detailed analysis of the results of \cref{sec:optimal_design} that do
not fit in the main text.\\

\begin{figure}[tbhp]
\centering
\subfloat[sample 1: small scenario]{\includegraphics[scale=0.55]{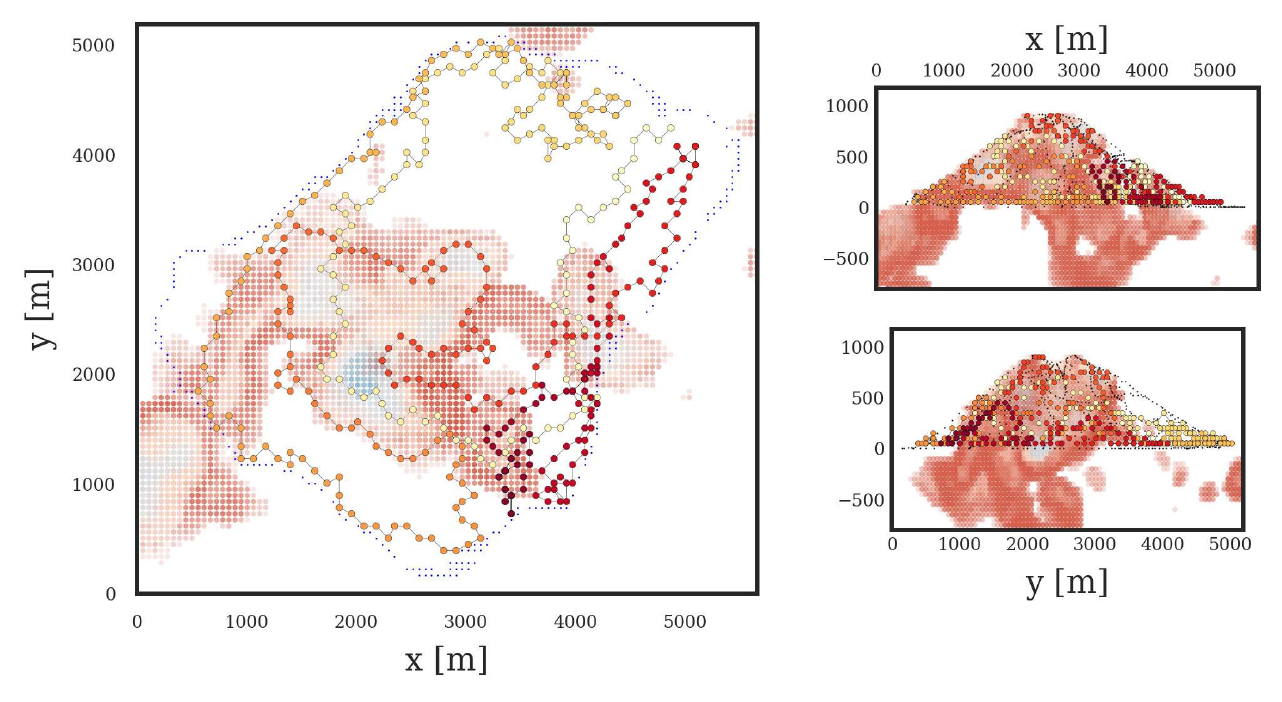}}
\subfloat[sample 1: large scenario]{\includegraphics[scale=0.55]{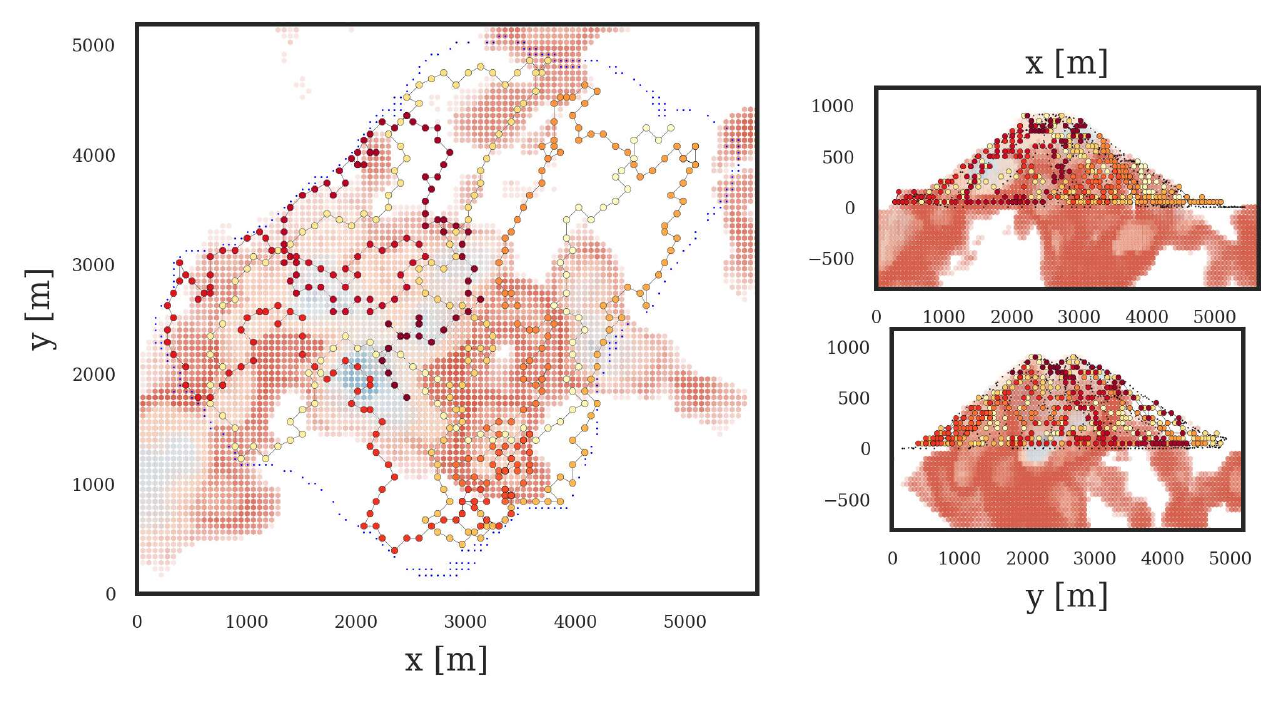}}\\
\subfloat[sample 2: small scenario]{\includegraphics[scale=0.55]{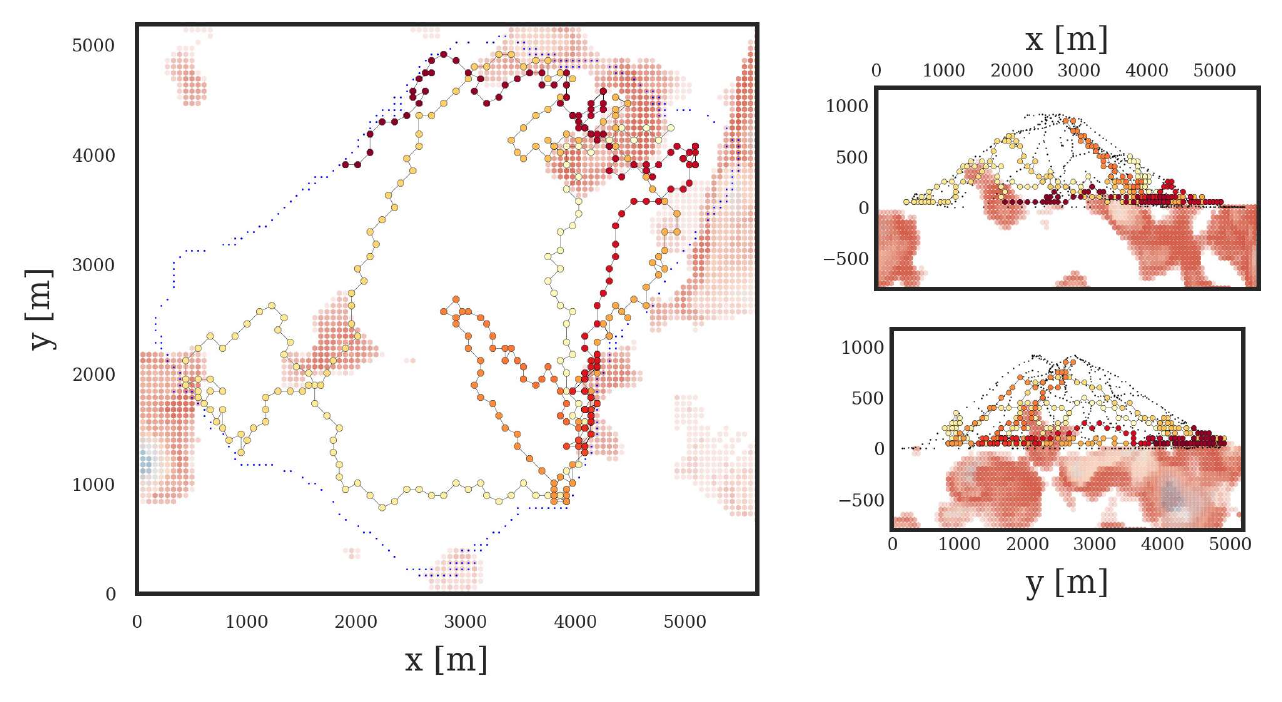}}
\subfloat[sample 2: large scenario]{\includegraphics[scale=0.55]{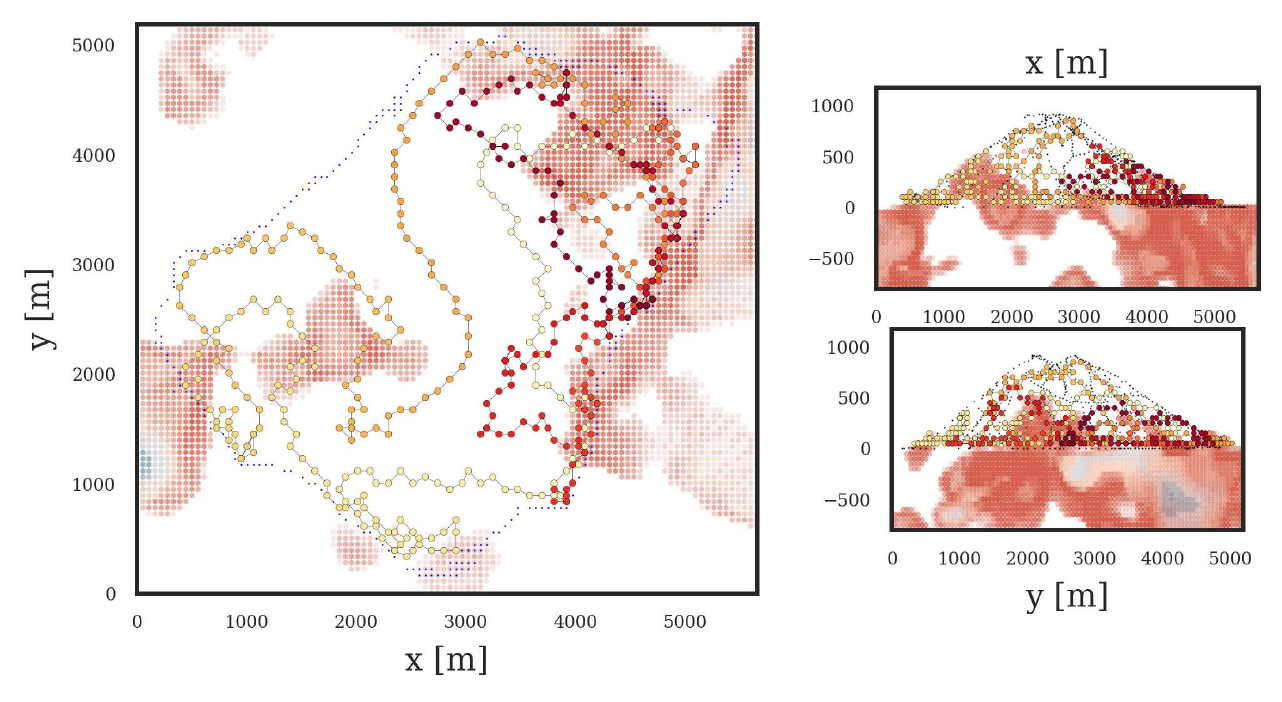}}\\
\subfloat[sample 3: small scenario]{\includegraphics[scale=0.55]{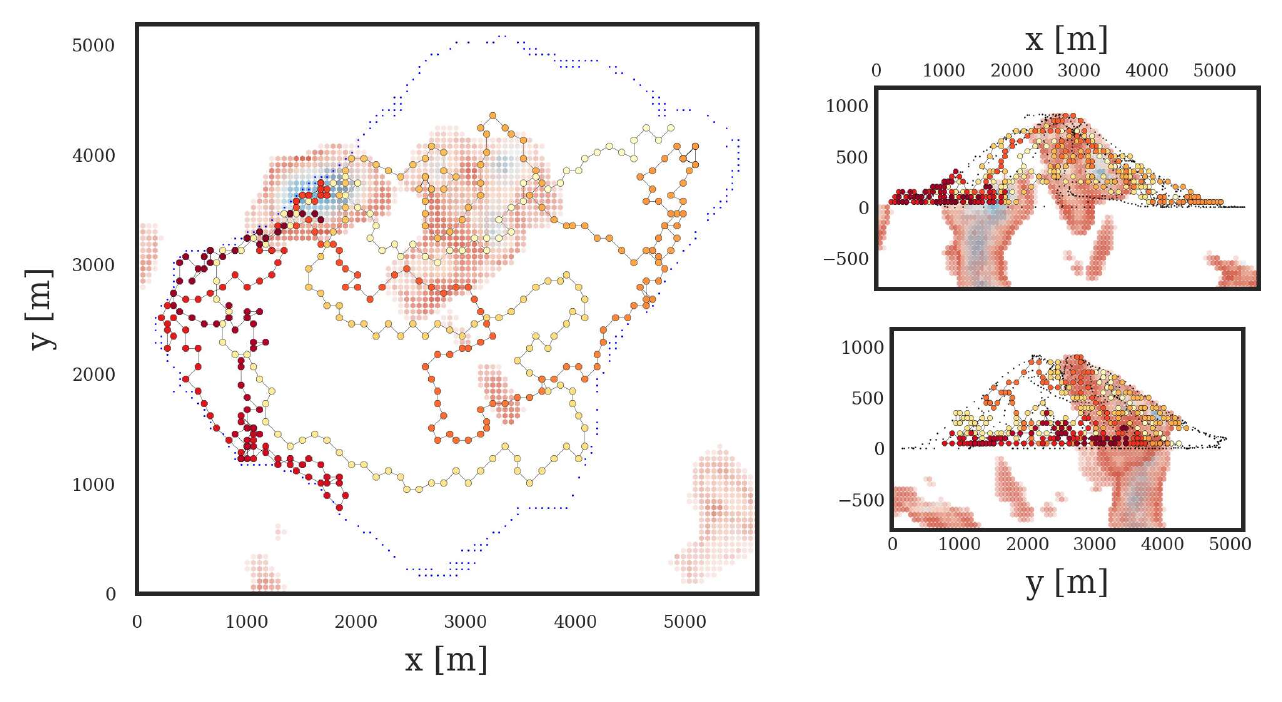}}
\subfloat[sample 3: large scenario]{\includegraphics[scale=0.55]{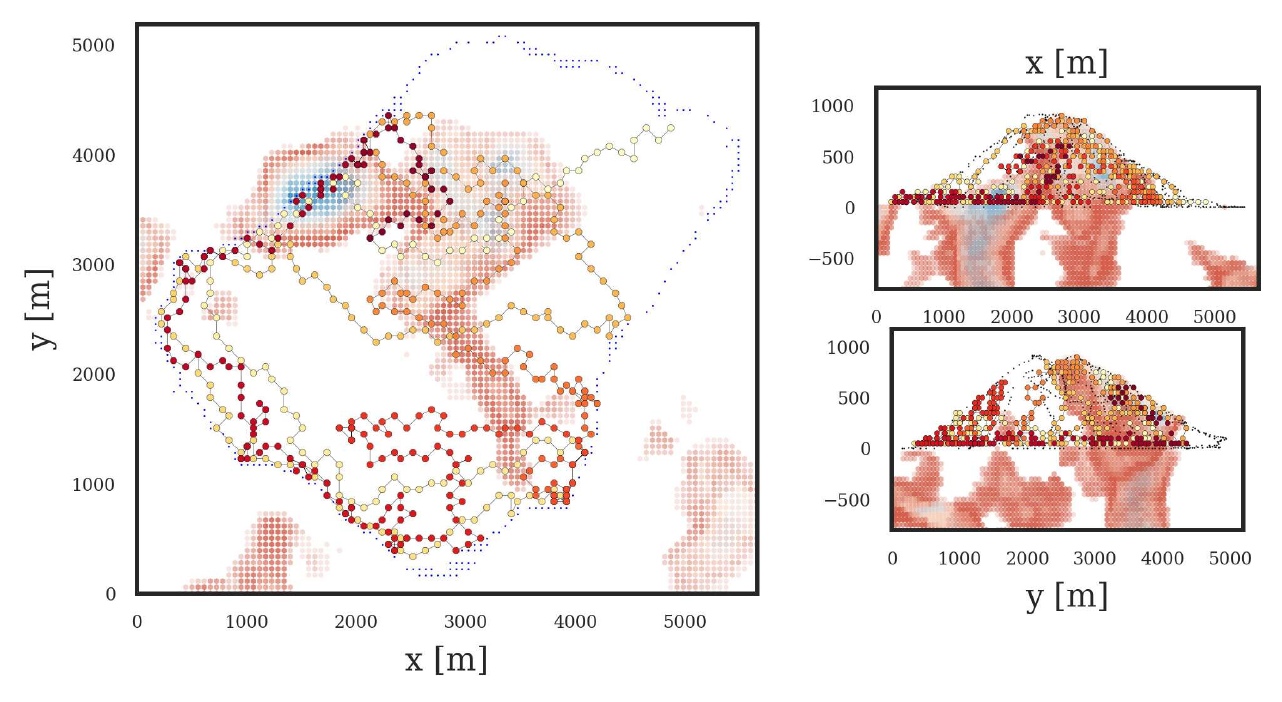}}\\
\subfloat[sample 4: small scenario]{\includegraphics[scale=0.55]{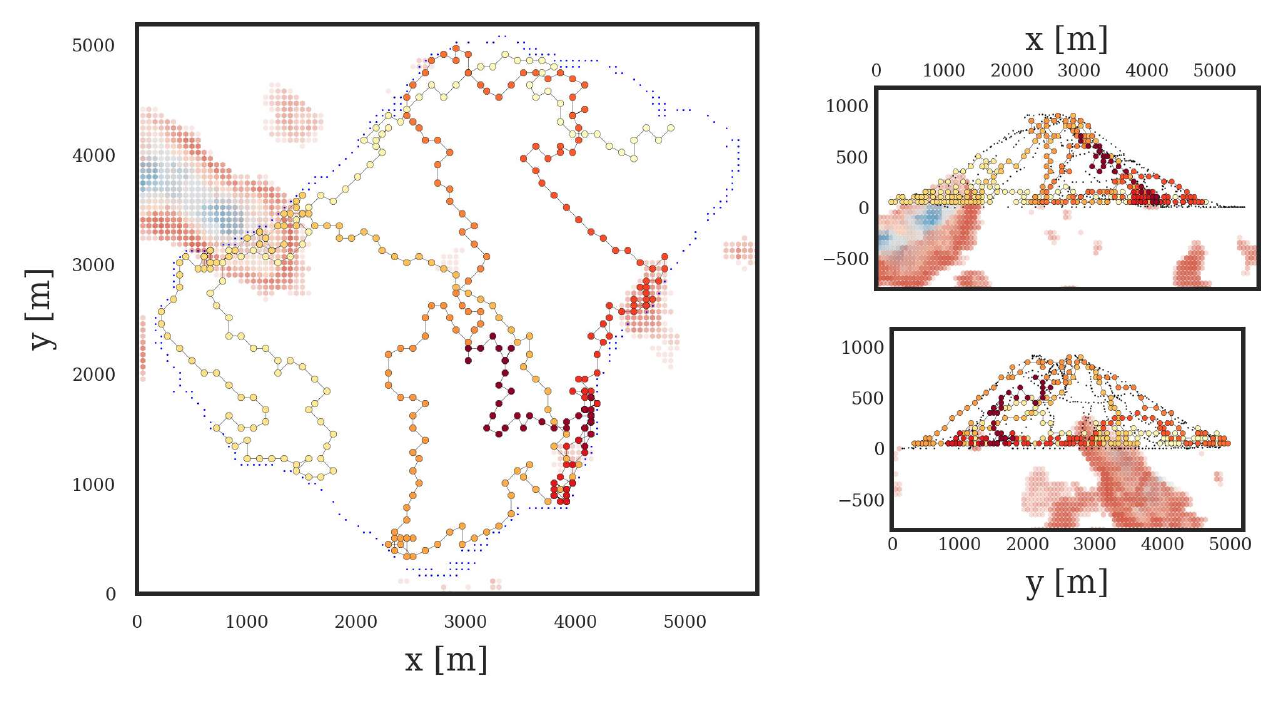}}
\subfloat[sample 4: large scenario]{\includegraphics[scale=0.55]{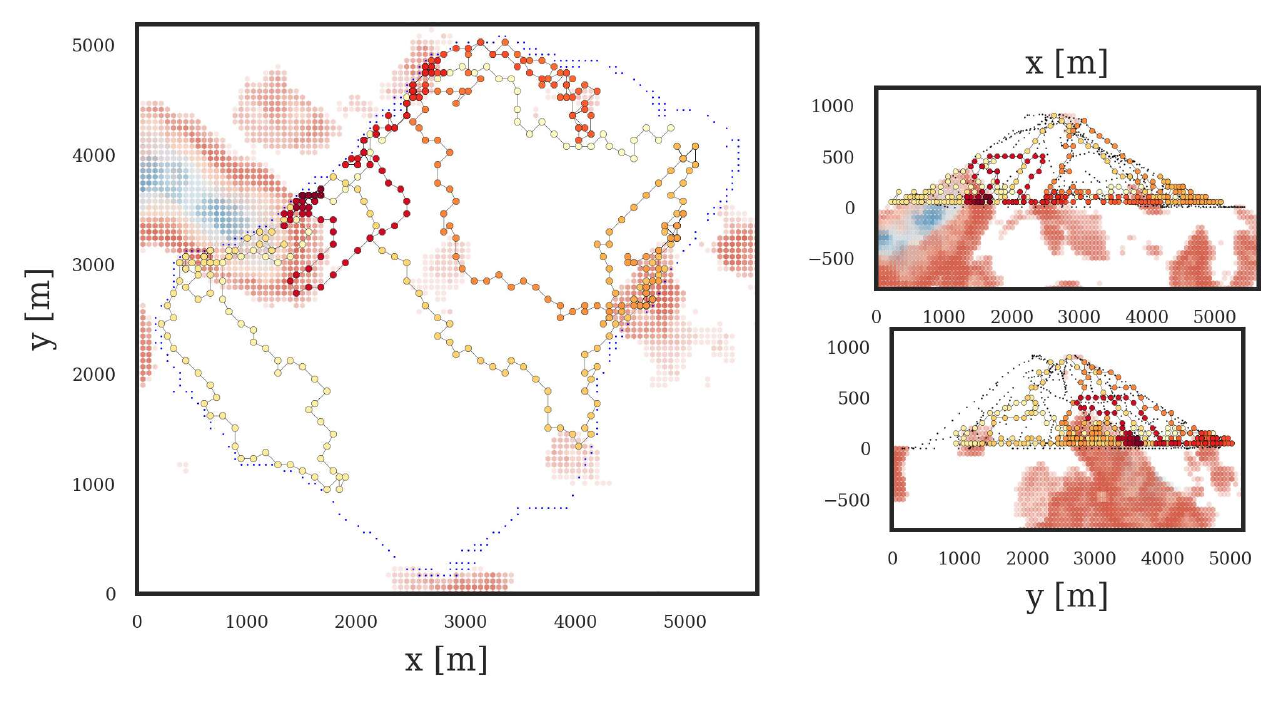}}\\
\subfloat[sample 5: small scenario]{\includegraphics[scale=0.55]{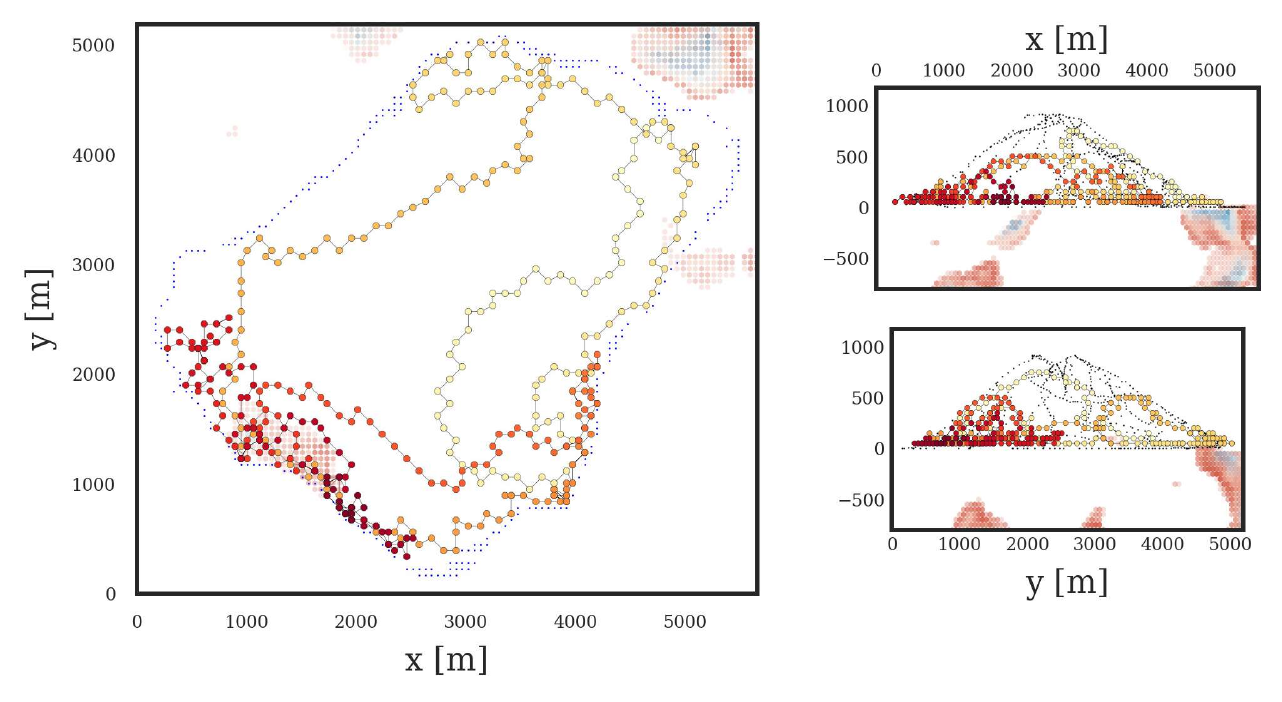}}
\subfloat[sample 5: large scenario]{\includegraphics[scale=0.55]{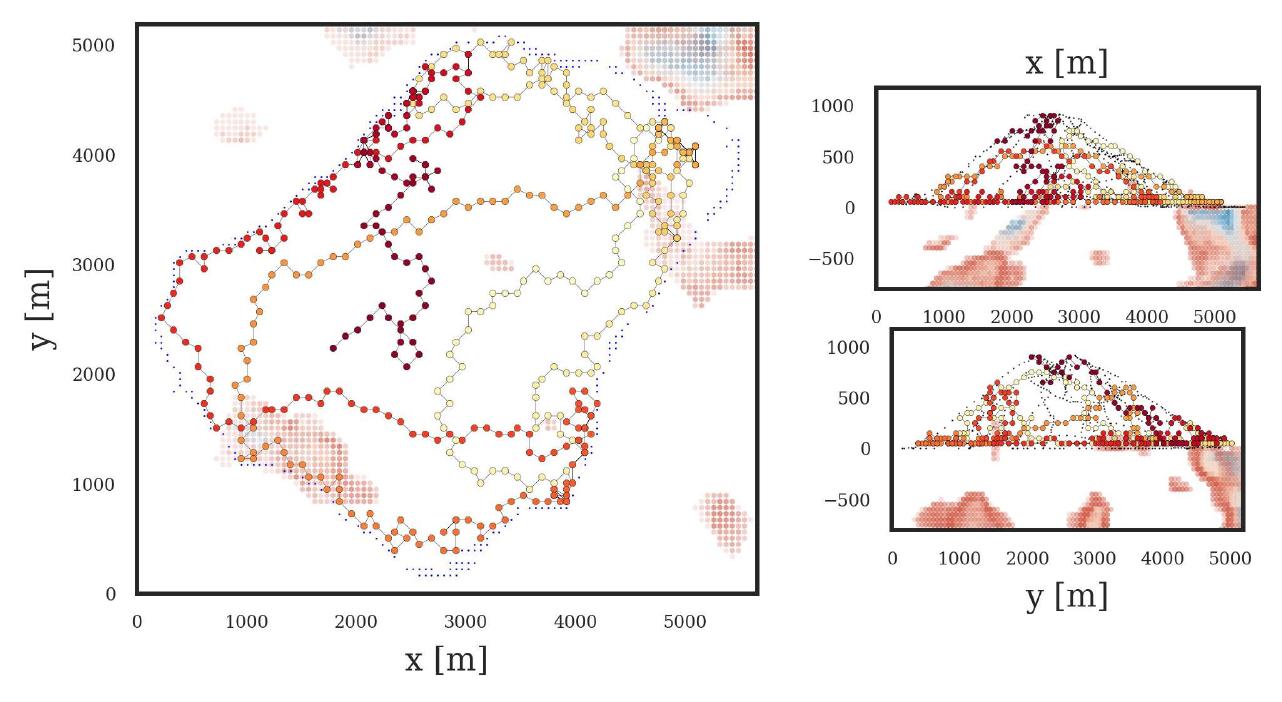}}\\
\caption{True excursion set and visited locations (wIVR strategy). Island boundary is shown in blue.}
\label{fig:excu_slice_appendix}
\end{figure}

\cref{fig:detection_evolution_big,fig:detection_evolution_small} showed that there are differences in detection performance for the different ground truths. These can be better understood by plotting the actual location of the excursion set for each of the ground truths as well as the observation locations chosen by the wIVR criterion, as done in \cref{fig:excu_slice_appendix}. 
One sees that the (comparatively) poor performance shown by \cref{fig:detection_evolution_small} for \textit{Sample 2} in the \textit{small} scenario may be explained by the fact that, for this ground truth, the excursion set is located mostly outside of the accessible data collection zone (island surface), so that the strategy is never able to collect data directly above the excursion.

\end{document}